\RequirePackage{ifxetex}
\RequirePackage{ifluatex}
\newif\ifxeorlua
\ifxetex\xeorluatrue\fi
\ifluatex\xeorluatrue\fi

\ifxeorlua

\RequirePackage{expl3}
\ExplSyntaxOn

\ExplSyntaxOff
\else
\RequirePackage{expl3}
\ExplSyntaxOn
\pdf_version_gset:n{1.5}
\ExplSyntaxOff
\fi

\documentclass[english, bibtex]{kththesis}

\ifbiblatex
    \usepackage[style=ieee,citestyle=numeric-comp]{biblatex}
\else
\fi

\ifxeorlua\relax
\else
\usepackage[scaled]{helvet}
\fi

\usepackage[force-eol=true]{scontents}              
\usepackage{pgffor}                 

\usepackage{xurl}                

\PassOptionsToPackage{dvipsnames, svgnames}{xcolor}
\usepackage{xcolor}

\usepackage[normalem]{ulem}
\usepackage{soul}
\usepackage{xspace}
\usepackage{braket}

\usepackage[binary-units=true, locale=US]{siunitx}

\usepackage{balance}
\usepackage{stmaryrd}
\usepackage{booktabs}
\usepackage{graphicx}	        
\usepackage{multirow}	        
\usepackage{tabularx}		    
\usepackage{mathtools}
\usepackage{algorithm} 
\usepackage{algorithmic}  
\usepackage{amsmath}
\usepackage[linesnumbered,ruled,vlined,algo2e]{algorithm2e}
\usepackage{optidef}
\usepackage{float}		        
\usepackage{pifont}

\usepackage{rotating}	    	
\usepackage{array}		        
\usepackage{mdwlist}            
\usepackage{setspace}           

\usepackage{enumitem}           

\usepackage{listings}		    

\usepackage{bytefield}          

\usepackage{todonotes}
\usepackage{notoccite} 

\usepackage{perpage}
\usepackage[perpage,para,symbol]{footmisc} 

\usepackage{tikz}
\usetikzlibrary{arrows,decorations.pathmorphing,backgrounds,fit,positioning,calc,shapes}
\usepackage{pgfmath}	

\usepackage{bm} 

\usepackage{csquotes}               
\usepackage{placeins}

\usepackage{comment}  
\usepackage{refcount}   

\usepackage{eso-pic}
\usepackage[absolute,overlay]{textpos}

\usepackage{xparse}  

\ifnomenclature
\usepackage[nocfg]{nomencl}  
\fi

\usepackage{longtable}  
\usepackage{lscape}     
\usepackage{needspace}  
\usepackage{metalogo}   

\usepackage{etoolbox}
\renewcommand{\bfseries}{\fontseries{b}\selectfont}
\robustify\bfseries
\newrobustcmd{\B}{\bfseries}

\ifluatex\empty
\else
\usepackage{morewrites}
\fi

\usepackage{svg}
\usepackage{pgfplots}
\usepackage{pgfplotstable}
 \pgfplotsset{compat=1.12}

    \tikzset{
        hatch distance/.store in=\hatchdistance,
        hatch distance=10pt,
        hatch thickness/.store in=\hatchthickness,
        hatch thickness=2pt
    }

    \makeatletter
    \pgfdeclarepatternformonly[\hatchdistance,\hatchthickness]{flexible hatch}
    {\pgfqpoint{0pt}{0pt}}
    {\pgfqpoint{\hatchdistance}{\hatchdistance}}
    {\pgfpoint{\hatchdistance-1pt}{\hatchdistance-1pt}}%
    {
        \pgfsetcolor{\tikz@pattern@color}
        \pgfsetlinewidth{\hatchthickness}
        \pgfpathmoveto{\pgfqpoint{0pt}{0pt}}
        \pgfpathlineto{\pgfqpoint{\hatchdistance}{\hatchdistance}}
        \pgfusepath{stroke}
    }
\makeatother

\definecolor{color1bg}{HTML}{1954a6}
\colorlet{color1bgFill}{color1bg!30!white}
\colorlet{color1bgDarkFill}{color1bg!90!white}

\definecolor{color2bg}{HTML}{24a0d8}
\colorlet{color2bgFill}{color2bg!30!white}
\colorlet{color2bgDarkFill}{color2bg!90!white}

\definecolor{color3bg}{HTML}{d85497}
\colorlet{color3bgFill}{color3bg!30!white}
\colorlet{color3bgDarkFill}{color3bg!90!white}

\definecolor{color4bg}{HTML}{b0c92b}
\colorlet{color4bgFill}{color4bg!30!white}
\colorlet{color4bgDarkFill}{color4bg!90!white}

\definecolor{color5bg}{HTML}{63666a}
\colorlet{color5bgFill}{color5bg!30!white}
\colorlet{color5bgDarkFill}{color5bg!90!white}

\pgfplotscreateplotcyclelist{rustcolors}{
  {color1bg,mark=none, pattern=
  {crosshatch}
  ,pattern color=color1bg},
  {color2bg,mark=none, pattern={vertical lines}, pattern color=color2bg},{color3bg,mark=none, pattern={horizontal lines}, pattern color=color3bg},{color4bg,mark=none, pattern={north east lines}, pattern color=color4bg},{color5bg,fill=color5bg,mark=none},
  {black,fill=gray,mark=none}
}

\pgfplotsset{cycle list name=rustcolors}
\pgfplotsset{/pgfplots/bar cycle list/.style={/pgfplots/cycle list name={rustcolors}}}

\usepackage{makecell}

\usepackage{pgf-pie}

\usetikzlibrary{tikzmark}

\ifluatex
\usepackage{luatexja-fontspec}
\setmainjfont{IPAexMincho} 
\fi

\usepackage{booktabs}
\usepackage{multirow}
\usepackage{adjustbox}
\usepackage{amssymb}
\usepackage{amsmath}
\usepackage{algorithm2e}
\usepackage{bbold}

\usepackage{xurl}

\usepackage[super]{nth}

\usepackage{subcaption}

\usepackage{tikz}
\usetikzlibrary{tikzmark}
\usetikzlibrary{calc}

\usepackage[all]{nowidow}
\usepackage{booktabs}       
\usepackage{colortbl}

\input{lib/kthcolors}
\makeatletter
\newcommand{\DeclareLatinAbbrev}[2]{%
  \DeclareRobustCommand{#1}{%
    \@ifnextchar{.}{\textit{#2}}{%
      \@ifnextchar{,}{\textit{#2.}}{%
        \@ifnextchar{!}{\textit{#2.}}{%
          \@ifnextchar{?}{\textit{#2.}}{%
            \@ifnextchar{)}{\textit{#2.}}{%
              {\textit{#2.,\ }}}}}}}}%
}
\makeatother
\DeclareLatinAbbrev{\eg}{e.g}
\DeclareLatinAbbrev{\Eg}{E.g}
\DeclareLatinAbbrev{\ie}{i.e}
\DeclareLatinAbbrev{\Ie}{I.e}
\DeclareLatinAbbrev{\etc}{etc}
\DeclareLatinAbbrev{\etal}{et~al}

\def\first {$(i)$\xspace}
\def\second{$(ii)$\xspace}
\def\third {$(iii)$\xspace}
\def\fourth{$(iv)$\xspace}

\definecolor{ForestGreen} {RGB}{34,  139,  34}
\definecolor{HeraldRed2}   {rgb}{0.81, 0.12, 0.15}

\newcommand{\refscolor} {blue}
\newcommand{\linkscolor}{HeraldRed2}
\newcommand{\urlscolor} {ForestGreen}

\ExplSyntaxOn
\newcommand\typestoredx[2]{\expandafter\__scontents_typestored_internal:nn\expandafter{#1} {#2}}
\ExplSyntaxOff
\makeatletter
\let\verbatimsc\@undefined
\let\endverbatimsc\@undefined
\lst@AddToHook{Init}{\hyphenpenalty=50\relax}
\makeatother

\lstnewenvironment{verbatimsc}
    {
    \lstset{%
        basicstyle=\ttfamily\tiny,
        backgroundcolor=\color{white},
        columns=[l]fixed,
        language=[LaTeX]TeX,
        keywordstyle=\color{red},
        breaklines=true,                 
        breakatwhitespace=true,          
        breakindent=0em,
        frame=none,                     
        postbreak={}                    
    }
}{}

\lstdefinestyle{[LaTeX]TeX}{
morekeywords={begin, todo, textbf, textit, texttt}
}

\newcolumntype{L}[1]{>{\raggedright\let\newline\\\arraybackslash\hspace{0pt}}p{#1}}

\ifbiblatex
    \usepackage[plainpages=false]{hyperref}
\else
    \usepackage[
    backref=page,
    pagebackref=false,
    plainpages=false,
    unicode=true,           
    bookmarks=true,         
    bookmarksopen=false,    
    pdfpagemode=UseNone,    
    destlabel,              
    pdfencoding=auto,       
    ]{hyperref}
    \makeatletter
    \ltx@ifpackageloaded{attachfile2}{
    }
    {\usepackage{backref}
    \renewcommand*{\backref}[1]{}
    \renewcommand*{\backrefalt}[4]{%
    \ifcase #1%
          \or [Page~#2.]%
          \else [Pages~#2.]%
    \fi%
    }
    }
    \makeatother

\fi
\usepackage[all]{hypcap}	

\usepackage[acronym, style=super, section=section, nonumberlist, nomain,
nopostdot]{glossaries}
\usepackage[]{glossaries-extra}
\ifinswedish
\fi

\newglossary[tlg]{readme}{tld}{tdn}{README acronyms}

\usepackage{doi}
\usepackage{cleveref}           
\usepackage{hyperxmp}           

\usepackage{attachfile2}

\makeglossaries

\ifxeorlua

\newacronym{UN}{UN}{United Nations}
\newacronym{SDG}{SDG}{Sustainable Development Goal}
\newacronym{RS}{RS}{Remote Sensing}
\newacronym{ESA}{ESA}{European Space Agency}
\newacronym{S2}{S2}{Sentinel-2}
\newacronym{ANN}{ANN}{Artificial Neural Network}
\newacronym{CNN}{CNN}{Convolutional Neural Network}
\newacronym{FCN}{FCN}{Fully Convolutional Network}
\newacronym{DNN}{DNN}{Deep Neural Network}
\newacronym{AI}{AI}{Artificial Intelligence}
\newacronym{DL}{DL}{Deep Learning}
\newacronym{SSL}{SSL}{Semi Supervised Learning}
\newacronym{EO}{EO}{Earth Observation}
\newacronym{BOA}{BOA}{Bottom-Of-Atmosphere}
\newacronym{TOA}{TOA}{Top-Of-Atmosphere}
\newacronym{GPU}{GPU}{Graphics Processing Unit}

\newacronym{CE}{CE}{Cross Entropy}
\newacronym{BCE}{BCE}{Binary Cross Entropy}
\newacronym{BBCE}{BBCE}{Balanced Binary Cross Entropy}
\newacronym{FL}{FL}{Focal Loss}

\newacronym{PA}{PA}{Pixel Accuracy}
\newacronym{IoU}{IoU}{Intersection over Union}
\newacronym{mIoU}{mIoU}{Mean Intersection over Union}

\newacronym{NaN}{NaN}{Not a number}                
\else

\newacronym{UN}{UN}{United Nations}
\newacronym{SDG}{SDG}{Sustainable Development Goal}
\newacronym{RS}{RS}{Remote Sensing}
\newacronym{ESA}{ESA}{European Space Agency}
\newacronym{S2}{S2}{Sentinel-2}
\newacronym{ANN}{ANN}{Artificial Neural Network}
\newacronym{CNN}{CNN}{Convolutional Neural Network}
\newacronym{FCN}{FCN}{Fully Convolutional Network}
\newacronym{DNN}{DNN}{Deep Neural Network}
\newacronym{AI}{AI}{Artificial Intelligence}
\newacronym{DL}{DL}{Deep Learning}
\newacronym{SSL}{SSL}{Semi Supervised Learning}
\newacronym{EO}{EO}{Earth Observation}
\newacronym{BOA}{BOA}{Bottom-Of-Atmosphere}
\newacronym{TOA}{TOA}{Top-Of-Atmosphere}
\newacronym{GPU}{GPU}{Graphics Processing Unit}

\newacronym{CE}{CE}{Cross Entropy}
\newacronym{BCE}{BCE}{Binary Cross Entropy}
\newacronym{BBCE}{BBCE}{Balanced Binary Cross Entropy}
\newacronym{FL}{FL}{Focal Loss}

\newacronym{PA}{PA}{Pixel Accuracy}
\newacronym{IoU}{IoU}{Intersection over Union}
\newacronym{mIoU}{mIoU}{Mean Intersection over Union}

\newacronym{NaN}{NaN}{Not a number}
\fi

\authorsLastname{Marini}
\authorsFirstname{Luca}
\email{lucamar@kth.se}
\kthid{u1lroecd}
\authorsSchool{\schoolAcronym{EECS}}

\authorCityCountryDate{15 December 2023}

\supervisorAsLastname{Meoni}
\supervisorAsFirstname{Gabriele}
\supervisorAsOrganization{Previously with $\Phi$-lab division, European Space Agency. Currently with Department of Space Engineering, Aerospace Engineering, Delft University of Technology}

\supervisorBsLastname{Gerard}
\supervisorBsFirstname{Sebastian}
\supervisorBsEmail{sgerard@kth.se}
\supervisorBsKTHID{u1tujxcj}
\supervisorBsSchool{\schoolAcronym{EECS}}
\supervisorBsDepartment{Computer Science}

\examinersLastname{Sullivan}
\examinersFirstname{Josephine}
\examinersEmail{sullivan@kth.se}
\examinersKTHID{u18xk10j}
\examinersSchool{\schoolAcronym{EECS}}
\examinersDepartment{Computer Science}

\hostcompany{European Space Agency, AI Sweden, TUDelft} 

\courseCycle{2}
\courseCode{DA233X}
\courseCredits{30.0}

\programcode{TMAIM}
\degreeName{Machine Learning}
\subjectArea{Machine Learning}
\nationalsubjectcategories{10201, 10207}

\title{Semi-supervised learning for marine anomaly detection on board satellites}

\alttitle{Semi-övervakad inlärning för upptäckt av marina anomalier ombord på satelliter}

\EnglishKeywords{Machine Learning, Computer Vision, Deep Leaning, Semi-Supervised Learning, Semantic Segmentation, Satellite Imagery, Marine Anomalies, U-Net, Onboard AI}
\SwedishKeywords{Maskininlärning, Datorseende, Deep Leaning, Semi-Supervised Learning, Semantisk segmentering, Satellitbilder, Marina Avvikelser, U-Net, AI ombord}

\presentationDateAndTimeISO{2023-12-15 14:00}
\presentationLanguage{eng}
\presentationRoom{via Zoom https://kth-se.zoom.us/j/64549441616}
\presentationAddress{}
\presentationCity{Stockholm}

\opponentsNames{Jorge Santiago Roman Avila}

\makeatletter
\ifx\@subtitle\@empty
    \newcommand{\mytitle}{\@title}
\else
    \ifinswedish
        \newcommand{\mytitle}{\@title\xspace–\xspace\@subtitle}
    \else
        \newcommand{\mytitle}{\@title: \@subtitle}
    \fi
\fi
\makeatother

\makeatletter
\ifx\@altsubtitle\@empty\relax
    \newcommand{\myalttitle}{\@alttitle}
\else
    \ifinswedish
        \newcommand{\myalttitle}{\@alttitle: \@altsubtitle}
    \else
    \newcommand{\myalttitle}{\@alttitle\xspace–\xspace\@altsubtitle}
    \fi
    
\fi
\makeatother
\hypersetup{
     pdfsubject={\myalttitle}        
}

\ifinswedish
\XMPLangAlt{en}{pdfsubject={\myalttitle}}
\else
\XMPLangAlt{sv}{pdfsubject={\myalttitle}}
\fi

\ifinswedish
\hypersetup{%
    pdflang={sv},
    pdfmetalang={sv},
    pdftitle={\mytitle}        
}
\XMPLangAlt{en}{pdftitle={\myalttitle}}
\else
\hypersetup{%
    pdflang={en},
    pdfmetalang={en},
    pdftitle={\mytitle}        
}
\XMPLangAlt{sv}{pdftitle={\myalttitle}}
\fi

\makeatletter
\ltx@ifpackageloaded{hyperxmp}{
\ifx\@secondAuthorsLastname\@empty
\StrSubstitute{\@authorsLastname}{,}{\hyxmp@comma}[\@authorsLastnameXMP]
    \newcommand{\myauthor}{\xmpquote{\@authorsFirstname\space\@authorsLastnameXMP}} 
\else
\StrSubstitute{\@authorsLastname}{,}{\hyxmp@comma}[\@authorsLastnameXMP]
\StrSubstitute{\@secondAuthorsLastname}{,}{\hyxmp@comma}[\@secondAuthorsLastnameXMP]
    \newcommand{\myauthor}{\xmpquote{\@authorsFirstname\space\@authorsLastnameXMP},
\xmpquote{\@secondAuthorsFirstname\space\@secondAuthorsLastnameXMP}}
\fi
}{
\ifx\@secondAuthorsLastname\@empty
    \newcommand{\myauthor}{\@authorsFirstname\space\@authorsLastname} 
\else
    \newcommand{\myauthor}{\@authorsFirstname\space\@authorsLastname,
\space\@secondAuthorsFirstname\space\@secondAuthorsLastname}
\fi
}
\makeatother

\hypersetup{
     pdfauthor={\myauthor}      
}

\makeatletter
\ifx\@EnglishKeywords\@empty
    \ifx\@SwedishKeywords\@empty
        \newcommand{\mykeywords}{}
    \else
    \newcommand{\mykeywords}{\@SwedishKeywords}
    \fi
\else
    \ifx\@SwedishKeywords\@empty
        \newcommand{\mykeywords}{\@EnglishKeywords}
    \else
        \ifinswedish
            \newcommand{\mykeywords}{\@SwedishKeywords, \@EnglishKeywords}
        \else
            \newcommand{\mykeywords}{\@EnglishKeywords, \@SwedishKeywords}
        \fi
    \fi
\fi
\makeatother

\hypersetup{
     pdfkeywords={\mykeywords}        
}        

\makeatletter
\ifx\@secondkthid\@empty\relax
    \newcommand{\mykthids}{author: \@kthid}
\else
    \newcommand{\mykthids}{author: \@kthid,\xspace
    secondauthor: \@secondkthid}
\fi
\makeatother

\hypersetup{
     pdfcontactemail={\mykthids}        
}

\makeatletter
\ifinswedish
\hypersetup{
        pdfvolumenum={\@thesisSeries},        
        pdfissuenum={\@thesisSeriesNumber},
        pdfpublisher={Kungliga Tekniska högskolan (KTH)},
        pdfpubtype={report}
}
\else
\hypersetup{
        pdfvolumenum={\@thesisSeries},        
        pdfissuenum={\@thesisSeriesNumber},
        pdfpublisher={KTH Royal Institute of Technology},
        pdfpubtype={report}
}
\fi
\makeatother

\hypersetup{
	colorlinks  = true,
	breaklinks  = true,
	linkcolor   = \linkscolor,
	urlcolor    = \urlscolor,
	citecolor   = \refscolor,
	anchorcolor = black
}

\begin{document}
%
\selectlanguage{english}

\pagenumbering{alph}
\kthcover
\clearpage\thispagestyle{empty}\mbox{} 

\titlepage
\bookinfopage

\frontmatter
\setcounter{page}{1}
\begin{abstract}
  \markboth{\abstractname}{}
\begin{scontents}[store-env=lang]
eng
\end{scontents}
\begin{scontents}[store-env=abstracts,print-env=true]
Aquatic bodies face numerous environmental threats caused by several marine anomalies. Marine debris can devastate habitats and endanger marine life through entanglement, while harmful algal blooms can produce toxins that negatively affect marine ecosystems. Additionally, ships may discharge oil or engage in illegal and overfishing activities, causing further harm.
These marine anomalies can be identified by applying trained deep learning (\gls{DL}) models on multispectral satellite imagery. Furthermore, the detection of other anomalies, such as clouds, could be beneficial in filtering out irrelevant images. However, \gls{DL} models often require a large volume of labeled data for training, which can be both costly and time-consuming, particularly for marine anomaly detection where expert annotation is needed. A potential solution is the use of semi-supervised learning methods, which can also utilize unlabeled data.
In this project, we implement and study the performance of \textit{FixMatch for Semantic Segmentation}, a semi-supervised algorithm for semantic segmentation.
Firstly, we found that semi-supervised models perform best with a high confidence threshold of 0.9 when there is a limited amount of labeled data.
Secondly, we compare the performance of semi-supervised models with fully-supervised models under varying amounts of labeled data. Our findings suggest that semi-supervised models outperform fully-supervised models with limited labeled data, while fully-supervised models have a slightly better performance with larger volumes of labeled data. We propose two hypotheses to explain why fully-supervised models surpass semi-supervised ones when a high volume of labeled data is used.
All of our experiments were conducted using a U-Net model architecture with a limited number of parameters to ensure compatibility with space-rated hardware. This approach is a preliminary step towards deploying \gls{AI} models onboard satellites, a move that could potentially reduce downlink bandwidth usage and expand the satellites’ data acquisition area.
The code of our project is open-source and available at \href{https://github.com/lucamarini22/marine-anomaly-detection}{\nolinkurl{github.com/lucamarini22/marine-anomaly-detection}}.
\end{scontents}

\subsection*{Keywords}
\begin{scontents}[store-env=keywords,print-env=true]
\InsertKeywords{english}
\end{scontents}

\end{abstract}
\cleardoublepage
\babelpolyLangStart{swedish}
\begin{abstract}
    \markboth{\abstractname}{}
\begin{scontents}[store-env=lang]
swe
\end{scontents}
\begin{scontents}[store-env=abstracts,print-env=true]
Vattenförekomster står inför många miljöhot som orsakas av flera marina anomalier. Marint skräp kan förstöra livsmiljöer och hota det marina livet genom att trassla in sig, medan skadliga algblomningar kan producera toxiner som påverkar de marina ekosystemen negativt. Dessutom kan fartyg släppa ut olja eller ägna sig åt olaglig verksamhet och överfiske, vilket orsakar ytterligare skada.
Dessa marina avvikelser kan identifieras genom att använda utbildade modeller för djupinlärning (\gls{DL}) på multispektrala satellitbilder. Dessutom kan detektering av andra avvikelser, t.ex. moln, vara till nytta för att filtrera bort irrelevanta bilder. Men \gls{DL}-modeller kräver ofta en stor mängd märkta data för träning, vilket kan vara både kostsamt och tidskrävande, särskilt för marin anomalidetektering där expertkommentarer behövs. En potentiell lösning är att använda semi-supervised learning-metoder, som också kan använda omärkta data.
I det här projektet implementerar vi \textit{FixMatch for Semantic Segmentation}, en semi-övervakad algoritm för semantisk segmentering, och studerar dess prestanda.
För det första fann vi att semi-övervakade modeller presterar bäst med en hög konfidensgräns på 0,9 när det finns en begränsad mängd märkta data.
För det andra jämför vi prestandan hos halvövervakade modeller med helt övervakade modeller under varierande mängder märkta data. Våra resultat tyder på att semi-övervakade modeller överträffar helt övervakade modeller med begränsad märkt data, medan helt övervakade modeller har en något bättre prestanda med större volymer märkt data. Vi föreslår två hypoteser för att förklara varför helt övervakade modeller överträffar semi-övervakade modeller när en stor mängd märkta data används.
Alla våra experiment genomfördes med en U-Net-modellarkitektur med ett begränsat antal parametrar för att säkerställa kompatibilitet med rymdklassad hårdvara. Detta tillvägagångssätt är ett preliminärt steg mot att distribuera \gls{AI}-modeller ombord på satelliter, ett steg som potentiellt kan minska bandbreddsanvändningen för nedlänkning och utöka satelliternas datainsamlingsområde.
Koden för vårt projekt är öppen källkod och finns tillgänglig på \href{https://github.com/lucamarini22/marine-anomaly-detection}{\nolinkurl{github.com/lucamarini22/marine-anomaly-detection}}.
\end{scontents}
\subsection*{Nyckelord}
\begin{scontents}[store-env=keywords,print-env=true]
\InsertKeywords{swedish}
\end{scontents}
\end{abstract}
\babelpolyLangStop{swedish}

\clearpage

\section*{Acknowledgments}
\markboth{Acknowledgments}{}

I am very grateful to my supervisors, Gabriele Meoni and Sebastian Gerard, for their invaluable and consistent feedback. Their guidance has enriched me both technically and personally, and I have gained immensely from their wisdom.
\newline
\newline
I would also like to thank Josephine Sullivan for guiding me through the initial steps of the thesis and for the contribution of being the examiner of this project. 
\newline
\newline
I want to express my gratitude to Vinutha Magal Shreenath from AI Sweden for setting up this opportunity. Her dedication and interest in ensuring the successful progression of my thesis were truly commendable. I am thankful to Pablo Gómez too for his code reviews and priceless support. I want to thank AI Sweden for granting me access to the Edge Learning Lab facilities. My time spent in the Stockholm office was truly enjoyable. The company of the wonderful people I met there greatly enriched my experience. Having fika, engaging in ping pong matches, and the stimulating conversations have left an indelible mark on me.
\newline
\newline
Finally, I would like to thank my friends, my family, and my girlfriend for supporting me through this journey. I could have not made it without you all.

\acknowlegmentssignature

\fancypagestyle{plain}{}
\renewcommand{\chaptermark}[1]{ \markboth{#1}{}} 
\tableofcontents
  \markboth{\contentsname}{}

\cleardoublepage
\listoffigures

\cleardoublepage

\listoftables
\cleardoublepage


\newglossarystyle{mylong}{%
  \setglossarystyle{long}%
  \renewenvironment{theglossary}%
     {\begin{longtable}[l]{@{}p{\dimexpr 2cm-\tabcolsep}p{0.8\hsize}}}
     {\end{longtable}}%
 }
\glsaddall
\printglossary[style=mylong, type=\acronymtype, title={List of acronyms and abbreviations}]
\label{pg:lastPageofPreface}
\mainmatter
\glsresetall
\renewcommand{\chaptermark}[1]{\markboth{#1}{}}
\selectlanguage{english}

\chapter{Introduction}
\label{ch:introduction}

This chapter describes the specific problem that this thesis addresses, 
states the research questions, introduces the research methodology, describes the delimitations, and outlines the structure of the thesis.

\section{Background}
\label{sec:background}

Interest in using remote sensing to detect marine anomalies has been growing in recent years \cite{themistocleous2020investigating, kikaki2022marida, mifdal2021towards}. In this project, marine anomalies are certain non-water entities in or above the water. Specifically, our focus is on marine debris, ships, clouds, and algae. Marine debris, primarily composed of slowly decomposing plastics, has a wide range of negative effects, including environmental damage and chemical pollution \cite{iniguez2016marine}. Ships are detected for maritime safety purposes, such as traffic monitoring, illegal fishing and pollution control  \cite{corbane2010complete}. Cloud detection is a key pre-processing step in optical satellite-based remote sensing to ensure clear images. Certain algae species produce harmful compounds when present in high concentrations \cite{carvalho2010satellite}.

Earth observation (\gls{EO}) satellites have revolutionized our understanding of Earth by providing unprecedented remote sensing data. They offer insights into global cloud patterns, vegetation cover, surface structures, etc. In particular, Sentinel-2 is an \gls{EO} mission developed by \gls{ESA}, consisting of two artificial satellites. These satellites acquire multispectral images, which consist of several channels known as bands. Several of these bands contain information beyond what is visible to the human eye.

Various studies have developed spectral indices to detect marine debris using multispectral satellite imagery \cite{biermann2020finding, themistocleous2020investigating}. However, differentiating marine debris from marine water and other floating objects like ships is a complex task \cite{kikaki2022marida}. It is particularly challenging to create spectral indexes that can effectively identify various types of marine anomalies.

Other studies utilized machine and deep learning techniques to perform semantic segmentation of marine debris and other anomalies \cite{kikaki2022marida, mifdal2021towards}. These works utilized a category of neural networks known as Convolutional Neural Networks (\glspl{CNN}), which has emerged over the past decade as the preferred deep learning approach for computer vision tasks, including semantic segmentation. This is due to their ability to integrate knowledge specific to images into their structure. The advantage of using deep learning techniques is that neural networks can autonomously learn to differentiate between various anomalies by learning from labeled data. This eliminates the need for manually creating spectral indices.

However, labeling data for semantic segmentation is a time-consuming task due to the requirement for high-quality pixel-level annotations for high-resolution images. This process is also costly because it needs specialist annotators to accurately identify pixels, particularly when trying to label marine anomalies in satellite images.

The need of labeled data could be reduced by using semi-supervised learning, a branch of machine learning that utilizes both labeled and unlabeled data \cite{van2020survey}. Semi-supervised learning techniques have proven effective in tasks such as image classification \cite{sohn2020fixmatch}, even with multispectral images \cite{gomez2021msmatch}. Given the abundance of unlabeled data from daily satellite acquisitions and the scarcity of labeled data in Earth observation, semi-supervised learning presents a promising approach.

The majority of \gls{EO} satellites do not use Artificial Intelligence (\gls{AI}) algorithms onboard.  Indeed, \gls{EO} satellites capture data, which is then transmitted to ground stations when the satellites enter their designated coverage area \cite{furano2020towards}. In this scenario, \gls{AI} algorithms are applied on the ground after the data has been transmitted by the satellite. However, applying \gls{AI} models directly on spacecraft could reduce downlink bandwidth usage \cite{furano2020towards} and extend the satellite acquisition area. In the context of marine anomaly detection, an onboard \gls{AI} model could be programmed to only transmit images containing anomalies back to ground stations. This approach would not only reduce the bandwidth usage but also free up storage space on the satellite. The additional storage could then be utilized to collect images beyond the satellite's originally designated image acquisition area, thereby expanding its coverage.

In this study, we employ semi-supervised learning to compare the performance of semi-supervised models (which use both labeled and unlabeled data) with fully-supervised models (which use only labeled data). Specifically, we examine these models in the context of semantic segmentation of marine anomalies using multispectral images. Moreover, we consider a model architecture that has been previously studied for space missions \cite{ghasemi2023feasibility} and whose computational operations are compatible with space-rated hardware.
The code of our project is open-source and available at \href{https://github.com/lucamarini22/marine-anomaly-detection}{\nolinkurl{github.com/lucamarini22/marine-anomaly-detection}}.



\section{Research Questions}
\label{sec:problem}




The research questions of our project are the following:
\begin{enumerate}
    \item Can semi-supervised models outperform fully-supervised models in the task of marine anomaly detection when trained with the same amount of labeled data?
    \item Is it feasible to achieve satisfactory performance in marine anomaly detection using a model that is compatible with space-rated hardware?
\end{enumerate}








\section{Research Methodology}


In this project, we conduct three experiments. The first two are hyperparameter selection experiments, whose results are applied to the third and main experiment. This final experiment is designed to compare the performance of semi-supervised and fully-supervised learning models.

The evaluation metric used across all experiments is the mean Intersection over Union (\gls{mIoU}). For a single model undergoing multiple epochs of training on a dataset, we assess its performance after each epoch based on the mIoU on the validation set. The version of the model trained during the epoch that yields the highest mIoU is chosen for the final evaluation on the test set.

The mIoU, calculated on the validation set, is also employed to compare models trained with different hyperparameters. The model with the highest mIoU is selected for the final evaluation on the test set.

This evaluation methodology is applied to both fully-supervised and semi-supervised learning models. Finally, we compare the performance of a fully-supervised model with its semi-supervised counterpart by computing the mIoU score on the test set.

In addition, we visualize the predictions of both types of models to better understand the cases in which the difference of performance among them is most pronounced.

\section{Delimitations}


In this project, our focus is not on comparing various semi-supervised methodologies. Instead, we only utilize FixMatch for Semantic Segmentation as our semi-supervised method.

Upon examining the data, we observed some unusual labels. Despite this, our study does not involve the re-labeling of these atypical semantic segmentation maps.

In addition, we did not perform perform extended hyperparameter spaces when conducting our experiments.

Lastly, we did not execute the deep learning models on space-rated hardware.

\section{Structure of the thesis}
\label{sec:structure}

Chapter \ref{ch:background} presents relevant background information about \first marine anomaly detection, \second remote sensing, Earth observation, and onboard artificial intelligence, \third \gls{CNN}-based semantic segmentation deep learning algorithms, and deep semi-supervised learning techniques. Chapter \ref{ch:methods} firstly presents the utilized dataset and justifies the choices of loss and model architecture. Secondly, it presents the semi-supervised algorithm (FixMatch for Semantic Segmentation) we implemented, and describes our evaluation metrics and experimental design used in our experiments. Thirdly, it outlines both the preliminary and main experiments that were conducted. It also reports libraries and packages used for the implementation, and the hardware utilized in this project. Chapter \ref{ch:resultsAndAnalysis} presents the results of our experiments. Chapter \ref{ch:discussion} delves into a comprehensive discussion of the experimental results. Chapter \ref{ch:conclusionsAndFutureWork} draws the final conclusions, acknowledges the limitations, and reflects on the ethical and environmental implications of this work. It also outlines potential future work for future enhancements and expansions of this project.

\cleardoublepage

\chapter{Background}
\label{ch:background}



This chapter presents the theoretical background of this project. Section \ref{section:marine_anomalies_detection_background} describes what marine anomalies are and why it is interesting to monitor them. Section \ref{section:remote_sensing} discusses basic concepts of remote sensing, Earth observation, satellites, and how marine anomalies can be detected by using traditional approaches. Section \ref{section:ai_dl_mad} introduces some notions of artificial intelligence (\gls{AI}) and deep learning (\gls{DL}), describes the task of semantic segmentation, explores possible problems that can arise when detecting marine anomalies with deep learning techniques, and introduces semi-supervised learning and some of its methods. 

\section{Marine Anomalies Detection}
\label{section:marine_anomalies_detection_background}

In recent years, there has been an increasing interest in detecting marine anomalies by means of remote sensing \cite{themistocleous2020investigating, kikaki2022marida, mifdal2021towards}. In this project, marine anomalies refer to certain entities that are in (or above) the water and are not marine water itself. In particular, marine debris, ships, clouds, and algae are the marine anomalies considered in the scope of this project.

Marine debris causes detrimental effects spanning environmental, economic, safety, health, and cultural domains. The majority of this marine litter, predominantly composed of plastics, decomposes at an extremely slow pace. This results in a steady yet substantial build-up in coastal and marine ecosystems. Over time, marine debris significantly contributes to the chemical pollution in the marine environment \cite{iniguez2016marine}.

Ships are another example of marine anomalies. Indeed, detecting them through imagery from remote sensing is a crucial application for ensuring maritime safety. This includes, but is not limited to, monitoring maritime traffic, safeguarding against unlawful fishing activities, controlling oil discharges, and observing sea pollution \cite{corbane2010complete}.

Clouds can also be considered as anomalies because in optical satellite-based remote sensing, the detection of clouds is a crucial pre-processing step. In fact, analyses of optical images are often disturbed by cloud coverage \cite{jeppesen2019cloud}. Therefore, the detection of cloud can be can be used to keep cloud-free images and discard images that are covered with too many clouds. 

Lastly, some species of algae, present in high concentrations, are also considered marine anomalies because they have the ability to generate poisonous compounds, frequently referred to as “harmful algal blooms” which can poison and produce negative effects on human health and sea life \cite{carvalho2010satellite}.

\section{Remote Sensing}
\label{section:remote_sensing}

Remote sensing (\gls{RS}) is defined as "the acquisition of information about an object without being in physical contact with it" \cite{doi:https://doi.org/10.1002/9781119523048.ch1}.

Information is obtained by detecting and measuring changes in the surrounding field, whether it is an electromagnetic, acoustic, or potential field. In particular, the term “remote sensing” is most commonly used in connection with electromagnetic techniques of information acquisition, where the entire electromagnetic spectrum is considered, beginning with low-frequency radio waves and progressing to microwave, submillimeter, far infrared, near infrared, visible, ultraviolet, x-ray, and gamma-ray spectrum's regions \cite{doi:https://doi.org/10.1002/9781119523048.ch1}.

In this project, the main application of remote sensing is Earth observation, which is intended as "the practice of deriving information about the Earth’s land and water surfaces using images acquired from an overhead perspective, using electromagnetic radiation in one or more regions of the electromagnetic spectrum, reflected or emitted from the Earth’s surface" \cite{campbell2011introduction}.
In particular, we will rely on satellite optical acquisitions to extract information on possible marine anomalies in coastal areas. The use of satellites for Earth observation applications is detailed later in this work.

\subsection{Satellites and Sentinel-2}
Artificial satellites have enabled the acquisition of information about Earth in a way that was not previously possible. Particularly, satellite sensors can provide data on global cloud patterns, surface vegetation cover, surface morphological structures, ocean surface temperature, and near-surface wind. Satellites allow the monitoring of various events such as atmospheric phenomena. Moreover, the periodic monitoring of satellites can be leveraged to observe seasonal, annual, and longer-term changes such as ice cover reduction, desert expansion, and deforestation \cite{doi:https://doi.org/10.1002/9781119523048.ch1}.

\subsubsection{Satellite Characteristics}
The satellite orbit is the path that a satellite follows. Satellite orbits vary depending on their orientation and rotation with respect to Earth and their height above Earth's surface (i.e. their altitude). 

\begin{figure}[H]
    \centering
    \includegraphics[width=0.4\textwidth]{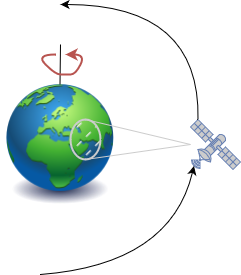}
    \caption{Illustration of a satellite with geostationary orbit. Adapted from \cite{satellite_characteristics}.}
    \label{Fig:sat_geostationary}
\end{figure}

Geostationary orbits have an altitude of approximately 36,000 kilometers. Satellites that are in a geostationary orbit revolve at a speed equal to the Earth's rotation speed, and they are called geostationary satellites (Figure \ref{Fig:sat_geostationary}). Their sensors always sense the same portion of the Earth. For instance, geostationary orbits are chosen for communications and weather satellites \cite{satellite_characteristics}.

\begin{figure}[H]
    \centering
    \includegraphics[width=0.3\textwidth]{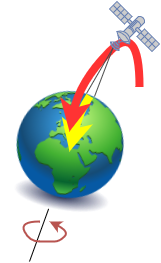}
    \caption{Illustration of a satellite with near-polar orbit. Adapted from \cite{satellite_characteristics}.}
    \label{Fig:sat_polar}
\end{figure}

Orbits that pass near the two poles are called near-polar orbits (Figure \ref{Fig:sat_polar}). Satellites with a sufficiently wide swath width that follow near-polar orbits are able to cover all of the Earth's surface in a certain time span. Indeed, the Earth rotates around itself (west-east) in the complementary direction of the satellite near-polar orbit (north-south). Often, near-polar orbits are also sun-synchronous, which means that a satellite in a sun-synchronous orbit will always pass over a certain surface at the same time of the day, which is called local sun time. So, at the same latitude in sun-synchronous orbits, the position of the sun with respect to the satellite will provide consistent lighting conditions during the same season. Having consistent illumination conditions over a season can be leveraged to better visualize changes over a period of days or over the same season but among different years \cite{satellite_characteristics} when optical sensors are used.

\begin{figure}[H]
    \centering
    \includegraphics[width=0.4\textwidth]{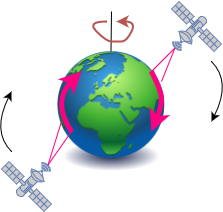}
    \caption{Illustration of ascending and descending passes of a satellite with near-polar orbit. Adapted from \cite{satellite_characteristics}.}
    \label{Fig:sat_asc_desc}
\end{figure}

Nowadays, the majority of low Earth orbit satellites used for remote sensing have near-polar orbits. Satellites in near-polar orbits make two passes at the same latitude per orbit: an ascending and a descending pass (Figure \ref{Fig:sat_asc_desc}). When the orbit is sun-synchronous and near-polar, the descending pass is usually performed on the sunny side of the Earth, whilst the ascending pass is done on the Earth's shadowed side. Therefore, satellites with sensors that can only acquire the visible spectrum are able to collect meaningful data only during the descending pass. Instead, satellites with sensors that can acquire also other portions of the spectrum, or acquire other non-visual information or use active sensors, such as Synthethic Aperture Radars (SARs),  are capable of collecting data in both passes \cite{satellite_characteristics}.

\begin{figure}[H]
    \centering
    \includegraphics[width=0.4\textwidth]{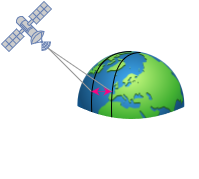}
    \caption{Illustration of the swath of a satellite. Adapted from \cite{satellite_characteristics}.}
    \label{Fig:sat_swath}
\end{figure}

Every remote sensing satellite has a swath (Figure \ref{Fig:sat_swath}), whose width determines the across-track portion of Earth that is captured by the sensor at a precise instant of time. Spaceborne sensors' swath widths generally have a range between tens and hundreds of kilometers. A satellite in a near-polar orbit is able to sense different parts of the Earth's surface after each revolution around the Earth since the Earth is rotating in a complementary direction (west-east) compared to near-polar satellites (north-south). So, with an adequate swath width, satellites are usually able to sense the entire Earth's surface thanks to this duality of Earth's rotation and satellite orbit \cite{satellite_characteristics}.

A satellite completes an orbit cycle when it passes again over the same point on the Earth's surface that is underneath it. The point below the satellite on the Earth's surface is called the nadir point, and the time to complete an orbit cycle differs depending on the satellite \cite{satellite_characteristics}. 

Another important measure is the revisit period, which can be different from the time needed for a satellite to sense over the same nadir point twice (orbit cycle time). For instance, the revisit time could be shorter compared to the orbit cycle time if steerable sensors are utilized. In this case, a satellite can acquire information about a certain  area of the Earth before and after the passage of its nadir point in that area. The revisit period is critical for a variety of remote sensing applications, particularly in cases that need multiple acquisitions of images (e.g. monitoring oil spills) \cite{satellite_characteristics}.

Lastly, satellites in near-polar orbits acquire information more frequently in areas at high latitudes compared to the areas close to the equator 
\cite{satellite_characteristics}.


\subsubsection{Sentinel-2}

Sentinel-2 is an Earth observation mission developed by the European Space Agency (\gls{ESA}) on behalf of the joint \gls{ESA}/European Commission initiative Copernicus \cite{esa_sentinel_overview}. The mission has the purpose of monitoring information for environment and security applications \cite{drusch2012sentinel}. Examples of such applications are urban and forest change detection.

The Copernicus Sentinel-2 (\gls{S2}) mission consists of a constellation of two polar-orbiting satellites that are phased at 180 degrees from one another and put in a sun-synchronous orbit at a mean altitude of 786 km. The two satellites are respectively called Sentinel-2A and Sentinel-2B. The satellites' swath width of 290 km and revisit time ("10 days at the equator with one satellite and 5 days with two satellites under cloud-free conditions, resulting in 2-3 days at mid-latitudes" \cite{esa_sentinel_2}) enable monitoring of changes to the Earth's surface.

Each Sentinel-2 satellite acquires multispectral images. In particular, 13 spectral bands are collected by the optical system mounted on each satellite, covering parts of the visible and near-infrared (VNIR) and short-wave-infrared (SWIR) sub-portions of the spectrum.
Four bands have a 10 m resolution, six bands a 20 m resolution, and three bands a 60 m resolution \cite{esa_sentinel_overview_2}. In particular, the bands having a 10 m resolution are bands B2, B3, B4, and B8. The bands with a 20 m resolution are bands B5, B6, B7, B8A, B11, and B12. Lastly, the bands with a 60 m resolution are bands B1, B9, and B10 \cite{esa_sentinel_2_bands}. 
The four bands with a resolution of 10 m respectively sense blue (490nm), green (560nm), red (665nm) and near infrared (842nm) sub-portions of the spectrum.
The six bands with a 20 m resolution include four narrow bands for vegetation characterization (705nm, 740nm, 783nm and 865nm), and two larger SWIR bands (1610nm and 2190nm) for applications such as snow/ice/cloud detection or vegetation moisture stress assessment.
Lastly, the three bands with a 60 m resolution are primarily used for cloud detection and atmospheric corrections (443nm for aerosols, 945 for water vapour and 1375nm for cirrus detection) \cite{sentinel-2-prod-levels}. Bands with 60 m spatial resolution have a higher frequency resolution and partially overlap the other bands in terms of spectral content.

\subsubsection{Product Levels}
\label{subsec:product-levels-s2}

The data acquired by the Sentinel-2 satellites undergoes several processing steps that transform it from the so called Level-0 to Level-2A. Specifically, the Sentinel-2 Product Levels are: Level-0, Level-1A, Level-1B, Level-1C, and Level-2A. Level-0 products consist in compressed raw images on board the satellite. Level-1A corresponds to decompressed raw images \cite{sentinel-2-prod-levels}. 

Level-1B products are multispectral image data that have been corrected for radiometric distortions, with spectral bands that have been coarsely co-registered with each other \cite{sentinel-2-prod-levels}.
Radiometric corrections are mathematical procedures that adjust for different causes of spectral distortion in the data \cite{levin_remote_2013}.
Image co-registration is the procedure of aligning two or more images geometrically \cite{huhdanpaa2014image}.

Level-1C products are orthorectified and provide Top-Of-Atmosphere (\gls{TOA}) reflectances \cite{sentinel-2-prod-levels}.
Orthorectification is the procedure of removing image distortions or displacements that arise due to sensor tilt and variations in topography. The goal of orthorectification is to represent each point on the image as if it were observed directly beneath the sensor (also known as nadir) \cite{orthorect-intro}.

Level-2A products are orthorectified and provide Bottom-Of-Atmosphere (\gls{BOA}) reflectances, along with a basic classification of pixels (which includes categories for various types of clouds) \cite{sentinel-2-prod-levels}.

Atmospheric correction is a crucial procedure in extracting land surface properties from satellite data. The surface reflectance signal that passive satellite devices measure is tainted by atmospheric effects. Phenomena such as the creation of thin cirrus clouds, and gas absorption prevent a clear view of the surface, resulting in a blurred image of radiation reflected by the atmosphere. Atmospheric correction is a method designed to eliminate the impact of these phenomena on the detected signal \cite{vermote2008atmospheric}. 

\gls{TOA} and \gls{BOA} reflectances can be interpreted as the reflectances that would be measured just above the Earth’s atmosphere and just above the land surface, respectively.




\subsection{Remote Sensing for Marine Anomalies Detection}

Some studies constructed spectral indices (the Floating Debris Index (FDI) \cite{biermann2020finding} and the Plastic Index (PI) \cite{themistocleous2020investigating}) to detect marine debris with multispectral satellite imagery. However, distinguishing marine debris from marine water and other floating objects (e.g. ships) is not straightforward \cite{kikaki2022marida}. It is significantly complex to build a spectral index that can generalize well across different types of marine litter. Furthermore, developing additional spectral indices to distinguish different marine anomalies adds to this complexity.

\subsection{Onboard Artificial Intelligence}
\label{subsec:onborad-ai}

The majority of \gls{EO} satellites acquire and store data internally and then send the data to ground stations when they enter their so-called coverage area. In particular, satellites transmit their data to ground after receiving commands from the ground stations. The described traditional approach is called the "bent-pipe" communication paradigm \cite{furano2020towards}. In this paradigm, no \gls{AI} algorithms are applied directly on-the-edge (i.e. by using satellite resources) but instead they are applied on the ground after the satellite sends data to them. Moreover, except for onboard compression and data-calibration, usually images are completely post-processed on ground stations.

It is important to note that a satellite may not collect data when it passes over certain areas because the downlink bandwidth that a satellite uses to send data to ground stations is limited. For this reason, generally Earth observation missions do not acquire data over oceanic areas unless they are specifically designed for this task. \cite{esa_sentinel_2_coverage} shows the coverage area of Sentinel-2 acquisitions. The coverage area of acquisitions of a satellite are the areas over which the satellite acquires data.

Using \gls{AI} models on board spacecraft, as opposed to downloading images and processing them on ground stations, could result in \first a reduction of the usage of the limited bandwidth downlink \cite{furano2020towards}, \second an extension of the acquisition area of the satellite, \third faster notifications of anomalies. 
More specifically, if the task is detecting marine anomalies, a satellite could send to ground stations only those images containing marine anomalies, filtering out all the others and significantly mitigate the bandwidth issue.
Moreover, having an \gls{AI} model onboard that could discard not relevant images increases the free storage space on the satellite. Therefore, the satellite could collect images outside the coverage area of acquisitions and expand it.
Finally, processing images with \gls{AI} on satellites could reduce the waiting time of marine anomalies notifications. In fact, if anomalies are detected onboard the spacecraft itself, notifications could be directly sent from satellites to ground stations.
This method contrasts with the traditional bent-pipe approach, which relies solely on \gls{AI} models on the ground and applies them to the data only after it has been transmitted by the satellites.


\section{Artificial Intelligence and Deep learning for Marine Anomalies}
\label{section:ai_dl_mad}

\subsection{Artificial Intelligence and Deep Learning}

\gls{AI} refers to the ability of a digital computer to carry out tasks that are usually associated with intelligent entities.

Deep learning (\gls{DL}) is a subfield of \gls{AI} that allows computational models to learn representations from data, with several levels of abstraction \cite{LeCun2015}.
Learning data representations is the crucial step that simplifies the process of extracting valuable insights when constructing classifiers \cite{bengio2013representation}.
The computational models used in deep learning are called \textit{Deep Neural Networks} (\glspl{DNN}), and they have multiple processing layers. During the last decade, there has been a growing interest in using deeper (more and more) layers in neural networks. This approach has been shown to outperform classical methods in various fields, particularly in pattern recognition \cite{albawi2017understanding}. Each layer is composed by one or more nodes that are called artificial neurons.

\subsubsection{Artificial Neurons}
\begin{figure}[H]
    \centering
    \includegraphics[width=0.5\textwidth]{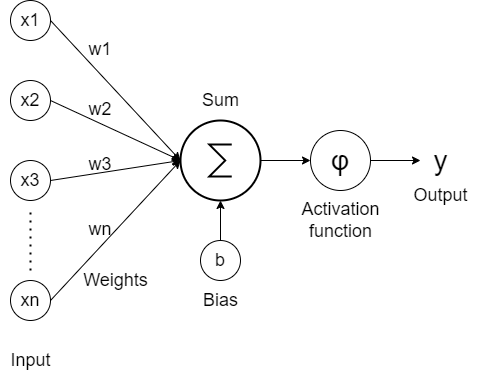}
    \caption{Illustration of an artificial neuron. Adapted from \url{https://commons.wikimedia.org/wiki/File:ArtificialNeuronModel.png} (\href{https://creativecommons.org/licenses/by-sa/3.0/deed.en}{CC BY-SA 3.0 Deed license}).}
    \label{Fig:artificial_neuron}
\end{figure}

Figure \ref{Fig:artificial_neuron} shows an illustration of an artificial neuron. An artificial neuron consists of three sets of operations: summation, multiplication, and activation \cite{krenker2011introduction}. It first takes $n$ input data ($x_1, \ldots, x_n$) and $n$ weight coefficients ($w_1, \ldots, w_n$). It then sums all the inputs, which are multiplied by their corresponding weight and summed with their bias (i.e. $\sum_{i=1}^n x_i w_i + b_i$). Afterward, it applies an activation function $\varphi$ to the previously computed sum. The result of the activation function is called the output of the neuron, and it is usually denoted as $y$.

\subsubsection{Artificial Neural Networks}
Artificial neural networks (\glspl{ANN}) \cite{mcculloch1943logical} are networks of artificial neurons communicating with each other. Each neuron is connected to at least one neuron, and each connection between two neurons has a weight coefficient. A weight coefficient is a real number that corresponds to a measure of the importance of the connection itself \cite{svozil1997introduction}. The neurons of a neural network are divided into layers. There exist three types of layers: input, hidden, and output layers. 

\begin{figure}[H]
    \centering
    \includegraphics[width=0.5\textwidth]{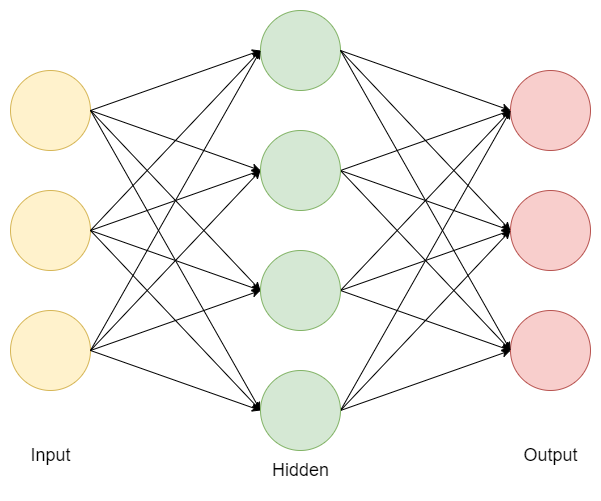}
    \caption{Example of a feed-forward artificial neural network with one input layer, one hidden layer, and one output layer. The input, hidden, and output layers have respectively three, four, and three neurons.}
    \label{Figure:ann}
\end{figure}

The simplest form of the neural network is the \textit{Feed Forward} Neural Network, which is an artificial neural network in which the connections between nodes do not form a cycle. Figure \ref{Figure:ann} shows an example of a feed-forward neural network. The input layer is the first layer, the hidden layers are in between, and the output layer is the last one. A neural network is considered \textit{deep} when it has multiple hidden layers.

\subsubsection{Convolutional Neural Networks}
\label{subsubsec:cnns}

Convolutional Neural Networks (\glspl{CNN}) are artificial neural networks that are mostly used for pattern recognition within images. In particular, they incorporate image-specific knowledge into the architecture, making the network more suited for image-focused tasks while reducing the number of parameters needed to set up the model \cite{o2015introduction}. 

The name of the network comes from the \textit{convolution}, which is a mathematical operator, even though the real operation a \gls{CNN} performs is called \textit{cross-correlation} (Eq. \ref{eq:cross-correlation}). In a 2D convolutional layer, a kernel $K$ of size $(A, B)$ slides over the data $I$ (input data or an intermediate feature map) of size $W \times H$. At a given position $(i, j)\text{, with } 1 \le i \le W \text{, } 1 \le j \le H$ the convolutional layer performs a cross-correlation (Eq. \ref{eq:cross-correlation}) between the kernel and the data, which consists of the dot product of the kernel $K$ and the overlapping portion of the data $I$.

\begin{equation}
    K \star I (i, j) = \sum_{x=1}^{A} \sum_{y=1}^{B} K(x, y) I(i+x, j+y) \cdot 
\label{eq:cross-correlation}
\end{equation}

\begin{figure}[H]
    \centering
    \includegraphics[width=1.0\textwidth]{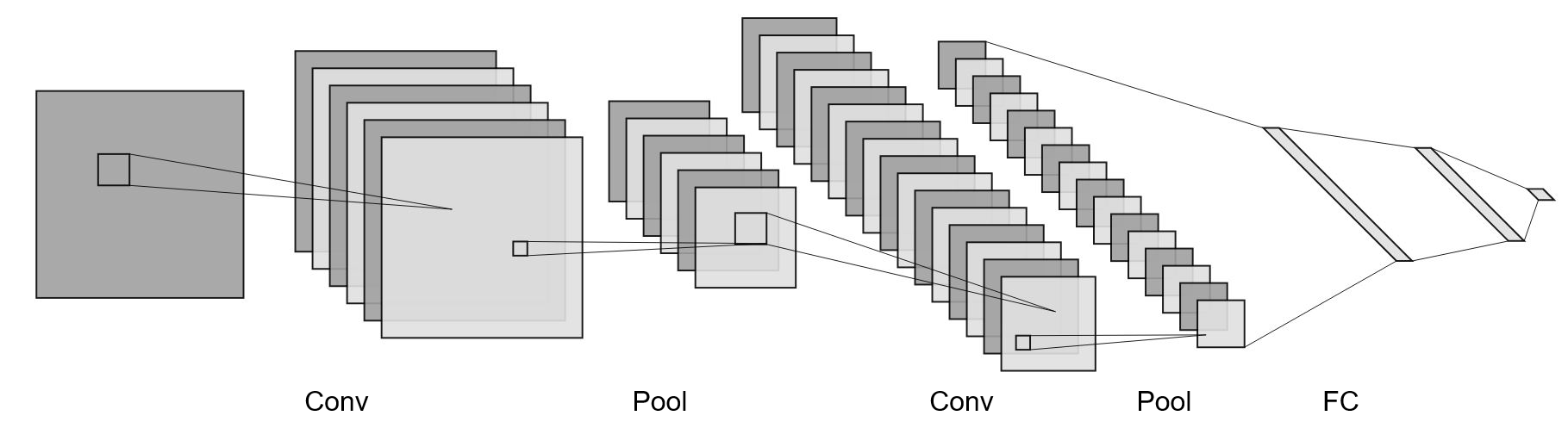}
    \caption{Example of a convolutional neural network with two convolutional layers (Conv), two pooling layers (Pool), and three fully connected layers (FC). Adapted from Figure 2 (a) of \cite{gu2018recent}.}
    \label{Figure:cnn}
\end{figure}

Figure \ref{Figure:cnn} shows an example of the architecture of a convolutional neural network, which consists of three types of layers: convolutional, pooling, and fully-connected layers. In particular, the objective of the convolutional layer is to learn feature representations of the inputs \cite{gu2018recent}, while pooling layers reduce the dimensions of the feature maps to decrease computational costs, and fully-connected layers are used to make a prediction based on the learnt features.

\subsection{Semantic Segmentation}

"The task of semantic segmentation is to segment the input image according to semantic information and predict the semantic category (also called class) of each pixel from a given label set" \cite{mo2022review}. Image segmentation is an active research field applied to various applications, some examples are automated disease detection and self-driving cars \cite{jadon2020survey}. In addition, identifying and categorizing objects within an image is a difficult task in the field of computer vision. However, the use of deep learning techniques has significantly improved the accuracy of semantic segmentation in recent years \cite{hao2020brief}.

\begin{figure}[H]
    \centering
    \includegraphics[width=\textwidth]{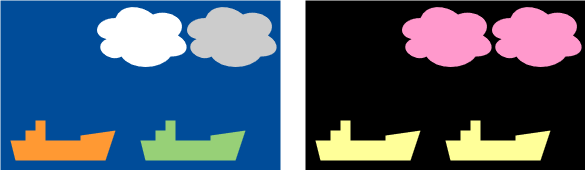} 
    \caption{Example of an image and its semantic segmentation map.}
    \label{Figure:sem_seg_example}
\end{figure}

So, given an input image of size $H$x$W$, the task of semantic segmentation consists of computing a segmentation map (i.e. an image), with the same spatial dimensions of the input image, that has a class (a color) assigned to each of its pixels. Different classes have different colors.
An example is shown in Figure \ref{Figure:sem_seg_example}. On the left is the input image and on the right is its ground truth segmentation map. The illustrated segmentation map has 3 classes: cloud, ship, and background, which correspond to pink, yellow, and black. 

A semantic segmentation map usually has 1 or $C$ channels, where $C$ is the number of classes. In the first case, each pixel's value of the unique channel is an integer number ranging from 0 to 255 due to the encoding of gray-scale images. Therefore, there is a limit of a maximum of 256 classes (255 classes and 1 background class). 
In the latter case, a segmentation map has one channel for each category, and the pixels of the $c$-th channel (with $c=1, \ldots C$) have a value that is $0$ or $1$ corresponding respectively to the background and the $c$-th semantic class.


In this work, we model the problem of detecting marine anomalies as a semantic segmentation task where the number of classes corresponds to the number of anomalies plus the background class. The background class is marine water.

\subsubsection{Fully Convolutional Networks}

Although other methods were popular in the past, deep learning has outperformed them in terms of accuracy and by a significant margin in many computer vision problems, including semantic segmentation \cite{garcia2017review}.

One popular type of neural networks used for semantic segmentation are the so called Fully Convolutional Networks (\gls{FCN}). Indeed, \glspl{FCN} can efficiently process input of any size and generate output of a corresponding size \cite{long2015fully}, which is very convenient for a pixel-wise prediction task as semantic segmentation.

\glspl{FCN} are \textit{fully convolutional} because they only utilize convolution, pooling, and upsampling layers. Hence, they do not use any fully connected layer, leading to a reduction in the number of parameters and faster training times. \glspl{FCN} are a subclass of \glspl{CNN}. The primary distinction between a \gls{FCN} and a \gls{CNN} is that the latter also uses fully connected layers.

A \gls{FCN} is composed of two sequential paths: a downsampling path, which extracts features, and an upsampling path, which enables localization (i.e. identifying the location of specific features or classes within the image). 

\glspl{FCN} use skip connections to recover the detailed spatial information lost during the downsampling process.

\subsubsection{U-Net}

U-Net \cite{ronneberger2015unet} is an extension of the \gls{FCN} architecture that expands the idea of skip connections used in \glspl{FCN} by incorporating an approximately symmetric encoder-decoder structure. The U-Net provides more precise segmentations compared to \glspl{FCN} and is able to work with a limited number of training images \cite{ronneberger2015unet}, which makes it as a suitable neural network for the task of semantically segment marine anomalies.

The encoder (downsampling path) is composed of convolutional and max pooling layers, while the decoder (upsampling path) has convolutional and upsample layers. 

The skip connections of the U-Net architecture concatenate (instead of summing as in \glspl{FCN}) feature maps of the upsampling path with the correspondingly cropped feature maps from the contracting path \cite{ronneberger2015unet}.

\subsubsection{State of the art of Deep Learning Methods for Marine Anomalies Detection}

In their work, Kikkaki et al. \cite{kikaki2022marida} employed machine and deep learning methods to semantically segment marine debris and other anomalies on the MARIDA dataset. In particular, the authors utilized three distinct random forest models and a U-Net. The first model was based solely on the spectral signatures of each pixel. The second model incorporated both spectral signatures and calculated spectral indices. The third model included spectral signatures, spectral indices, and textural features extracted from the Gray-Level Co-occurrence Matrix. 

The best results were achieved by the latter random forest model, which performed even better than the U-Net. However, the U-Net was trained for only 44 epochs. We think that training for more epochs, increasing the hyperparameter space, and employing other types of loss functions could have increased significantly the results of the fully convolutional neural network.

Mifdal et al. \cite{mifdal2021towards} studied the use of deep learning methods to automatically detect (via semantic segmentation) floating objects from a hand-labeled dataset without exploiting manually defined spectral indices. Their experiments show that employing a U-Net \cite{ronneberger2015unet} model to the task leads to better performance compared to traditional machine learning models that use manually-crafted spectral indices.

\subsubsection{Weakly Semantic Segmentation}
\label{sec:weakly-sem-seg}

Unfortunately, training supervised deep neural networks for semantic segmentation tasks can be challenging due to the need for a large amount of high-quality pixel-level annotations for high-resolution images. Producing these annotations is both labor-intensive and time-consuming. It is time-consuming due to the need for fine-grained labeling at the pixel level. Additionally, this process may even require the expertise of specialist annotators to correctly identify pixels \cite{hua2021semantic}.

Indeed, expert annotators are needed to annotate multispectral satellite imagery for semantic segmentation of marine anomalies. It is in fact difficult to clearly identify anomalies from satellite images. To properly label data, experts may need to collect reports regarding anomalies events before annotating images. Moreover, experts could also necessitate to visualize computed spectral indices to highlight and distinguish the different marine anomalies in the image.

Sometimes not all the pixels of an image are labeled since \first labeling all of them is a time-consuming and expensive task, \second it can be difficult to recognize and properly annotate each pixel. The task of predicting the categories of pixels when having not fully-annotated images is called weakly-semantic segmentation.

%

\subsection{Semi-Supervised Learning}
Semi-supervised learning (\gls{SSL}) is the subfield of machine learning that focuses on performing specific learning tasks using both labeled and unlabeled data \cite{van2020survey}, whilst standard supervised learning uses only labeled data.

Semi-supervised learning usually performs well when \first labeled data are scarce compared to unlabeled data, \second and the labeled data is not enough to obtain state-of-the-art results on the task that is tackled. The latter implies that if a task can be easily solved just by using labeled data, then unlabeled data will probably not provide great improvements.  

Utilizing unlabeled data could be very beneficial because \first annotating data is time-consuming and costly (e.g. hiring annotators) \cite{oliver2018realistic}, \second expert knowledge is often needed, \third learning with unlabeled data could help improve the model's generalization \cite{patel2021evaluating}, and \fourth there are many real-world applications that have a lack of labeled but not of unlabeled data that could benefit from semi-supervised learning. The latter point is particularly true for Earth observation data, for which labeling requires expert annotators, whilst unlabeled data are acquired every day by Earth observation satellites.

In this work, we use semi-supervised learning to study how deep learning models can perform when trained with a small set of labeled and a larger set of unlabeled multispectral satellite data.

\subsubsection{Pseudo-Labeling}
\label{subsec:pseudo-labeling}
Pseudo-labeling is a semi-supervised learning technique that leverages a model to generate labels for unlabeled data \cite{sohn2020fixmatch}.  

In particular, pseudo-labeling is closely related to Self-Training \cite{zhu2005semi}. A simple implementation of pseudo-labeling based on self-training is thus described. Firstly, a network is trained to minimize a loss function $\mathcal{L}$ with the small amount of labeled data $D^l$. The updated parameters of the network are denoted by $\theta^*$. Secondly, the trained model predicts a pseudo-label for each unlabeled sample $x_i^u$. If the prediction of an unlabeled sample is highly confident (e.g. if its probability is higher than a threshold $\tau$), then the sample is added to the dataset with its pseudo-label generated by the model. Subsequently, the parameters $\theta^*$ are updated by training with both labeled data and pseudo-labeled data. This process is repeated for a number of $m$ steps, where $m$ can be a pre-determined number, a hyperparameter, or a pre-defined condition. The pseudo-code can be seen in Algorithm \ref{alg:pseudo-labeling}. 

Different measures could be used to estimate the confidence of pseudo-labels. A simple measure is the softmax function. For instance, a pseudo-label could be defined as the class that is predicted with the highest probability \cite{lee2013pseudo}.

\begin{algorithm}[hbt!]
\caption{Example of a simple realization of the Pseudo-Labeling algorithm.}\label{alg:pseudo-labeling}
$D = D^l$\\
\For{$i = 1:m$}{
  $\theta^* = \text{argmin}_{\theta}  \mathcal{L}(D)$\\
  $D_{conf}^{u} = \{\}$\\
  \For{$j = 1:n^u$}{
      \If{$conf(f(x_j^u)) > \tau$}{
        $D_{conf}^{u} = D_{conf}^{u} \cup x_j^u$\\
      }
  }
  $D = D^l \cup D_{conf}^u$\\
  \Return $\theta^*$
}
\end{algorithm}

A downside of pseudo-labeling is that it can suffer from the confirmation bias \cite{arazo2020pseudo}. The so-called confirmation bias happens when the network overfits to incorrect pseudo-labels generated by itself. Pseudo-labels can be incorrect because sometimes network outputs are wrong. Moreover, the network outputs will further deteriorate if incorrect outputs are utilized as pseudo-labels for unlabeled data.

\subsubsection{Consistency Regularization}
\label{subsec:consistency-regularization}
Consistency regularization is a frequently employed method for semi-supervised learning. It is an auxiliary objective function \cite{englesson2021consistency} whose aim is to have similar outputs when the network is fed with a perturbed sample and its corresponding original sample. So, in computer vision applications, when given perturbed versions of the same image, consistency regularization enforces that the model should produce similar outputs \cite{sohn2020fixmatch}. 

A perturbed sample is a sample that had been modified by a perturbation function. A perturbation function could for instance be a simple translation if the input data are 2D points, or an image augmentation if the input data are images. There exist other methods to perturb an image too. Indeed, an image can be perturbed by stochastic model perturbations too. For instance, the $\Pi$-model \cite{laine2016temporal} enforces the similarity of network outputs between two versions of the same input sample that are obtained through both dropout regularization (model perturbation) and image augmentation (data augmentation).

A general definition of consistency regularization is presented as follows. Suppose to have an unlabeled sample $x_i$, its perturbed version $\tilde{x}_i$, a neural network defined as $f_{\theta}()$ where $\theta$ are its parameters, a threshold $\tau$, and a distance function $d$. A distance function is chosen to express the concept of similarity, but any other function that can compute a form of similarity between two examples could be utilized. Given these premises, consistency regularization wants the distance between the outputs of the network on the perturbed and on the original sample to be lower than a threshold $\tau$ (i.e. it wants the outputs of perturbed and original samples to be similar):

\begin{equation}
    d(f_{\theta}(x_i), f_{\theta}( \tilde{x}_i )) < \tau
\end{equation}

So, the objective function can be defined as:
\begin{equation}
    \mathcal{L}^u = \sum_{i=1}^n d(f_{\theta}(x_i), f_{\theta}( \tilde{x}_i ))
\end{equation}
where $n$ is the number of samples.

In semi-supervised methods that use consistency regularization, the loss function usually consists of two components. The first component is the standard supervised loss ($\mathcal{L}^s$), evaluated for labeled inputs only. The second component ($\mathcal{L}^u$), evaluated for unlabeled inputs (or both unlabeled and labeled inputs), penalizes different predictions for the same training input $x_i$ by computing the distance between the predictions on the original and the perturbed sample. So, the loss is defined as: 
\begin{equation}
    \mathcal{L} = \mathcal{L}^s + \lambda \mathcal{L}^u
    \label{eq:consist-regulariz-loss}
\end{equation}
where $\lambda$ is a hyperparameter that weighs the contribution of the unsupervised loss function.

\subsubsection{Semi-Supervised Learning for Earth Observation}

This section outlines related work in the field of semi-supervised learning applied to Earth observation data.

Patel et al. \cite{patel2021evaluating} applied self- and semi-supervised state-of-the-art methods on three remote sensing tasks to study their performance with limited labeled data and geographical domain shifts. In particular, the work investigated the use of SimCLR \cite{chen2020simple} as the self-supervised method and FixMatch \cite{sohn2020fixmatch} as the semi-supervised one. The results highlighted that when there is enough annotated data, then the self- and semi-supervised methods do not provide substantial improvements compared to the fully-supervised setting. However, self- and semi-supervised methods (also in combination) can provide notable benefits in terms of performance when labeled data is scarce.

Wang et al. \cite{wang2020semi} also demonstrated the effectiveness of a semi-supervised method on three remote sensing datasets. Their method uses consistency regularization during training. Afterwards, the trained model is utilized for average update of pseudo-label. Finally, pseudo-labels and strong-labels are combined to train a neural network for semantic segmentation. The effectiveness of their proposed method was demonstrated on three remote sensing datasets, where it achieved superior performance without the need for additional labeled data.

Gómez et al. \cite{gomez2021msmatch} presented MSMatch, a semi-supervised learning method to tackle scene classification on RGB and multispectral images. The findings from their study showed that the semi-supervised approach they developed has a superior performance compared to the supervised method.



\cleardoublepage

\chapter{Methods}
\label{ch:methods}



The purpose of this chapter is to provide an overview of the research method
used in this thesis. Section~\ref{sec:meth_dataset} focuses on the data used in this work. Various losses applicable to semantic segmentation tasks are explored in Section~\ref{subsec:losses_sem_seg}, with a justification for our chosen loss. The neural network architecture we used is detailed in Section~\ref{sec:meth_model_arch}. Section~\ref{sec:meth_fixmatch_sem_seg} outlines the utilized semi-supervised approach, which will be called FixMatch for Semantic Segmentation.
Section~\ref{sec:meth_eval_metrics} describes the selected method to evaluate the deep-learning models.
Finally, Section~\ref{sec:meth_exp_design} describes the experimental design.

\section{Dataset}
\label{sec:meth_dataset}

This section contains all information about the data that was used in our work. Subsection \ref{subsec:marida} provides details about the MARIDA dataset \cite{kikaki2022marida}, which is weakly-labeled. It further explains the modifications we made to this dataset to address issues of class imbalance and to enhance its relevance for marine anomaly cases.

\subsection{MARIDA}
\label{subsec:marida}
\begin{figure}[H]
    \centering
    \begin{subfigure}[t]{\textwidth}
    \centering
        \raisebox{-\height}{\includegraphics[width=0.20\textwidth]{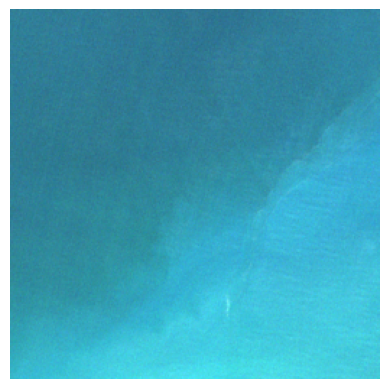}}
        \raisebox{-\height}{\includegraphics[width=0.20\textwidth]{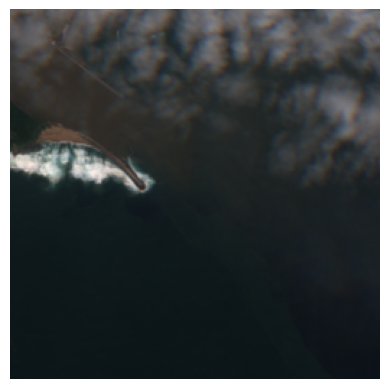}}
        \raisebox{-\height}{\includegraphics[width=0.20\textwidth]{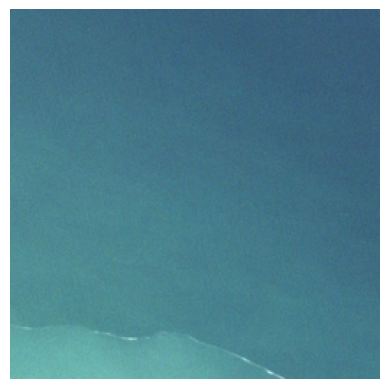}}
        \raisebox{-\height}{\includegraphics[width=0.20\textwidth]{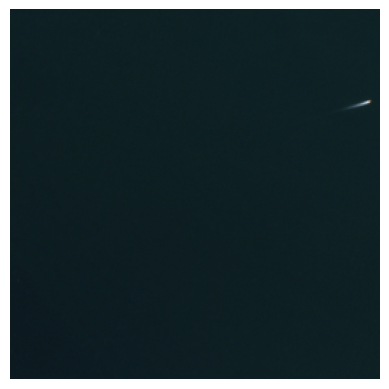}}
        \raisebox{-\height}{\includegraphics[width=0.20\textwidth]{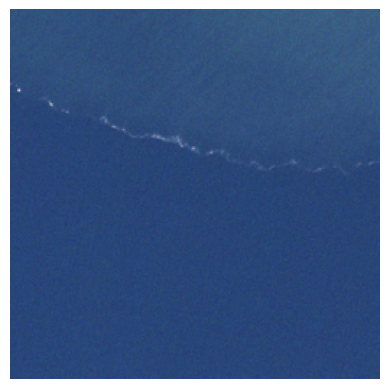}}
        \raisebox{-\height}{\includegraphics[width=0.20\textwidth]{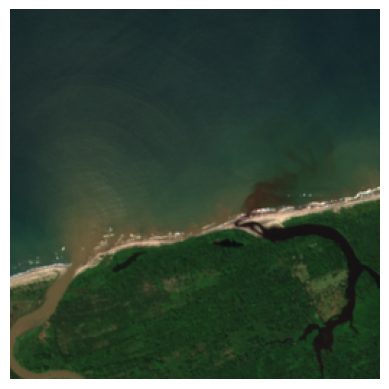}}
        \raisebox{-\height}{\includegraphics[width=0.20\textwidth]{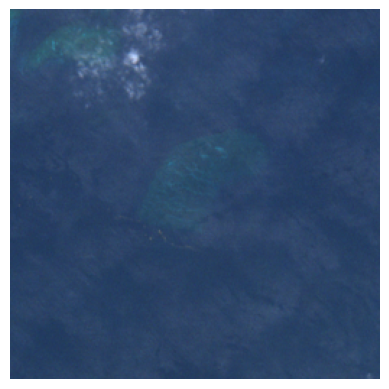}}
        \raisebox{-\height}{\includegraphics[width=0.20\textwidth]{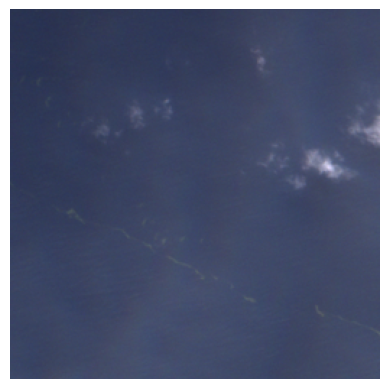}}
        \raisebox{-\height}{\includegraphics[width=0.20\textwidth]{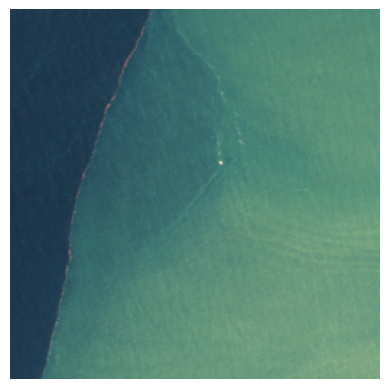}}
        \raisebox{-\height}{\includegraphics[width=0.20\textwidth]{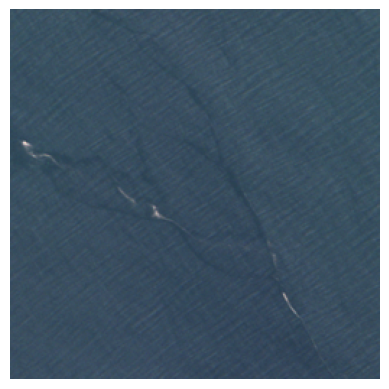}}
        \raisebox{-\height}{\includegraphics[width=0.20\textwidth]{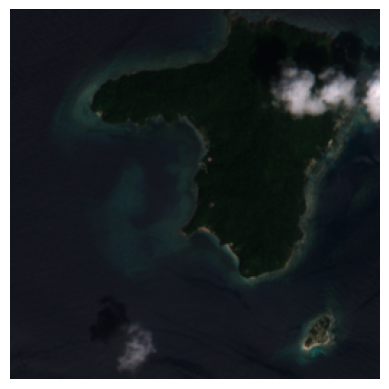}}
        \raisebox{-\height}{\includegraphics[width=0.20\textwidth]{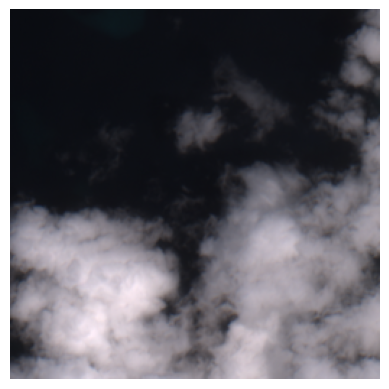}}
    \end{subfigure}
    \hfill
    \caption{Examples of MARIDA patches. The figure shows the RGB version of the patches. Therefore, only bands B4, B3, and B2 are shown for each patch.}
    \label{Figure:marida-rgb-images-samples}
\end{figure}

Marine Debris Archive (MARIDA) is a marine-debris dataset of multispectral Sentinel-2 satellite images \cite{kikaki2022marida}. In the context of the MARIDA dataset, the term \textit{patch} indicates an image of shape $256 \times 256$ with 11 channels corresponding to the 11 bands. Specifically, MARIDA patches contain 11 out of 13 bands of Sentinel-2 images. Indeed, the authors of \cite{kikaki2022marida} applied an atmospheric correction to the patches that originally were Level-1C products (refer to Subsubsection \ref{subsec:product-levels-s2}) with 13 bands. Therefore, bands B9 (Vapour) and B10 (Cirrus) were excluded. Figure \ref{Figure:marida-rgb-images-samples} shows some examples of MARIDA patches. 

\begin{figure}[H]
    \centering
    \begin{minipage}{0.45\textwidth}
        \centering
        \includegraphics[width=0.9\textwidth]{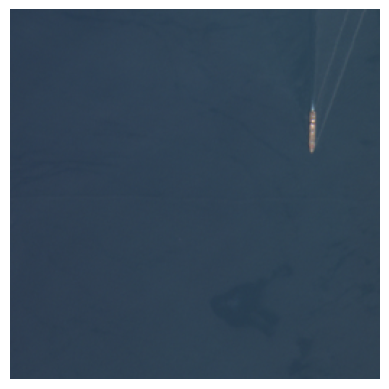} 
    \end{minipage}\hfill
    \begin{minipage}{0.45\textwidth}
        \centering
        \includegraphics[width=0.9\textwidth]{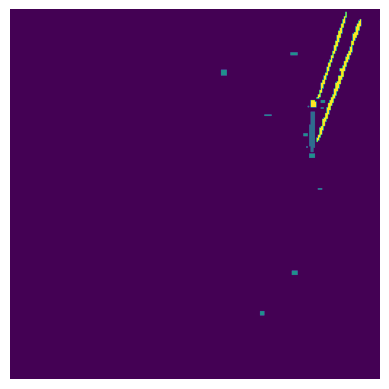} 
    \end{minipage}
    \caption{A MARIDA patch and its corresponding semantic segmentation map. The purple pixels of the semantic segmentation map are unlabeled.}
    \label{Figure:weakly-labeled-example}
\end{figure}

The MARIDA images are weakly-labeled (refer to Subsubsection \ref{sec:weakly-sem-seg}), meaning that not every pixel of an image is assigned a category. An example of MARIDA patch and its corresponding weakly-labeled semantic segmentation map is shown in Figure \ref{Figure:weakly-labeled-example}. It can be noticed that the majority of pixels of the semantic segmentation map are unlabeled (purple pixels).

The MARIDA dataset has a total of 15 classes, which are: \textit{Marine Debris}, \textit{Dense Sargassum}, \textit{Sparse Sargassum}, \textit{Natural Organic Material}, \textit{Ship}, \textit{Clouds}, \textit{Marine Water}, \textit{Sediment-Laden Water}, \textit{Foam}, \textit{Turbid Water}, \textit{Shallow Water}, \textit{Waves}, \textit{Cloud Shadows}, \textit{Wakes}, and \textit{Mixed Water}.

The original dataset contains 837,377 labeled pixels in total and its splitting scheme for labeled pixels of training/validation/test sets is approximately 50/25/25\% (it is 51.28/25.45/23.27\%). In particular, the training, validation and test sets contain respectively 694 patches (429,412 labeled pixels), 328 patches (213,102 labeled pixels) and 359 patches (194,843 labeled pixels). The annotation procedure, the distribution of classes, and additional details about the MARIDA dataset are described in \cite{kikaki2022marida}.

Some classes of the MARIDA dataset contain a very low number of labeled pixels (e.g. the classes \textit{Wakes} and \textit{Natural Organic Material} respectively have 410 and 864 labeled pixels in the whole dataset).

\subsubsection{Class Imbalance}

Class imbalance is a problem that arises when one class has more samples than the others. Generally, algorithms tend to focus on classifying the majority samples, often at the expense of ignoring or mis-classifying the minority samples. These minority samples, although infrequent, can be of significant importance \cite{longadge2013class}, especially in the detection of marine anomalies.

Performing semantic segmentation on satellite imagery to detect marine anomalies is a highly imbalanced task since most of the pixels in one satellite image would be marine water pixels, or also land if it is an image of a coastal area. Moreover, most satellite images collected over the oceans and seas show no anomaly, only seawater. 
Therefore, there are very few pixels with marine anomalies.

\subsubsection{Grouping of classes}

Some of the classes of the MARIDA dataset \first have a very limited number of labeled pixels and/or \second are not anomalies (or not anomalies of interest to our work). For these two reasons, we decided to group together some of the classes, leading to a total of 5 classes.


In particular, the categories related to water (\textit{Marine Water}, \textit{Sediment-Laden Water}, \textit{Foam}, \textit{Turbid Water}, \textit{Shallow Water}, \textit{Waves}, \textit{Cloud Shadows}, \textit{Wakes}, \textit{Mixed Water}) were combined together in the \textit{Water} class. \textit{Dense Sargassum}, \textit{Sparse Sargassum}, and \textit{Natural Organic Material} were grouped into the class \textit{Algae/Organic Material}. Lastly, the classes \textit{Cloud}, \textit{Marine Debris}, and \textit{Ship} were kept as separate classes.

Table 
\ref{table:multi-setting-pixels} 
shows the number and percentage of pixels of the new classes. The class with the least number of labeled pixels is now \textit{Marine Debris}
, with 3,399 pixels. The 
new setting is still very imbalanced. Indeed, some classes have less than 1\% labeled pixels of the total labeled pixels of the dataset. The focal loss is used to deal with class imbalance, and is later described (see Subsection \ref{subsubsec:focal-loss}). 



\begin{table}[H]
\caption{Number and percentage of pixels of classes of the new setting with 5 classes.}
\label{table:multi-setting-pixels}
\centering
\begin{adjustbox}{width=\columnwidth,center}
\begin{tabular}{lll}
    \toprule
    Class          & Percentage of labeled pixels (\%) & \multicolumn{1}{c}{Number of labeled pixels} \\
    \midrule
    Marine Debris          & \multicolumn{1}{r}{0.41}                         & \multicolumn{1}{r}{3,399}                                         \\
    Algae/Organic Material & \multicolumn{1}{r}{0.72}                         & \multicolumn{1}{r}{6,018}                                         \\
    Ship                   & \multicolumn{1}{r}{0.69}                           & \multicolumn{1}{r}{5,803}                                         \\
    Cloud                  & \multicolumn{1}{r}{14.02}                          & \multicolumn{1}{r}{117,400}                                       \\
    Water                  & \multicolumn{1}{r}{84.16}                         & \multicolumn{1}{r}{704,757}                                       \\
    \bottomrule
\end{tabular}
\end{adjustbox}
\end{table}

\subsubsection{Onboard Usage}
\label{subsubsec:onboard_marida}

Considering the deployment of an \gls{AI} model on board a satellite for the purpose of semantically segmenting marine anomalies, one potential method would be to send to the ground stations only the pixels predicted as anomalies, along with their georeferenced coordinates. This approach could substantially reduce the utilization of downlink bandwidth.

Nonetheless, certain aspects of the data warrant consideration. The MARIDA patches could be compared to Level-2A products, given that they are Level-1C products that have undergone atmospheric correction \cite{kikaki2022marida}. As a result, the image distortion typically caused by atmospheric effects has been compensated in the MARIDA dataset. Moreover, the bands within each patch are not only co-registered but also georeferenced. This means that if a satellite detects an anomaly, it has the capability to transmit the georeferenced coordinates of that anomaly to ground stations. However, it is important to note that these are strong assumptions, as typically, artificial satellites do not have atmospherically corrected, co-registered and georeferenced patches at their disposal on board.

\subsubsection{Patches that contain NaN values}
\label{subsubsec:nan_patches}
In the MARIDA dataset, a total of seven patches contain \gls{NaN} values. In particular, there is one patch with \gls{NaN} values in the training set (\textit{21-2-17\_16PCC\_0}), three in the validation set (\textit{18-9-20\_16PCC\_47, 18-9-20\_16PCC\_48, 18-9-20\_16PCC\_50}), and three in the test set (\textit{30-8-18\_16PCC\_0, 30-8-18\_16PCC\_1, 30-8-18\_16PCC\_2}). The mentioned patches containing \gls{NaN} values are not considered in our experiments.

\section{Losses}
\label{subsec:losses_sem_seg}

There exist various losses to train a neural network for the task of semantic segmentation.
In semantic segmentation, the ground truth label of each pixel can be represented as a discrete probability distribution having $C$ events. The $i$-th ($i=1, \ldots, C$) event corresponds to the probability of the pixel of being of the $i$-th class. In particular, the event corresponding to its true class has a probability of $1$ while all the other events have a probability equal to $0$.

This section describes several loss functions that are commonly used in semantic segmentation tasks and explain the reasoning behind our selection of the focal loss function.

\subsection{Cross Entropy}
\label{subsec:cross-entropy-loss}
Cross entropy is a measure of the difference between two probability distributions for a given random variable or set of events. The cross entropy between two discrete distributions $p$ and $t$ that both have $C$ events is defined as:
\begin{equation}
    CE = -\sum_{i=1}^{C} t_i \cdot log(p_i)
\label{eq:ce}
\end{equation}

where $t_i$ is the true probability distribution of a pixel and $p_i$ is the predicted probability distribution of the same pixel. 
In this way, when the two probability distributions $t_i$ and $p_i$ perfectly correspond (i.e. they both have a probability of $1$ assigned to the event corresponding to the true class and a probability of $0$ to all the other events) the cross entropy loss is zero, while it is greater than zero when they differ. And the more the two probability distributions differ the greater the loss will be.

Cross entropy is called \textit{Binary} Cross Entropy if the classification problem has only two classes ($C=2$). Equation \ref{eq:ce} can be rewritten as:
\begin{equation}
\begin{aligned}
    BCE &= -\sum_{i=1}^{2} t_i \cdot \log(p_i)\\
    &= -\left[t_1 \cdot \log(p_1) + t_2 \cdot \log(p_2) \right] \\
    &= - \left[ y \cdot \log(\hat{y}) + (1 - y) \cdot \log(1 - \hat{y}) \right]
\label{eq:bce}
\end{aligned}
\end{equation}
where $y$ is the probability of class 1, $1 - y$ is the probability of class 2, $\hat{y}$ is the predicted probability for class 1, and $1 - \hat{y}$ is the predicted probability for class 2. Equation \ref{eq:bce} can be used to compute the binary cross entropy of a single pixel.


Using the notation in \cite{lin2017focal}, the binary cross entropy can be re-defined as follows:
\begin{equation}
BCE(\hat{y}, y) =
\begin{cases}
     -\log(\hat{y}) & \text{if } y = 1,\\
     -\log(1 - \hat{y}) &   \text{otherwise.}
\end{cases}
\label{eq:bce_rewritten}
\end{equation}
In Equation \ref{eq:bce_rewritten}, $y \in \{0, 1 \}$ specifies the ground-truth class and $\hat{y} \in \left[0, 1\right]$ is the model’s estimated probability for the class
with label $y = 1$. $\hat{y_t}$ is then denoted as:
\begin{equation}
\hat{y}_t =
\begin{cases}
     \hat{y} & \text{if } y = 1,\\
     1 - \hat{y} & \text{otherwise.}
\end{cases}
\end{equation}
Thus, we can define \gls{BCE} as:
\begin{equation}
BCE(\hat{y}, y) = BCE(\hat{y}_t) = - \log(\hat{y_t})
\label{eq:bce_re_deifined}
\end{equation}

\subsection{Balanced Cross Entropy}

Balanced cross entropy is a variation of binary cross entropy that addresses class imbalance. It introduces a weighting factor $\alpha \in \left[0,1\right]$ for class 1 and $1 - \alpha$ for class -1. Starting from Equation \ref{eq:bce_re_deifined}, it is defined as:
\begin{equation}
BBCE(\hat{y}_t) = - \alpha_t \log(\hat{y_t})
\label{eq:bbce}
\end{equation}
where:
\begin{equation}
\hat{\alpha}_t =
\begin{cases}
     \alpha & \text{if } y = 1,\\
     1 - \alpha & \text{otherwise.}
\end{cases}
\end{equation}

In practice $\alpha$ may be set by inverse class frequency or treated as a hyperparameter \cite{lin2017focal}.

\subsection{Focal Loss}
\label{subsubsec:focal-loss}

The \textit{Focal} loss \cite{lin2017focal} works well in highly imbalanced class scenarios. In particular, it down-weights the contribution of easy examples and enables the model to focus more on learning hard examples \cite{jadon2020survey}. 

Indeed, when the focal loss is not used, easily classified negatives (e.g. background pixels correctly classified with a high probability) constitute the majority of the loss and monopolize the gradient. The factor $\alpha_t$ of Equation \ref{eq:bbce} assigns different importance to positive and negative samples, but it is not able to distinguish between easy and hard samples \cite{lin2017focal}.

Hence, the focal loss brings an advantage over the balanced cross entropy and drastically reduces the weights of easy samples to place more emphasis on hard negatives by using a modulating factor $(1 - \hat{y}_t)^{\gamma}$. 

The mathematical definition of the focal loss is:
\begin{equation}
\begin{aligned}
    FL(\hat{y_t}) = -\alpha_t(1 - \hat{y}_t)^{\gamma}\log(\hat{y}_t)
\label{eq:focal_loss}
\end{aligned}
\end{equation}
where $\gamma >= 0$. When $\gamma = 0$, \gls{FL} is equivalent to \gls{BBCE}. The factor $\alpha_t$ weights the loss of a sample based on its class, as it was for \gls{BBCE}. 

An example is now described to better understand the concept of positive/negative and easy/hard samples. Suppose to have the task of binary segmenting marine debris versus background on satellites images acquired over the ocean. Negative samples are background pixels while positive samples are marine debris pixels. In particular, easy negative examples could be marine water pixels since they would probably be present in almost every image and therefore be classified correctly with a high probability (e.g. higher than 60\%). Instead, hard negative examples could be pixels of ships or of algae. They are negative because are not marine debris and hard because probably not so many images contain them and so the model would probably classify them as background with a low probability (e.g. less than 40\%).  

\begin{figure}[H]
    \centering
    \includegraphics[width=0.8\textwidth]{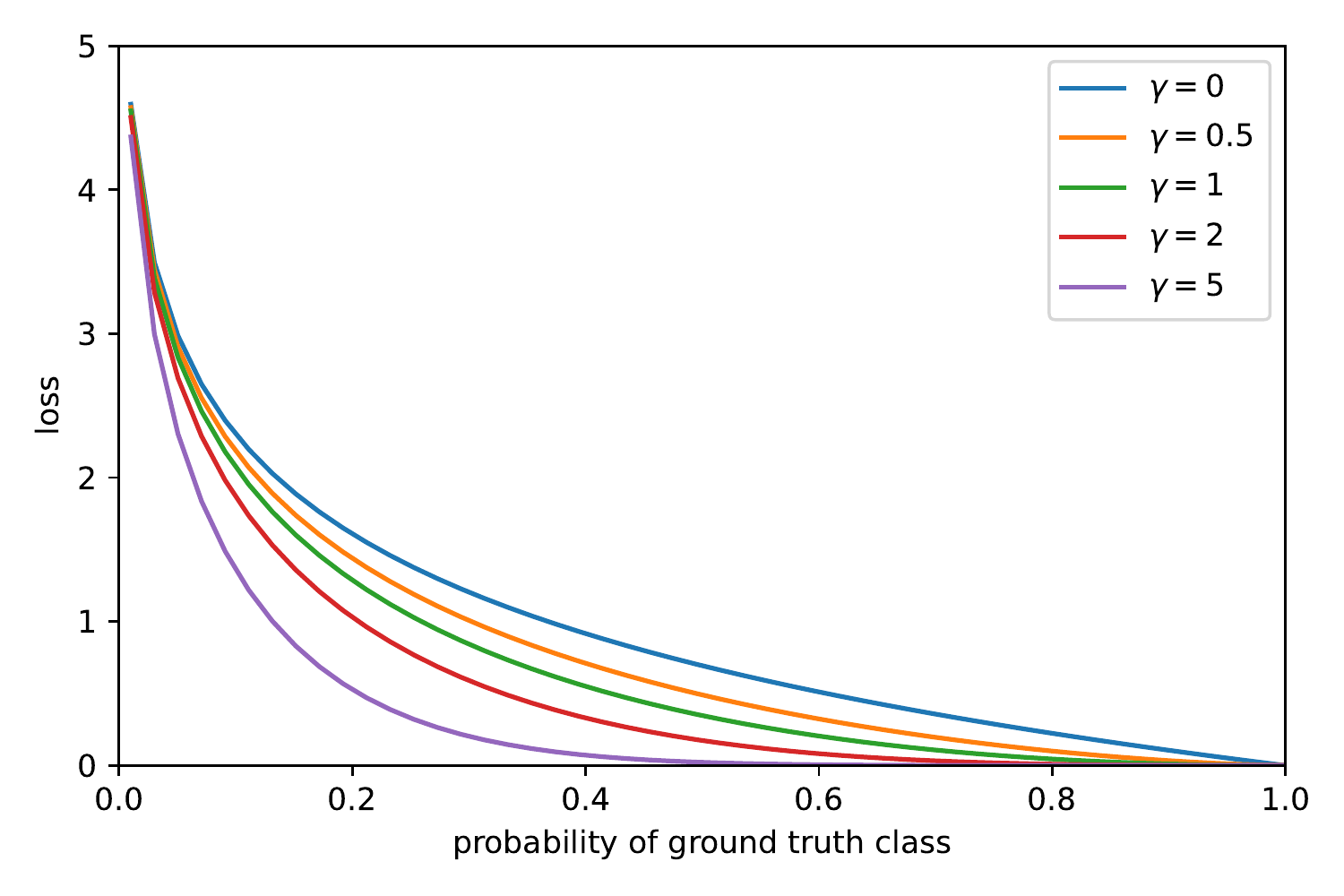}
    \caption{Example of focal loss results with different focusing parameter values. Adapted from \cite{lin2017focal}.}
    \label{Figure:focal_loss}
\end{figure}

Figure \ref{Figure:focal_loss} shows the effect of $\gamma$. When $\gamma$ is zero, the focal loss corresponds to the cross entropy loss. A higher $\gamma$ decreases the loss everywhere. However, the reduction is much greater for well-classified examples (easy samples), which makes the loss very small for samples classified correctly with a high probability (e.g. above 60\%).
On the other hand, the standard cross entropy is very small only when the predicted probability of ground truth class of a sample is close to 1.

In this work we use the focal loss for its capability in \first handling class imbalance, \second differentiating between easy and hard samples.

\section{Model}
\label{sec:meth_model_arch}

\begin{figure}[H]
    \centering
    \includegraphics[width=\textwidth]{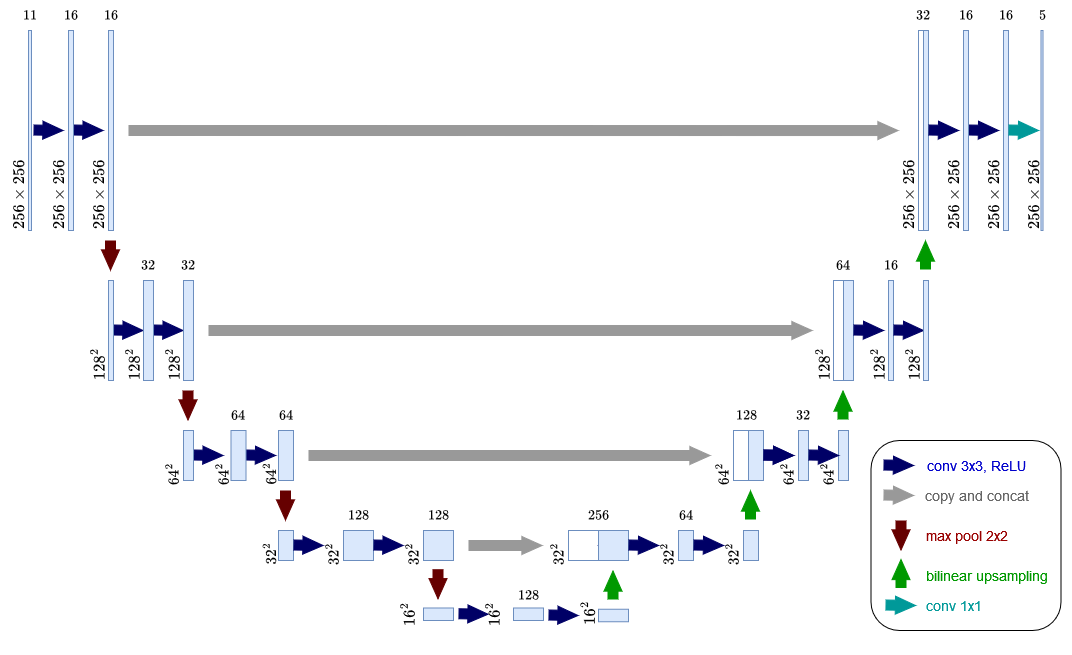}
    \caption{Architecture of the U-Net used in this work. Adapted from \cite{ronneberger2015unet}.}
    \label{Figure:unet_architecture}
\end{figure}

Figure \ref{Figure:unet_architecture} illustrates the U-Net \cite{ronneberger2015unet} architecture that was used to train each model in this work. The figure also shows the sizes of the input image, the output segmentation map, and the intermediate feature maps. 

The downsampling path consists of convolutional layers with $3 \times 3$ kernels and max pooling layers with $2 \times 2$ kernels. The upsampling is composed of bilinear upsampling layers and convolutional layers, with $2 \times 2$ and $3 \times 3$ kernels respectively. A final $1 \times 1$ convolutional layer is applied to adjust the number of channels to match the number of classes.

The gray arrows in Figure \ref{Figure:unet_architecture} represent the skip connections, which concatenate the feature maps of the upsampling path with the correspondingly feature maps from the downsampling path.

\subsection{Onboard characteristics}
The U-Net \cite{ronneberger2015unet} was already considered in feasibility studies for onboard missions \cite{ghasemi2023feasibility}.

Every model we used in our work has 841 099 parameters and requires 2.07 GFLOPs (giga floating point operations) to perform inference on a patch of size (11, 256, 256). Therefore, the model could be deployed to space-rated hardware, such as the Unibap SpaceCloud iX5-106, whose \gls{GPU} has a compute performance of 87 GFLOPS \cite{unibap_ix_5_106}. The number of parameters and the floating point operations of the model were estimated using \textit{flopth}\footnote{https://pypi.org/project/flopth/}. The \textit{.pth} file of each model weighs 3.25 MB, which is very light-weight compared to the SpaceCloud iX5-106's storage of 240 GB \cite{unibap_ix_5_106}.

\section{FixMatch for Semantic Segmentation}
\label{sec:meth_fixmatch_sem_seg}

We adapt FixMatch \cite{sohn2020fixmatch} to our problem, which consist in performing semantic segmentation on multispectral images. The inspiration came from \cite{patel2021evaluating}. However, the authors of \cite{patel2021evaluating} did not work with weakly-labeled annotations. Moreover, the work of \cite{patel2021evaluating} is not open-source. 

The implementation of this project builds on and extends the FixMatch-pytorch\footnote{https://github.com/kekmodel/FixMatch-pytorch} repository, which is an implementation of FixMatch \cite{sohn2020fixmatch} based on PyTorch. Compared to the FixMatch-pytorch repository, this project adapts FixMatch to be used for weakly-semantic segmentation (refer to Subsubsection \ref{sec:weakly-sem-seg}) tasks and to work with multispectral images. FixMatch \cite{sohn2020fixmatch} was initially developed to tackle image classification tasks. This project adapts it to tackle semantic segmentation tasks. The main difference is that our implemented algorithm makes a classification for each pixel in one image instead of just one classification per image. 

\subsection{FixMatch}
\label{subsection:sota_ssl_ic}

FixMatch \cite{sohn2020fixmatch} is one of the most popular state-of-the-art methods in the application of semi-supervised learning to the task of image classification. It gained substantial attention due to both its simplicity and its effectiveness.

FixMatch uses both pseudo-labeling (Subsection \ref{subsec:pseudo-labeling}) and consistency regularization (Subsection \ref{subsec:consistency-regularization}) \gls{SSL} techniques. In particular, it creates pseudo-labels starting from weakly-augmented unlabeled images if their prediction is highly confident. Then, it applies consistency regularization by enforcing the prediction on strongly-augmented images to be similar to the pseudo-labels by computing the cross-entropy loss on them. The final loss FixMatch uses has the form of Equation \ref{eq:consist-regulariz-loss}.

To weakly augment images, the authors randomly flipped images horizontally and translated images vertically and horizontally, whilst the RandAugment \cite{cubuk2020randaugment} and CTAugment \cite{berthelot2019remixmatch} methods followed by Cutout \cite{devries2017improved} method are utilized to strongly augment images.

FixMatch achieved state-of-the-art results on standard datasets used for evaluating semi-supervised learning methods such as CIFAR-10 \cite{cifar-10}. It also reached an accuracy of 88.61\% on it just by using 4 labels per class.

\subsection{Adaptating FixMatch for Semantic Segmentation}

A difference between the original implementation of FixMatch \cite{sohn2020fixmatch} and the implementation of FixMatch for semantic segmentation concern the application of the augmentations to the labels of the data. 

In the case of image classification, an augmentation that applies a geometric transformation to an input image (e.g. rotating the image) does not necessarily imply a modification of its label. For example, when performing image classification among ships and algae, the label of a rotated ship image is still ship. 

On the other hand, in a semantic segmentation task, the geometric transformations applied to an image must be applied to its corresponding semantic segmentation map too. Indeed, geometric transformations change the coordinates of pixels in an image. Therefore, the coordinates of their corresponding labels needs to be changed too. In this way, the label of the pixel at location $(x_i, y_i)$ of the augmented segmentation map will correspond to the input pixel $(x_i, y_i)$ of the augmented image.

In addition, the loss used in FixMatch has to be modified to suit the task of semantic segmentation, and will now be defined.

\subsubsection{Total loss}

As mentioned in Subsection \ref{subsection:sota_ssl_ic}, FixMatch is a semi-supervised learning that makes use of both pseudo-labeling (Subsubsection \ref{subsec:pseudo-labeling}) and consistency regularization (Subsubsection \ref{subsec:consistency-regularization}), and its loss is of the form of Eq. \ref{eq:consist-regulariz-loss}, i.e. $\mathcal{L} = \mathcal{L}^s + \lambda \mathcal{L}^u$.


The supervised and unsupervised components of the loss of FixMatch for semantic segmentation will be now defined.

\subsubsection{Supervised loss}
\label{subsubsec:sup_loss}

The supervised loss is defined as:

\begin{equation}
    \mathcal{L}^s = \frac{1}{H W B} \sum_{h=1}^{H} \sum_{w=1}^{W} \sum_{b=1}^{B} \overbrace{\mathbb{1}(\text{arg max}(p_{h, w, b}) \neq -1)}^{\text{ignoring unlabeled pixels}} \cdot F(\overbrace{p_{h, w, b}}^{\text{label}},   \overbrace{p_m(y \vert \underbrace{\kappa(x_{h, w, b})}_{\substack{\text{weakly-}\\ \text{augmented}\\ \text{pixel}}})}^{\substack{\text{predicted class} \\\text{distribution}}} )
    \label{eq:Ls_fixmatch_sem_seg}
\end{equation}

In Eq. \ref{eq:Ls_fixmatch_sem_seg}, $F$ is the focal loss  computed between two probability distributions. The focal loss is used because it addresses class imbalance (\ref{subsubsec:focal-loss}). 

$\kappa(\cdot)$ is a function that applies a weak augmentation to an input. 

$p_{h, w, b}$ denotes the one-hot label for the pixel located at coordinates $(h, w)$ of the $b$-th input image of the current weakly-labeled batch. 

$p_m(y \vert \kappa(x_{h, w, b}))$ is the predicted class distribution produced by the model ($m$) for the weakly-augmented pixel $x$ located at coordinates $(h, w)$ of the $b$-th input image of the current weakly-labeled batch. 

The supervised loss (Eq. \ref{eq:Ls_fixmatch_sem_seg}) ignores pixels that are not labeled. Indeed, the indicator function $\mathbb{1}(\text{arg max}(p_{h, w, b}) \neq -1)$ is equal to zero when a pixel is unlabeled
. Otherwise, the indicator function is equal to $1$. 

$B, H, W$ are respectively the 
batch size, and the height and width of input images. The denominator of Eq. \ref{eq:Ls_fixmatch_sem_seg} does not exactly correspond to $H \cdot W \cdot B$, but it corresponds to the number of labeled pixels of the current batch because unlabeled pixels are ignored in the computation of this loss.

\subsubsection{Unsupervised loss}
\label{subsubsec:unsup_loss}

The unsupervised loss is defined as:

\begin{equation}
    \mathcal{L}^u = 
    \frac{1}{H W \mu B} \sum_{h=1}^{H} \sum_{w=1}^{W} \sum_{b=1}^{\mu \cdot B} \underbrace{\mathbb{1}(\max(q_{h, w, b}) \geq \tikzmarknode{tau}{\tau} )}_{\substack{\text{ignoring low-confidence}\\\text{pseudo-labels}}} \cdot CE(\underbrace{\hat{q}_{h, w, b}}_{\substack{\text{pseudo-}\\\text{labels}}}, \overbrace{p_m(y \vert \underbrace{\mathcal{K}(u_{h, w, b})}_{\substack{\text{strongly-}\\\text{augmented}\\\text{pixel}}})}^{\substack{\text{predicted class} \\\text{distribution}}})
    \label{eq:Lu_fixmatch_sem_seg}
\end{equation}

\begin{tikzpicture}[overlay,remember picture,black,>=stealth,shorten
 <=0.2ex,nodes={font=\scriptsize,align=left,inner ysep=1pt},<-]
  \draw (tau.north) -- ++ (0,1.5em) node[anchor=south west,align=left,xshift=-1.2ex] (o) {threshold};
\end{tikzpicture}

In Eq. \ref{eq:Lu_fixmatch_sem_seg}, $\tau$ is a scalar hyperparameter that indicates the probability threshold above which a pseudo-label is kept. Indeed, the indicator function ignores the pixels whose maximum predicted probability is less than $\tau$. 

In addition, all pixels that were added due to padding (when applying strong augmentations) are also ignored when computing the loss.

\gls{CE} is the cross-entropy loss computed between two probability distributions. In our unsupervised loss calculation, we opted to use the cross-entropy loss instead of the focal loss. The reason behind this choice is that the focal loss reduces the impact of samples that are predicted with high confidence. However, these high-confidence samples are the only ones that are considered in the unsupervised loss, unless the value of $\tau$ is low.

$\mathcal{K}(\cdot)$ is a function that applies a strong augmentation to an input. 

$q_{h, w, b} = \hat{\mathcal{K}}(p_m(y \vert \kappa(u_{h, w, b})))$ is the predicted class distribution by the model ($m$) for the weakly-augmented unlabeled pixel $u$ located at coordinates $(h, w)$ of the $b$-th input image of the current unlabeled batch. $\hat{\mathcal{K}}$ is a function that only applies geometric strong augmentations to an input. 
Its purpose is to ensure that the coordinates of the predicted class distribution align with the pixel coordinates for which the prediction was made.

$\hat{q}_{h, w, b} = \text{arg max}(q_{h, w, b})$. In this case, it is assumed that $\text{arg max}$ applied to a probability distribution produces a one-hot probability distribution. So, $\hat{q}_{h, w, b}$ is the one-hot encoded pseudo-label obtained for the unlabeled pixel $u_{h, w, b}$.

$p_m(y \vert \mathcal{K}(u_{h, w, b}))$ is the predicted class distribution produced by the model for the strongly-augmented pixel $u$ located at coordinates $(h, w)$ of the $b$-th input image of the current unlabeled batch.

$B, H, W$ are respectively the 
labeled batch size, and the height and width of input images. The denominator of Eq. \ref{eq:Lu_fixmatch_sem_seg} does not exactly correspond to $H \cdot W \cdot \mu \cdot B$, but it corresponds to the number of unlabeled pixels (excluding the ones added due to padding) of the current batch whose pseudo-label probability is greater than the threshold $\tau$.

$\mu$ is also a scalar hyperparameter and denotes the ratio between the unlabeled batch and the labeled batch.

\begin{figure}[H]
    \centering
    \includegraphics[width=\textwidth]{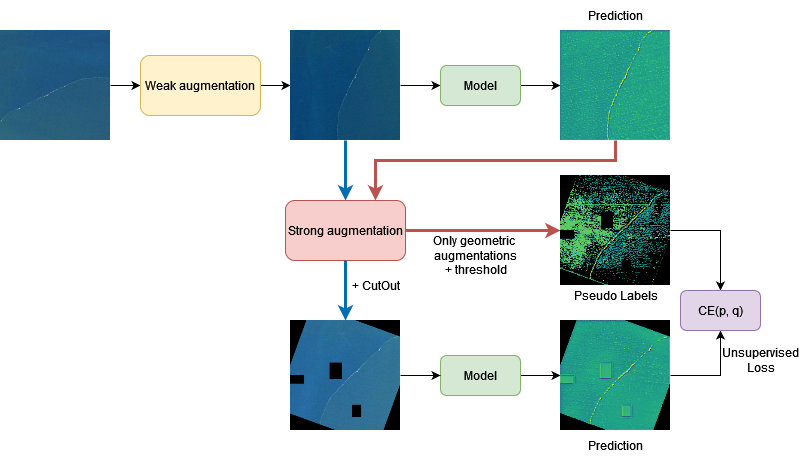}
    \caption{Diagram that illustrates how the unsupervised loss of FixMatch for semantic segmentation is computed using an unlabeled input image.  Adapted from \cite{patel2021evaluating}.}
    \label{Figure:ssl-loss-diagram}
\end{figure}

Figure \ref{Figure:ssl-loss-diagram} illustrates how the unsupervised component $\mathcal{L}^u$ (Eq. \ref{eq:Lu_fixmatch_sem_seg}) of the loss of FixMatch for Semantic Segmentation is computed. A weak augmentation is applied to an unlabeled input image. The model predicts the semantic segmentation map of the weakly augmented image. 

A strong augmentation is then applied to the weakly augmented image. The strong augmentation is also applied to the predicted pseudo-label map to make sure its coordinates spatially correspond to the ones of the strongly-augmented image. However, only geometric augmentations, which do not alter the intensity of pixels, are applied to the predicted pseudo-label map. This is done to avoid changing the model's predicted class distribution values.

Additionally, Cutout \cite{devries2017improved} is applied to the strongly augmented image. Cutout stimulates the model to use a more diverse set of features to overcome the varying combinations of image portions that may be visible or masked \cite{french2019semi}. The pixels masked by Cutout are not considered later when computing the unsupervised loss.

The threshold $\tau$ of Eq. \ref{eq:Lu_fixmatch_sem_seg} is applied to the predicted pseudo-label map. The objective of the threshold is to ignore the predicted pixels whose maximum probability value is lower than the threshold.
This is done to give more importance to the quality rather than the quantity of pseudo-labels. Indeed, \cite{sohn2020fixmatch} found that 
the correctness of pseudo-labels for unlabeled data is higher with a higher threshold value (increasing the quality of pseudo-labels) rather than with a low one (increasing the quantity of pseudo-labels).

The pseudo-labels of the input image are therefore generated. Afterwards, the model predicts the segmentation map of the strongly augmented input image. The unsupervised loss is then computed between this strongly augmented image and the pseudo-labels.

\subsection{Data Augmentation}

The image augmentations that can be applied during training by FixMatch for Semantic Segmentation will be now described. 

\subsubsection{Weak Augmentation}

The weak augmentations that the function $\kappa(\cdot)$ (refer to Subsubsection \ref{subsubsec:sup_loss}) applies to an input image are now listed. The function consecutively applies a horizontal flip, a vertical flip and a rotation of $-90$ or $90$ degree to each input image. The horizontal and vertical flip are modeled as independent events drawn from a Bernoulli random variable $FP \sim Ber(0.5)$. The probability mass function of a Bernoulli random variable with $p=0.5$ is:
\begin{equation}
    \begin{cases}
      P(FP = 1) = p = 0.5\\
      P(FP = 0) = 1 - p = 0.5 .
    \end{cases}
    \label{eq:flipping-prob}
\end{equation}
In Eq. \ref{eq:flipping-prob}, $P(FP = 1)$ indicates the probability of flipping the input image and $P(FP = 0)$ the probability of not flipping the input image.

\subsubsection{Strong Augmentation}
\label{subsubsec:strong_augs_description}
The strong augmentations that the function $\mathcal{K}(\cdot)$ (refer to Subsubsection \ref{subsubsec:unsup_loss}) applies to an input image are listed in Table \ref{table:strong-augmentations}. The strong augmentations are applied utilizing the two Python libraries Albumentations \cite{buslaev2020albumentations} and imgaug \cite{jung2019imgaug}.

Given an input image, the function $\mathcal{K}(\cdot)$ consecutively applies two strong augmentations. In particular, two strong augmentations and their values are respectively drawn from the list of strong augmentations and their parameter ranges (Table \ref{table:strong-augmentations}) with a uniform distribution. In addition, the two strong augmentations are drawn with repetition, thus the same strong augmentation could be drawn and applied twice. During training, two strong augmentations are drawn at each training step and applied to all input images of the batch.

The utilized strong augmentations are divided into two types: \textit{Geometric} and \textit{Color}. Geometric augmentations alter an image by changing the position of pixels and possibly adding padding pixels. On the other hand, Color augmentations modify the intensity of pixels. The Color augmentations, specifically Sharpness and Solarize, are only applied as first augmentation. They are not applied as a second transformation to avoid altering  the intensity of padding pixels (introduced by Geometric augmentations), which should be ignored when computing the loss. Moreover, as already detailed in Subsubsection \ref{subsubsec:unsup_loss}, only geometric augmentations are applied to pseudo-labels.

In Table \ref{table:strong-augmentations}, the \textit{Identity} augmentation returns the input image that received as input, unchanged. Hence, it has no parameter range. 

\begin{table}[H]
\caption{List of applied strong augmentations and the range of their parameters.}
\label{table:strong-augmentations}
\centering
\begin{tabular}{lc}
    \toprule
    \textbf{Augmentation} & \textbf{Parameter Range} \\ 
    \midrule
    \textbf{Geometric Augmentations} & \\ 
    \midrule
    Identity     & -               \\ 
    Rotate       & $ \left[ -30, -5 \right] \cup  \left[ 5, 30 \right] $   \\ 
    ShearX       & $ \left[ -30, -5 \right] \cup  \left[ 5, 30 \right] $     \\
    ShearY       & $ \left[ -30, -5 \right] \cup  \left[ 5, 30 \right] $     \\ 
    TranslateX   & $ \left[ -0.2, -0.1 \right] \cup  \left[ 0.1, 0.2 \right]$  \\ 
    TranslateY   & $ \left[ -0.2, -0.1 \right] \cup  \left[ 0.1, 0.2 \right]$  \\
    \midrule
    \textbf{Color Augmentations} & \\
    \midrule
    Solarize     & $ \left[ 0.01, 0.99 \right]$    \\ 
    Sharpness    & $ \left[ 0.2, 0.5 \right]$  \\ 
    \bottomrule
\end{tabular}
\end{table}

\subsubsection{Cutout}

Cutout \cite{devries2017improved} is a simple regularization
technique that randomly masks rectangular regions of the input when training a neural network.

As mentioned in Subsubsection \ref{subsubsec:unsup_loss}, in this project, Cutout is applied to the strongly augmented image. In particular, three rectangles randomly cut out each strongly augmented image by setting the cutout pixels' intensity to $0$. The three rectangles could possibly overlap among themselves.

The height and width of each rectangle are respectively drawn from the two uniform random variables $H_{rect}\sim U(0.05, 0.15) \cdot H$ and $W_{rect}\sim U(0.05, 0.15) \cdot W$, where $H$ and $W$ respectively denote the height and the width of the input image. Afterwards, the coordinates $x$ and $y$ of the center of each rectangle are drawn from other two uniform random variables $X\sim U(1, W)$ and $Y\sim U(1, H)$, where $H$ and $W$ respectively denote the height and the width of the input image. Hence, the center of a rectangle could be placed on the border of an input image and mask out less than $H \cdot W$ pixels.

\section{Evaluation Metrics}
\label{sec:meth_eval_metrics}

There exist different metrics to evaluate the predicted semantic segmentation maps against the ground truth ones. This section defines several evaluation metrics used for semantic segmentation and justifies the chosen evaluation metric. 

\subsection{Pixel Accuracy}

Pixel accuracy (\gls{PA}) is a semantic segmentation metric that denotes the percentage of
pixels that are accurately classified in the dataset.

Considering one single image, the pixel accuracy on an image \textit{I} of dimensions $H$x$W$ is:
\begin{equation}
PA = \frac{\sum_{c=1}^{C} TP_c}{H \cdot W}
\end{equation}
where the numerator is the sum of all true positive pixels of all classes (i.e. all pixels in image \textit{I} both classified and labeled with the same class), and the denominator is the total number of pixels in the image.

The pixel accuracy on a dataset that contains $N$ images is:  
\begin{equation}
PA = \frac{\sum_{n=1}^{N}\sum_{c=1}^{C} TP_c}{H \cdot W \cdot N}
\end{equation}
where the numerator is the sum of all true positive pixels of all classes across all $N$ images, and the denominator is the total number of pixels in the dataset.

\begin{figure}[H]
    \centering
    \includegraphics[width=0.7\textwidth]{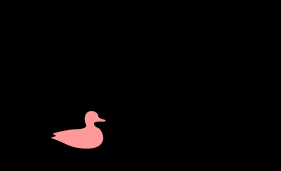} 
    \caption{Example of ground truth segmentation map with class imbalance.}
    \label{Figure:sem_seg_imbalance_bad_accuracy-example}
\end{figure}

However, this metric is not ideal when having imbalanced classes. Figure \ref{Figure:sem_seg_imbalance_bad_accuracy-example} shows an example of an image with two classes (bird and background). The background class clearly contains more pixels than the bird class so there is an imbalance of pixels between the two classes. Suppose that the background class contains 90\% of the pixels of the image, while the bird class only contains 10\% of them (ignore the void class). A semantic segmentation model that always predicts the background class for every pixel will obtain a pixel accuracy of 90\% on this image. 

Thus, a model that always predicts the same class will still perform well in this case, and that is the reason why pixel accuracy is not a good measure when having class imbalance.  




\subsection{mIoU}
\label{subsubsection:miou}

The \gls{IoU} (intersection over union), also known as the Jaccard index \cite{jaccard1912distribution}, between two sets is defined as the intersection between them divided by their union \cite{taha2015metrics}, that is:
\begin{equation}
IoU = \frac{A \cap B}{A \cup B}
\end{equation}

By considering a dataset of $N$ images of size $H \text{x} W$ with $C$ different classes, the intersection over union of the $c$-th class is defined as:
\begin{multline}
\begin{aligned}
    &IoU_c =\\ &=\frac{\sum_{n=1}^N \sum_{i=1}^{H} \sum_{j=1}^{W} \mathbb{1}_{inter}(n, i, j, c)}{\sum_{n=1}^N \sum_{i=1}^{H} \sum_{j=1}^{W} \mathbb{1}_{pred}(n, i, j, c) + \mathbb{1}_{gt}(n, i, j, c) - \mathbb{1}_{inter}(n, i, j, c)}\\
    &= \frac{TP_c}{TP_c + FP_c + FN_c}
\end{aligned}
\label{eq:iou_c}
\end{multline}
where 
\begin{align*}
\mathbb{1}_{inter}(n, i, j, c) &=    
\begin{cases}
     1 & \text{if } y_n(i, j) = c \text{ and } \hat{y}_n(i, j) = c,\\
     0 & \text{otherwise,}
\end{cases}
\\
\mathbb{1}_{pred}(n, i, j, c) &=    
\begin{cases}
     1 & \text{if } y_n(i, j) = c,\\
     0 & \text{otherwise,}
\end{cases}
\\
\mathbb{1}_{gt}(n, i, j, c) &=    
\begin{cases}
     1 & \text{if } \hat{y}_n(i, j) = c,\\
     0 & \text{ otherwise.}
\end{cases}
\end{align*}

$y_n(i,j)$ and $\hat{y}_n(i,j)$ are respectively the true and predicted class of the ($i$, $j$)-th pixel of the $n$-th input image of the dataset. 

So, $\mathbb{1}_{inter}(n, i, j, c)$ is an indicator function that is equal to $1$ when both the predicted and the true class of the ($i$, $j$)-th pixel of the $n$-th sample are equal to class $c$, which is a true positive pixel of class $c$. The numerator of Equation \ref{eq:iou_c} is therefore the total number of true positives of class $c$. The denominator of Equation \ref{eq:iou_c} is the union of pixels that are predicted with class $c$ and the pixels whose true label is class $c$. Therefore, the denominator corresponds to the sum of the total true positive, false positive, and false negative pixels of class $c$.

The \gls{mIoU} (mean intersection over union) is the mean of the \glspl{IoU} of all the classes:
\begin{equation}
mIoU = \frac{1}{C} \sum_{c=1}^{C} IoU_c
\end{equation}
$mIoU \in \left[0,1\right]$, and a higher value of the \gls{mIoU} signifies a better performance.

The \gls{mIoU} solves the problem of that the \gls{PA} evaluation metric has  with class imbalance because it evaluates the overlap between the predicted and the ground truth pixels. Therefore, we use the \gls{mIoU} as evaluation metric in this work.

\section{Experimental Design}
\label{sec:meth_exp_design}

This section elucidates the methodology employed for conducting and assessing our experiments. 
It begins with an introduction to the method used to split the dataset into two distinct subsets when training semi-supervised learning models (Subsection \ref{subsec:two-train-set}).
This is followed by an overview of the hyperparameters in Subsection \ref{subsec:exp_design_hyperparams}, and a description of the evaluation techniques used for model selection and comparison in Subsection \ref{subsec:exp-eval-metric}.



\subsection{Two Training Sets Setting}
\label{subsec:two-train-set}

We used two separate training sets to train semi-supervised learning models.
The "Two Training Sets" setting consists in splitting the original training set ($\mathcal{D}$) into two distinct subsets: one labeled ($\mathcal{D}^l$) and one unlabeled ($\mathcal{D}^u$). So, for example, if $\mathcal{D}$ contains $p$ images, then  $p_l$ ($0 < p_l < p$) images will be assigned to $\mathcal{D}^l$ and $\mathcal{D}^u$ will contain the remaining $p - p_l$ images.

Every image has labeled and unlabeled pixels. In this setting, each image is either in the labeled set $\mathcal{D}^l$ or unlabeled set $\mathcal{D}^u$. For images in $\mathcal{D}^l$, only the labeled pixels are utilized during the training process. Conversely, all pixels of images in $\mathcal{D}^u$, regardless of whether they are labeled or not, are treated as unlabeled and used in training.

\subsubsection{Training Set Configurations}
\label{subsec:lab-train-config}

The "Two Training Sets" setting considers nine distinct configurations of the training set. The labeled subset, $\mathcal{D}^l$, corresponds to a percentage ($n \%, \text{with } n = 5, 10, 20, 30, \ldots, 80$) of the total number of training images contained in the entire training set $\mathcal{D}$. Consequently, the unlabeled subset, $\mathcal{D}^u$, contains the remaining ($100 - n$) $\%$ of the total number of training images in $\mathcal{D}$.

The labeled subsets $\mathcal{D}^l$ from two different percentage configurations might have an intersection, but it’s not a certainty. In fact, the images of each labeled subset $\mathcal{D}^l$ are selected randomly from $\mathcal{D}$ each time.

In addition, the labeled subset $\mathcal{D}^l$ should also contain, for each class, $n \%$ of the total number of labeled training pixels. However, the MARIDA dataset \cite{kikaki2022marida} is weakly-labeled and its classes are imbalanced. Hence, two images can contain a different number of labeled pixels and have a different distribution of labeled classes. Therefore, it is complex to find a subset $\mathcal{D}^l$ of $\mathcal{D}$ that contains a exactly $n\%$ of the total number of training labeled pixels for each class. For this reason, a subset $\mathcal{D}^l$ of $\mathcal{D}$ is chosen so that it contains $n \in \left[ m - 5, m + 5 \right] \%$ of the total number of training labeled pixels for each class, with $m = 5, 10, 20, 30, \ldots, 80$.

When training using the "Two Training Sets" setting, a training epoch is defined as one complete cycle of training the model using the entire labeled subset $\mathcal{D}^l$. Each epoch consists of several training steps. Hence, the number of training steps increases with the size (the percentage of labeled data) of the labeled subset $\mathcal{D}^l$.

The validation and test sets are the same as the ones stated in \cite{kikaki2022marida}, except for the patches containing \gls{NaN} values listed in Subsubsection \ref{subsubsec:nan_patches}, which were excluded. The patch having \gls{NaN} values (Subsubsection \ref{subsubsec:nan_patches}) in the training set is also excluded because the applied augmentations substantially increase the number of \gls{NaN} values in the augmented images.

\subsection{Hyperparameters}
\label{subsec:exp_design_hyperparams}

Table \ref{table:hyperparams_descritption} contains the description of the hyperparameters used in the experiments of this chapter. The ones listed under \textit{General} are for both fully- and semi-supervised models, whilst the hyperparameters listed under \textit{Semi-supervised} are utilized only in semi-supervised models. The values of the hyperparameters contained in Table \ref{table:hyperparams_descritption} will be listed in the methodology of each experiment. 

In all experiments, we set the batch size and the mu hyperparameters to 5. This was due for computational efficiency during training. More specifically, the mu parameter, which determines the ratio between unlabeled and labeled batches, was calibrated to limit the number of images in one semi-supervised training step that would otherwise cause the \gls{GPU} to run out of memory.




Every deep learning model was trained by using Adam \cite{kingma2014adam} as the optimization algorithm.

In our training process, we discard the last incomplete batch, by setting the parameter \texttt{drop\_last=True}\footnote{https://pytorch.org/docs/stable/data.html\label{pytorch-data}}. This approach ensures that the backward pass of the network is not influenced by a smaller number of images than the designated batch size. Despite discarding the last batch, all training data is still utilized over time, as the data is reshuffled at the start of each epoch, which is ensured by setting \texttt{shuffle=True}\footref{pytorch-data}. 

\begin{table}[H]
\caption{Description of the hyperparameters used in the following experiments.}
\label{table:hyperparams_descritption}
\centering
\begin{tabular}{ll}
    \toprule
    \multicolumn{2}{l}{\rule{0pt}{4ex}\textbf{General}}        \\ 
    \midrule
    epochs                         & Number of epochs for which the model \\ & is trained         \\ 
    lr                         & Learning rate                             \\ 
    batch size                 & Size of the batch                               \\ 
    alphas ($\alpha$)                    & coefficients of the Focal loss (see Subsection \ref{subsubsec:focal-loss}) \\ 
    gamma ($\gamma$)                      & coefficient of the Focal loss (see Subsection \ref{subsubsec:focal-loss}) \\ 
    \midrule             
    \multicolumn{2}{l}{\rule{0pt}{4ex}\textbf{Semi-supervised}}                                                      \\ 
    \midrule
    threshold ($\tau$)                & Threshold for pseudo-labels (see Subsubsection \ref{subsubsec:unsup_loss})                                \\ 
    lambda coeff ($\lambda$)                & Coefficient that weighs the contribution of \\ & the unsupervised loss function (see Subsubsection Eq. \ref{eq:consist-regulariz-loss})                              \\ 
    mu ($\mu$)                &    Ratio between the unlabeled batch and the \\ & labeled batch (see \ref{subsubsec:unsup_loss})                            \\ 
    \bottomrule
\end{tabular}
\end{table}

\subsection{Evaluation}
\label{subsec:exp-eval-metric}

The methods to evaluate and select \first the best version of a single model and \second the best model among two or more of them will be now outlined.

When having a single model that is trained for several epochs on a dataset, its performance is evaluated after each epoch based on the \gls{mIoU} (refer to Subsection \ref{subsubsection:miou}) on the validation set. The version of the model that achieves the highest \gls{mIoU} during these epochs is selected for final evaluation on the test set.

The \gls{mIoU} computed on the validation set is also the evaluation metric used to compare two or more models that were trained using different hyperparameters' values. The model that achieves the highest \gls{mIoU} among the tested models is selected for final evaluation on the test set.

The described evaluation method is used for both fully- and semi-supervised learning models.

Lastly, the performance of a fully-supervised model is compared to its semi-supervised counterpart using the \gls{mIoU} score on the test set.

\section{Preliminary Experiments}
\label{sec:exp-preliminary-1}
This section delves into the specifics of two preliminary hyperparameter selection experiments. The findings from these preliminary experiments will be used in the main experiment, which is described in Section \ref{subsec:exp_design_exp_2_train_sets}.

Firstly, the experiment aimed at determining the optimal probability threshold ($\tau$ in Eq. \ref{eq:Lu_fixmatch_sem_seg}) under conditions of limited labeled data is described in Subsection \ref{exp_finding_best_thresh_results}.
Secondly, Subsection \ref{subsec:exp_design_exp_ce_vs_focal_loss} discusses an experiment comparing the performance of cross-entropy and focal loss in the unsupervised component of semi-supervised model loss.

\subsection{Finding the Best Probability Threshold}
\label{exp_finding_best_thresh_results}

The purpose of our experiment is to identify an optimal value for the probability threshold that can effectively train semi-supervised models. The threshold parameter, denoted as $\tau$ in Eq. \ref{eq:Lu_fixmatch_sem_seg}, is responsible for the exclusion of predicted pixels from the computation of the loss when their maximum probability value is lower than the threshold. $\tau$ can take any value in the interval $\left[ 0, 1 \right]$. If we set $\tau$ to 0, it implies that the pseudo-labels of all pixels are taken into account when calculating Eq. \ref{eq:Lu_fixmatch_sem_seg}. Conversely, if $\tau$ is set to a value greater than $1$, it results in the exclusion of all pseudo-labels.

Hence, in this experiment, we evaluate the performance of various semi-supervised models by keeping all their hyperparameters constant, except for the threshold $\tau$. Our aim is to investigate how varying the threshold influences the performance of the models and identify the value of $\tau$ that yields the highest scores.

We conducted tests using the following thresholds: 0.5, 0.6, 0.7, 0.8, 0.9. Due to constraints on time, we limited our testing to the scenario with 10\% labeled data from the “Two Training Sets” configuration (refer to Subsection \ref{subsec:two-train-set}). The evaluation methodology employed is detailed in Subsection \ref{subsec:exp-eval-metric}.

\subsubsection{Hyperparameters' Values}

The hyperparameters' values of the tested semi-supervised models are listed in Table \ref{table:hyperparams_best_prob_threshold}.

The learning rate and the batch size are the same used in \cite{kikaki2022marida}. 
We decided to train for 500 epochs as we noticed that extending the training to 2000 epochs did not yield significant additional insights. The use of focal loss leads to additional hyperparameters, which are listed in Table \ref{table:hyperparams_best_prob_threshold}. Leveraging on the focal loss represents an element of novelty compared to \cite{kikaki2022marida}.

We set the coefficient that weighs the contribution of the unsupervised loss function to $1$. This was done to ensure that the supervised and unsupervised components of the total loss are equally weighted.

A comprehensive description of these hyperparameters can be found in Table \ref{table:hyperparams_descritption}.

\begin{table}[H]
\caption{Hyperparameters' values of the tested semi-supervised models.}
\label{table:hyperparams_best_prob_threshold}
\centering
\begin{tabular}{ll}
    \toprule
    epochs                         & \multicolumn{1}{r}{500}                            \\ 
    lr                         & \multicolumn{1}{r}{2e-4}                            \\ 
    batch size                 & \multicolumn{1}{r}{5}                                \\ 
    alphas ($\alpha$)                    & \multicolumn{1}{r}{$\left[ 1, 1, 1, 1, 1\right]$} \\ 
    gamma ($\gamma$)                     & \multicolumn{1}{r}{2.0} \\ 
    threshold ($\tau$)                        & \multicolumn{1}{r}{0.5, 0.6, 0.7, 0.8, 0.9}                             \\ 
    lambda coeff ($\lambda$)                & \multicolumn{1}{r}{1}                                \\ 
    mu ($\mu$)                & \multicolumn{1}{r}{5}                                \\ 
    \bottomrule
\end{tabular}
\end{table}

\subsection{Cross Entropy Vs Focal loss in the Unsupervised Component}
\label{subsec:exp_design_exp_ce_vs_focal_loss}

As described in Subsubsection \ref{subsubsec:unsup_loss}, the unsupervised loss calculation employs the cross-entropy loss instead of the focal loss. This choice was motivated by the fact that the focal loss diminishes the influence of samples predicted with high probability. However, these high-confidence samples are exactly what we aim to emphasize in our unsupervised loss, especially considering that the optimal threshold identified in Subsection \ref{exp_finding_best_thresh_results} was 0.9. In this experiment, we test whether using cross-entropy in the unsupervised component ($\mathcal{L}^u$, Eq. \ref{eq:Lu_fixmatch_sem_seg}) outperforms the focal loss. It is important to note that this experiment compares the two types of losses in the unsupervised component. However, in both cases, the supervised component ($\mathcal{L}^s$, Eq. \ref{eq:Ls_fixmatch_sem_seg}) utilizes the focal loss.

Due to time constraints, our tests were confined to the 10\% labeled data case of the “Two Training Sets” setup (see Subsection \ref{subsec:two-train-set} for more details). The evaluation approach we used is explained in Subsection \ref{subsec:exp-eval-metric}.

\subsubsection{Hyperparameters' Values}

The values of the hyperparameters for the tested semi-supervised models are detailed in Table \ref{table:hyperparams_ssl_one_train_set}. 

The learning rate and the batch size are the same used in \cite{kikaki2022marida}. In this experiment, the hyperparameter \textit{gamma} represents the coefficient $\gamma$ of the focal loss for both the supervised and unsupervised components of the loss. This applies when the focal loss is used as the unsupervised component. However, when we use cross entropy loss as the unsupervised component of the loss, gamma remains at $2.0$ for the supervised component but is set to $0$ for the unsupervised component. This is because a gamma value of $0$ equates to cross entropy, as indicated in Subsection \ref{subsubsec:focal-loss}. 

We set the threshold to $0.9$ because it was the optimal value we found among the ones tested in the "Finding the Best Probability Threshold" experiment (Subsection \ref{exp_finding_best_thresh_results}). The coefficient that weighs the contribution of the unsupervised loss function is set to $1$. This was done to ensure that the supervised and unsupervised components of the total loss are equally weighted.

The description of the hyperparameters is noted in Table \ref{table:hyperparams_descritption}. 

\begin{table}[H]
\caption{Hyperparameters' values of the tested semi-supervised models.}
\label{table:hyperparams_ssl_one_train_set}
\centering
\begin{tabular}{ll}
    \toprule
    epoch                         & \multicolumn{1}{r}{2000}                             \\ 
    lr                         & \multicolumn{1}{r}{2e-4}                             \\ 
    batch size                 & \multicolumn{1}{r}{5}                                \\ 
    alphas ($\alpha$)                     & \multicolumn{1}{r}{$\left[1, 1, 1, 1, 1\right]$} \\ 
    gamma ($\gamma$)                      & \multicolumn{1}{r}{2.0} \\
    threshold ($\tau$)                & \multicolumn{1}{r}{0.9}                                \\ 
    lambda coeff ($\lambda$)                & \multicolumn{1}{r}{1}                                \\ \bottomrule       
\end{tabular}
\end{table}

\section{Main Experiment - Comparing Semi-Supervised and Fully-Supervised Learning Models}
\label{subsec:exp_design_exp_2_train_sets}

This section outlines the main experiment of our work. In this experiment, we use the hyperparameters' values that we selected by conducting two preliminary experiments (refer to Section \ref{sec:exp-preliminary-1}). 

The main experiment aims to evaluate the performance of fully-supervised and semi-supervised learning models in the "Two Training Sets" setting. Specifically, it investigates whether FixMatch for Semantic Segmentation can enhance the performance by utilizing additional unlabeled data, compared to fully-supervised models solely that rely on labeled data.

The comparison involves training these models with subsets of the labeled training images, as outlined in the “Two Training Sets” setting (refer to Subsection \ref{subsec:two-train-set}). Specifically, a fully-supervised model is trained exclusively on the labeled subset $\mathcal{D}^l$, while a semi-supervised model is trained on both $\mathcal{D}^l$ and $\mathcal{D}^u$ subsets. The subset $\mathcal{D}^u$ is used only to compute the unsupervised component of the total loss. Therefore, all the labeled pixels of the training images of $\mathcal{D}^u$ are treated as unlabeled. While they are excluded from the computation of the supervised component of the loss, they are incorporated in the calculation of the unsupervised component.

To ensure a fair comparison, the same labeled subset $\mathcal{D}^l$ is employed when training both the fully-supervised model and its semi-supervised counterpart.

With a fixed percentage of labeled data, a training epoch, which consists in a full cycle of model training using the entire labeled subset $\mathcal{D}^l$, results in an equivalent number of training steps when comparing a fully-supervised model with its semi-supervised counterpart.
In particular, during each of these steps, a fully-supervised model is trained on a batch of labeled data. Conversely, a semi-supervised model's training step involves a batch that includes both labeled and unlabeled data.

The used evaluation method is the one described in Subsection \ref{subsec:exp-eval-metric}.

\subsection{Hyperparameters' Values}

The values of the hyperparameters for the semi-supervised and fully-supervised models tested in the main experiment are detailed in Table \ref{table:hyperparams_ssl_two_train_sets}.

The learning rate and the batch size are the same used in \cite{kikaki2022marida}.

For the threshold, we tested not only the optimal value of $0.9$ found among the ones tested in the "Finding the Best Probability Threshold" experiment (see Subsection \ref{exp_finding_best_thresh_results}), but also the value of $0.999$. This was motivated by our observation that when using 20\% percentage of labeled data, a threshold of $0.999$ produced superior results. Therefore, we used a threshold of $0.999$ for all percentages greater or equal than 20\%.

The coefficient that weighs the contribution of the unsupervised loss function is set to 1. This was done to ensure that the supervised and unsupervised components of the total loss are equally weighted.

A comprehensive description of these hyperparameters can be found in Table \ref{table:hyperparams_descritption}.

\begin{table}[H]
\caption{Hyperparameters' values of the fully- and semi-supervised models tested in the "Two Training Sets" setting. The threshold and mu parameters are only used for semi-supervised models.}
\label{table:hyperparams_ssl_two_train_sets}
\centering
\begin{tabular}{ll}
    \toprule
    epoch                         & \multicolumn{1}{r}{2000}                             \\ 
    lr                         & \multicolumn{1}{r}{2e-4}                             \\ 
    batch size                 & \multicolumn{1}{r}{5}                                \\ 
    alphas ($\alpha$)                     & \multicolumn{1}{r}{$\left[1, 1, 1, 1, 1\right]$} \\ 
    gamma ($\gamma$)                      & \multicolumn{1}{r}{2.0} \\ 
    threshold ($\tau$)                & \multicolumn{1}{r}{0.9, 0.999}                                \\ 
    lambda coeff ($\lambda$)                & \multicolumn{1}{r}{1}                                \\ 
    mu ($\mu$)                & \multicolumn{1}{r}{5}                                \\ 
    \bottomrule
\end{tabular}
\end{table}

\section{Implementation Details and Hardware}

In this project, we primarily utilized Python  3.10.9 \cite{van1995python} as our programming language, and PyTorch 1.12.1 \cite{paszke2019pytorch} as our deep learning framework. Our experiment scheduling and tracking were managed through Weights \& Biases \cite{wandb}, which also facilitated the generation of our experiment results and visualizations. In addition, Matplotlib \cite{hunter2007matplotlib} was employed for further visualization needs.

A variety of other Python packages and libraries were integral to our project, including Rasterio \cite{gillies2013rasterio}, GDAL \cite{gdal}, NumPy \cite{harris2020array}, OpenCV \cite{bradski2000opencv}, scikit-learn \cite{pedregosa2011scikit}, imgaug \cite{jung2019imgaug}, and Albumentations \cite{buslaev2020albumentations}.

Regarding the hardware, all deep learning models were trained using a NVIDIA A100-SXM4-40GB \gls{GPU}. The training time increases with increasing amount of labeled data since in this project a training epoch consists in a full cycle of model training using the entire labeled subset.

The code of our project is open-source and available at \href{https://github.com/lucamarini22/marine-anomaly-detection}{\nolinkurl{github.com/lucamarini22/marine-anomaly-detection}}.

\cleardoublepage

\chapter{Results}
\label{ch:resultsAndAnalysis}


The aim of this chapter is to present the results of the experiments conducted in our work. 

Section \ref{sec:res_prel_exp} presents the results of two preliminary hyperparameter selection experiments. The findings from these preliminary experiments will be used in the main experiment, which compares the performance of semi-supervised and fully-supervised models. The results of the main experiment are detailed in  Section \ref{subsec:res-fs-vs-ssl-2-train-sets}.

\section{Preliminary Experiments}
\label{sec:res_prel_exp}

We conducted two hyperparameter selection experiments (refer to Section \ref{sec:exp-preliminary-1}). This section outlines their results. 
Firstly, Subsection \ref{exp_results:finding_best_threshold_max_val_miou_10} contains the results of the experiment aimed at determining the optimal probability threshold ($\tau$ in Eq. \ref{eq:Lu_fixmatch_sem_seg}) under conditions of limited labeled training data.
Finally, Subsection \ref{subsec:res-ce-vs-focal-loss} compares the performance of cross-entropy and focal loss in the unsupervised component of semi-supervised model loss.

\subsection{Finding the Best Probability Threshold}
\label{exp_results:finding_best_threshold_max_val_miou_10}

Table \ref{table:finding_best_threshold_max_val_miou_10} compares semi-supervised models when varying the probability threshold. Setting the threshold to 0.9 yielded the best outcomes on both the validation and test sets.

\begin{table}[H]
\caption{Maximum validation \gls{mIoU} and corresponding test \gls{mIoU} scores of the tested semi-supervised models trained on 10\% of labeled training images across probability thresholds ranging from 0.5 to 0.9.}
\label{table:finding_best_threshold_max_val_miou_10}
\centering
\begin{tabular}{lll}
    \toprule
    Threshold & Max Val mIoU & Test mIoU \\ 
    \midrule
    0.5      & 0.68         & 0.57      \\ 
    0.6       & 0.72         & 0.59      \\ 
    0.7       & 0.70         & 0.57      \\ 
    0.8       & 0.70         & 0.54      \\ 
    0.9       & \textbf{0.74}         & \textbf{0.63}      \\ 
    \bottomrule
\end{tabular}
\end{table}

\begin{figure}[H]
    \centering
    \includegraphics[width=\textwidth]{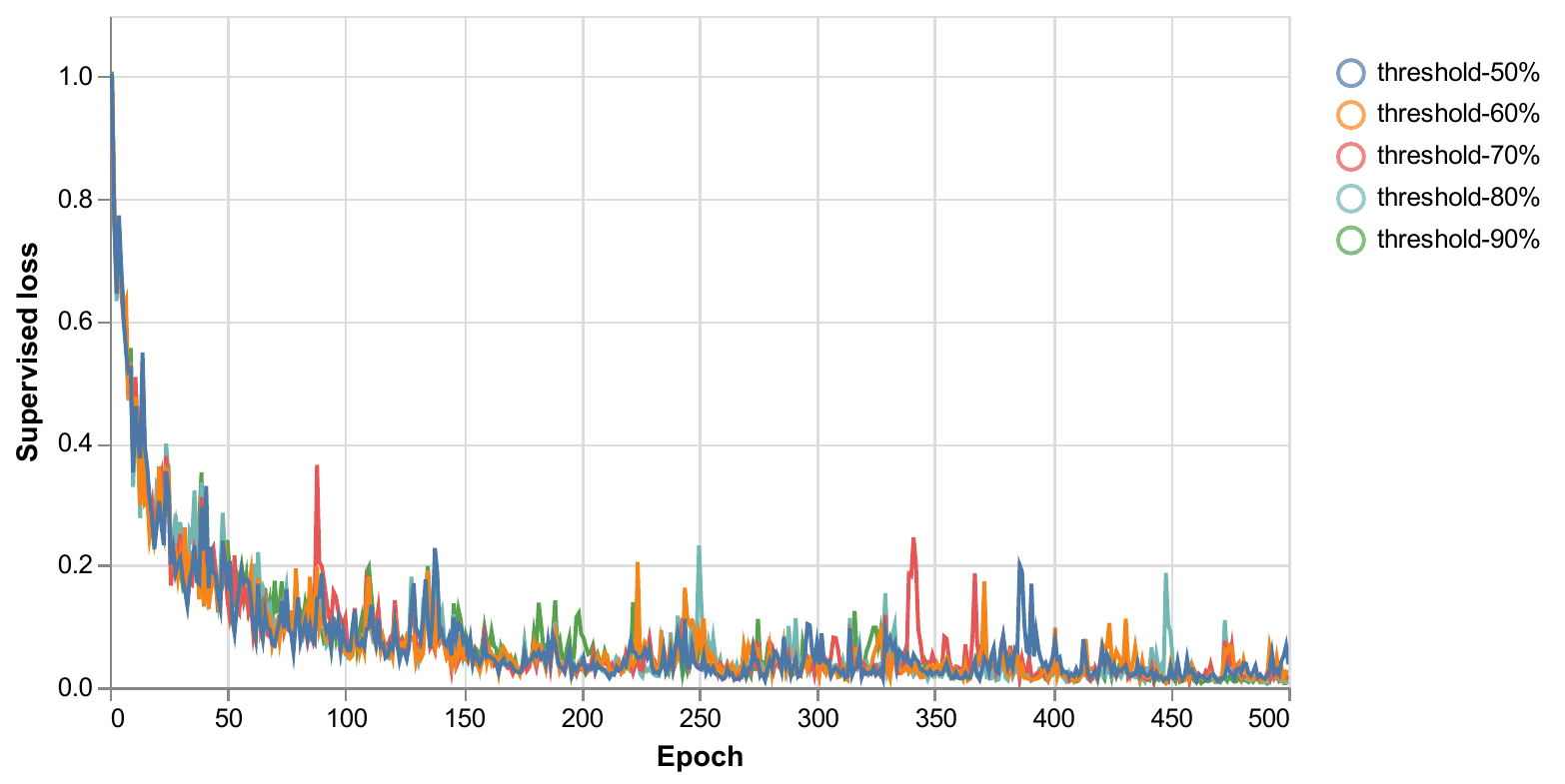}
    \caption{Supervised component of the training loss of the tested semi-supervised models trained on 10\% of labeled training images for probability thresholds ranging from 0.5 to 0.9.}
    \label{Figure:finding_best_threshold_sup_loss_prob_thresholds}
\end{figure}

\begin{figure}[H]
    \centering
    \includegraphics[width=\textwidth]{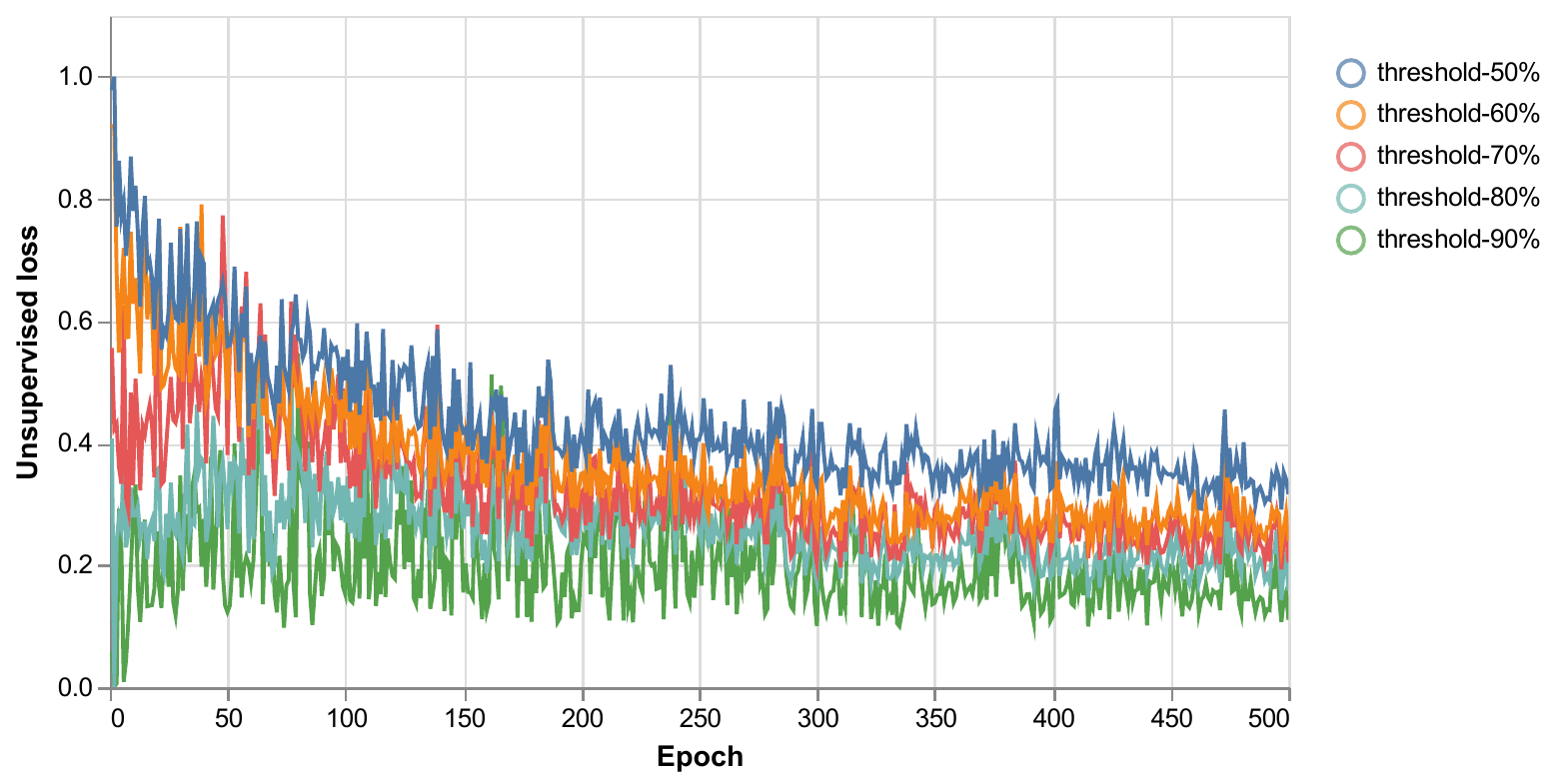}
    \caption{Unsupervised component of the training loss of the tested semi-supervised models trained on 10\% of labeled training images for probability thresholds ranging from 0.5 to 0.9.}
    \label{Figure:finding_best_threshold_unsup_loss_prob_thresholds}
\end{figure}

Figures \ref{Figure:finding_best_threshold_sup_loss_prob_thresholds}, and \ref{Figure:finding_best_threshold_unsup_loss_prob_thresholds} respectively show the supervised component ($\mathcal{L}^s$, Eq. \ref{eq:Ls_fixmatch_sem_seg}) and the unsupervised component ($\mathcal{L}^u$, Eq. \ref{eq:Lu_fixmatch_sem_seg}) of the total loss of the tested semi-supervised models across different probability thresholds.

The trend of the supervised component of the loss remains relatively consistent across different probability thresholds. On the other hand, the unsupervised component’s trend varies significantly with the threshold. In fact, Figure \ref{Figure:finding_best_threshold_unsup_loss_prob_thresholds} shows that the unsupervised component of the loss is lower with a higher threshold, especially at the beginning of training.

\subsection{Cross Entropy Vs Focal loss in the Unsupervised Component}
\label{subsec:res-ce-vs-focal-loss}

The hypothesis of the third experiment claims: “Employing the cross entropy loss as the unsupervised component in FixMatch for Semantic Segmentation results in higher \gls{mIoU} scores on the test set, as opposed to using the focal loss, particularly when using a high probability threshold”.

Table \ref{table:mIoU_exp_ce_vs_focal} compares semi-supervised models, which are trained using either \first cross entropy loss or \second focal loss for the unsupervised component ($\mathcal{L}^u$, Eq. \ref{eq:Lu_fixmatch_sem_seg}) of the total loss ($\mathcal{L} = \mathcal{L}^s + \lambda \cdot \mathcal{L}^u$). As reported in Table \ref{table:hyperparams_ssl_one_train_set}, the $\gamma$ coefficient of the focal loss is set to 2.0. Each test \gls{mIoU} score is averaged over two independently trained models.

\begin{table}[H]
\caption{Test \gls{mIoU} of semi-supervised models trained with cross entropy versus focal loss in the unsupervised component of the loss ($\mathcal{L}^u$, Eq. \ref{eq:Lu_fixmatch_sem_seg}) using 10\% of labeled data.}
\label{table:mIoU_exp_ce_vs_focal}
\centering
\begin{tabular}{ll}
    \toprule
    Unsupervised Loss & Test mIoU                 \\  
    \midrule
    Cross entropy     & 0.643   \\ 
    Focal loss             & 0.642    \\ 
    \bottomrule
\end{tabular}
\end{table}

\begin{figure}[H]
    \centering
    \includegraphics[width=\textwidth]{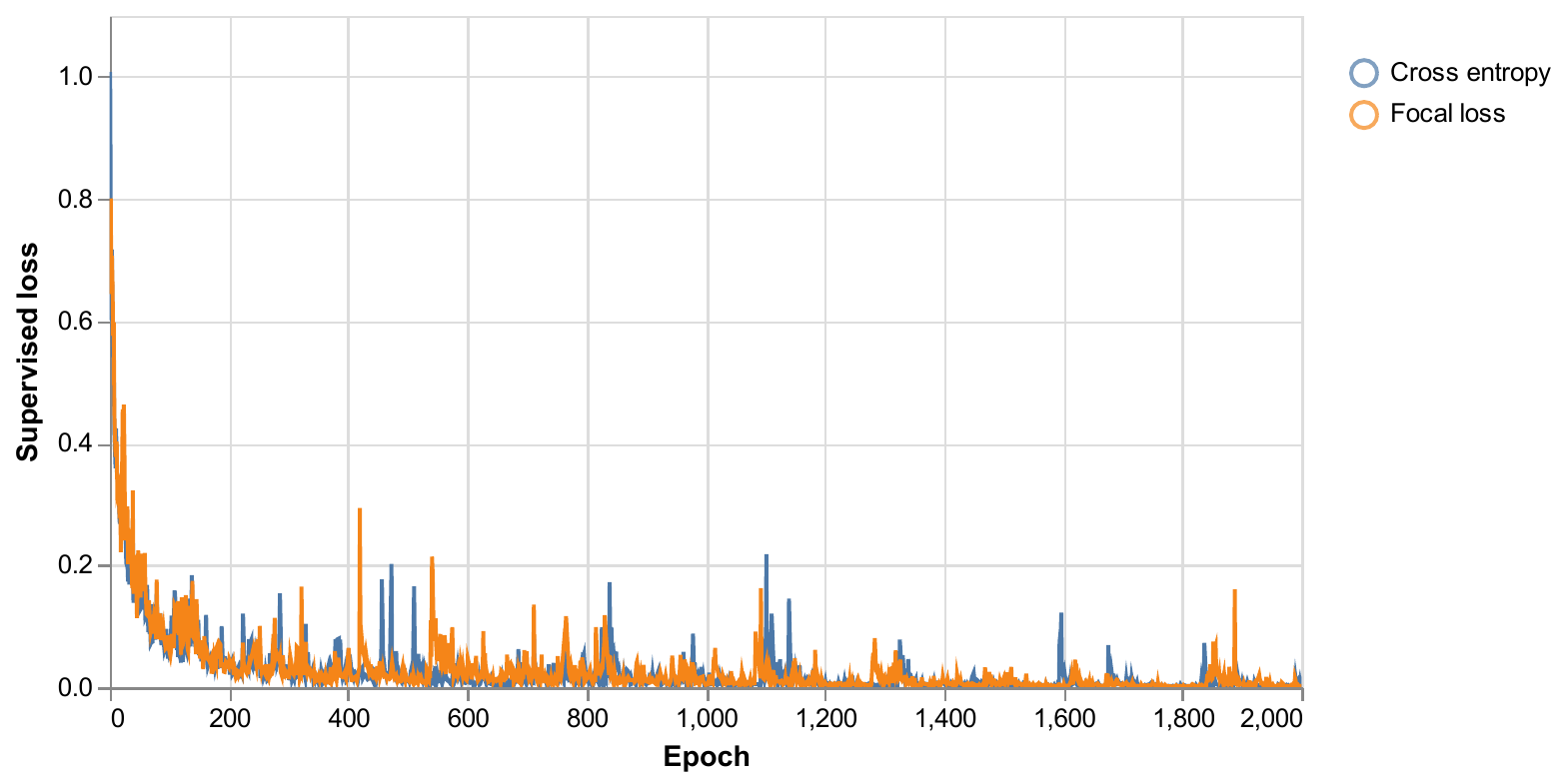}
    \caption{Supervised component of the training loss of the tested semi-supervised models. The models were trained using either cross entropy or focal loss as the unsupervised component of the loss. The training was conducted on 10\% of the labeled training images.}
    \label{Figure:ce_vs_focal_sup_loss}
\end{figure}

\begin{figure}[H]
    \centering
    \includegraphics[width=\textwidth]{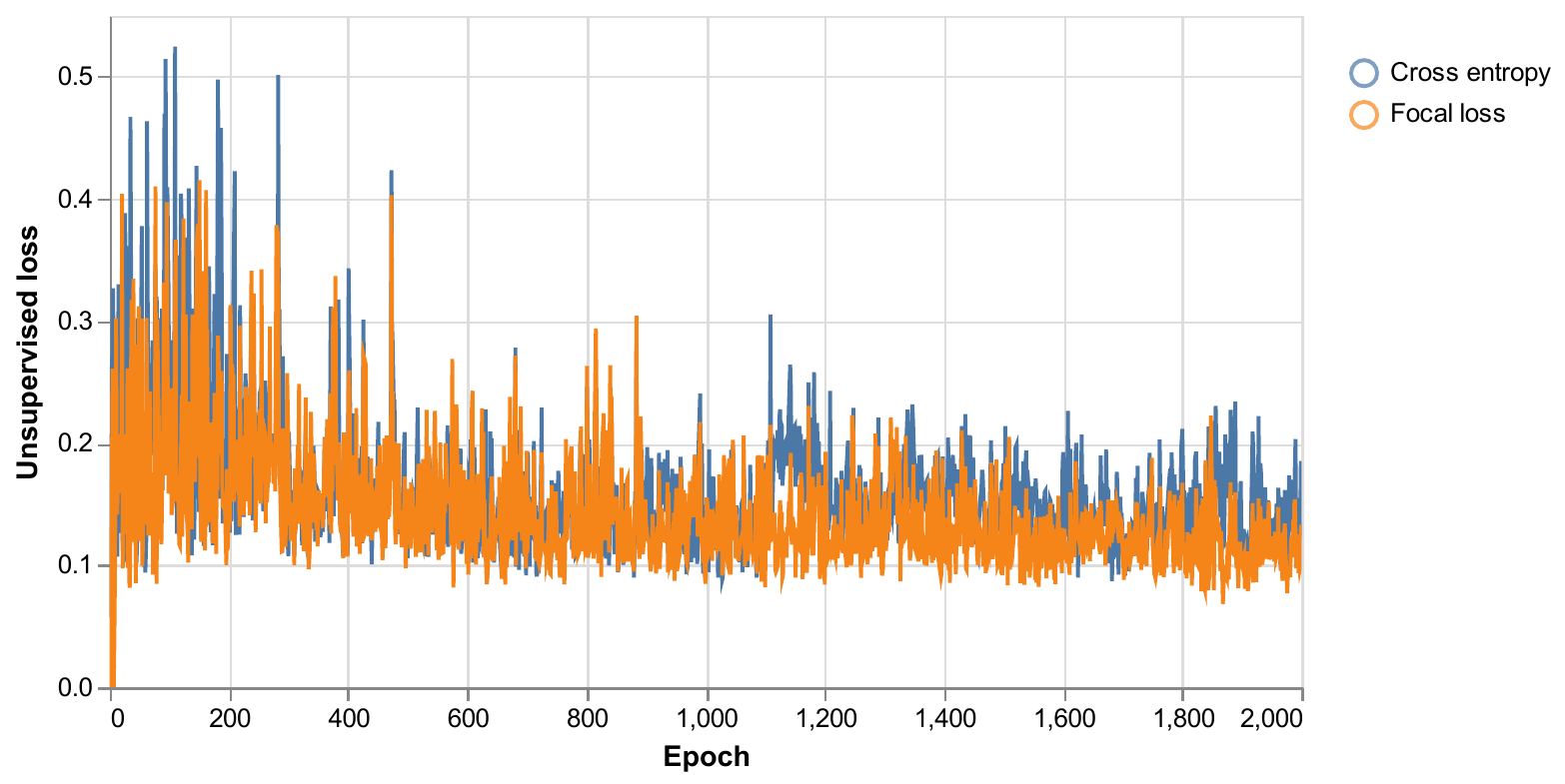}
    \caption{Unsupervised component of the training loss of the tested semi-supervised models. The models were trained using either cross entropy or focal loss as the unsupervised component of the loss. The training was conducted on 10\% of the labeled training images.}
    \label{Figure:ce_vs_focal_unsup_loss}
\end{figure}

Figures \ref{Figure:ce_vs_focal_sup_loss}, and \ref{Figure:ce_vs_focal_unsup_loss} respectively depict the supervised ($\mathcal{L}^s$, Eq. \ref{eq:Ls_fixmatch_sem_seg}) and the unsupervised ($\mathcal{L}^u$, Eq. \ref{eq:Lu_fixmatch_sem_seg}) components of the total loss of the evaluated semi-supervised models. These models utilized different unsupervised losses, either cross entropy or focal loss. Interestingly, both the unsupervised and supervised losses exhibited a similar trend, despite the variation in the type of the unsupervised loss.

\section{Main Experiment - Comparing Semi-Supervised and Fully-Supervised Learning Models}
\label{subsec:res-fs-vs-ssl-2-train-sets}

This section contains the results of the main experiment of our work (refer to Section \ref{subsec:exp_design_exp_2_train_sets} for the details of the methodology of this experiment).
Specifically, Subsection \ref{subsec:perf-comp-ssl-vs-fs-res} compares the performance of fully-supervised and semi-supervised learning models, while a visual comparison of the predictions of semi-supervised and fully-supervised model is contained Subsection \ref{subsec:viz-ssl-vs-fs-res}.

\subsection{Comparing the Performance of Semi-Supervised and Fully-Supervised Learning Models}
\label{subsec:perf-comp-ssl-vs-fs-res}

This Subsection compares the performance of semi-supervised and fully-supervised models. Moreover, Subsubsection \ref{subsubsec:connection-val-loss-miou} outlines a connection between the validation loss and \gls{mIoU} that we observed when semi- and fully-supervised models have the same performance on the validation set but not on the test set.

Tables \ref{table:mIoU-fully-vs-ssl-val} and \ref{table:mIoU-fully-vs-ssl} respectively compare semi-supervised and fully-supervised models when varying the amount of labeled data in the validation and test sets. All models were trained with the "Two Training Sets" (see Subsection \ref{subsec:two-train-set}) setting. 

With low percentages of labeled data (i.e., $\leq30\%$), the semi-supervised models consistently outperform or match the \gls{mIoU} scores of fully-supervised models on both validation and test sets. However, when the labeled data increases to 40\% or more, fully-supervised models tend to achieve higher \gls{mIoU} scores on the test set in four out of five instances, with the exception being at 60\% labeled data.

The performance gap between fully- and semi-supervised models is more pronounced when the labeled data is less than or equal to 30\%. In these cases, semi-supervised models average a 2.25\% higher \gls{mIoU} on the validation set and a 6.50\% higher \gls{mIoU} on the test set compared to their fully-supervised counterparts.

On the other hand, when the labeled data is 40\% or more, fully-supervised models are on average better. If we exclude the outlier at 60\%, they outperform the semi-supervised models by 1.33\% on the validation set and by 3\% on the test set. However, semi- and fully-supervised models perform more similarly as the percentage of labeled data increases.

\begin{table}[H]
\caption{Validation \gls{mIoU} scores of fully- and semi-supervised models trained with the "Two Training Sets" setting and with diverse percentages of labeled data.}
\label{table:mIoU-fully-vs-ssl-val}
\centering
\begin{adjustbox}{width=\columnwidth,center}
\begin{tabular}{llllllllll}
    \toprule
    \multirow{2}{*}{Method} & \multicolumn{9}{c}{Labeled data (\%)}                                                                                                                                                                                                              \\ \cmidrule{2-10} 
                            & \multicolumn{1}{l}{5}   & \multicolumn{1}{l}{10}   & \multicolumn{1}{l}{20}   & \multicolumn{1}{l}{30}        & \multicolumn{1}{l}{40}   & \multicolumn{1}{l}{50}   & \multicolumn{1}{l}{60}   & \multicolumn{1}{l}{70}   & 80   \\ 
                            \midrule
    Fully-supervised        & \multicolumn{1}{l}{0.50} & \multicolumn{1}{l}{0.72} & \multicolumn{1}{l}{0.78} & \multicolumn{1}{l}{0.89}      & \multicolumn{1}{l}{0.87} & \multicolumn{1}{l}{\textbf{0.89}} & \multicolumn{1}{l}{0.90} & \multicolumn{1}{l}{0.91} & \textbf{0.92} \\ 
    Semi-supervised         & \multicolumn{1}{l}{\textbf{0.53}}     & \multicolumn{1}{l}{\textbf{0.76}}     & \multicolumn{1}{l}{\textbf{0.80}}     & \multicolumn{1}{l}{0.89} & \multicolumn{1}{l}{\textbf{0.88}}     & \multicolumn{1}{l}{0.86}     & \multicolumn{1}{l}{\textbf{0.91}}     & \multicolumn{1}{l}{0.91}     &  0.91    \\ 
    \bottomrule
\end{tabular}
\end{adjustbox}
\end{table}

\begin{table}[H]
\caption{Test \gls{mIoU} scores of fully- and semi-supervised models trained with the "Two Training Sets" setting and with diverse percentages of labeled data.}
\label{table:mIoU-fully-vs-ssl}
\centering
\begin{adjustbox}{width=\columnwidth,center}
\begin{tabular}{llllllllll}
    \toprule
    \multirow{2}{*}{Method} & \multicolumn{9}{c}{Labeled data (\%)}                                                                                                                                                                                                              \\ \cmidrule{2-10} 
                            & \multicolumn{1}{l}{5}   & \multicolumn{1}{l}{10}   & \multicolumn{1}{l}{20}   & \multicolumn{1}{l}{30}        & \multicolumn{1}{l}{40}   & \multicolumn{1}{l}{50}   & \multicolumn{1}{l}{60}   & \multicolumn{1}{l}{70}   & 80   \\ 
                            \midrule
    Fully-supervised        & \multicolumn{1}{l}{0.46} & \multicolumn{1}{l}{0.61} & \multicolumn{1}{l}{0.72} & \multicolumn{1}{l}{0.86}      & \multicolumn{1}{l}{\textbf{0.89}} & \multicolumn{1}{l}{\textbf{0.86}} & \multicolumn{1}{l}{0.88} & \multicolumn{1}{l}{\textbf{0.88}} & \textbf{0.91} \\
    Semi-supervised         & \multicolumn{1}{l}{\textbf{0.57}}     & \multicolumn{1}{l}{\textbf{0.65}}     & \multicolumn{1}{l}{\textbf{0.80}}     & \multicolumn{1}{l}{\textbf{0.89}} & \multicolumn{1}{l}{0.86}     & \multicolumn{1}{l}{0.82}     & \multicolumn{1}{l}{\textbf{0.90}}     & \multicolumn{1}{l}{0.84}     &  0.90    \\ 
    \bottomrule
\end{tabular}
\end{adjustbox}
\end{table}

\begin{figure}[H]
    \centering
    \includegraphics[width=\textwidth]{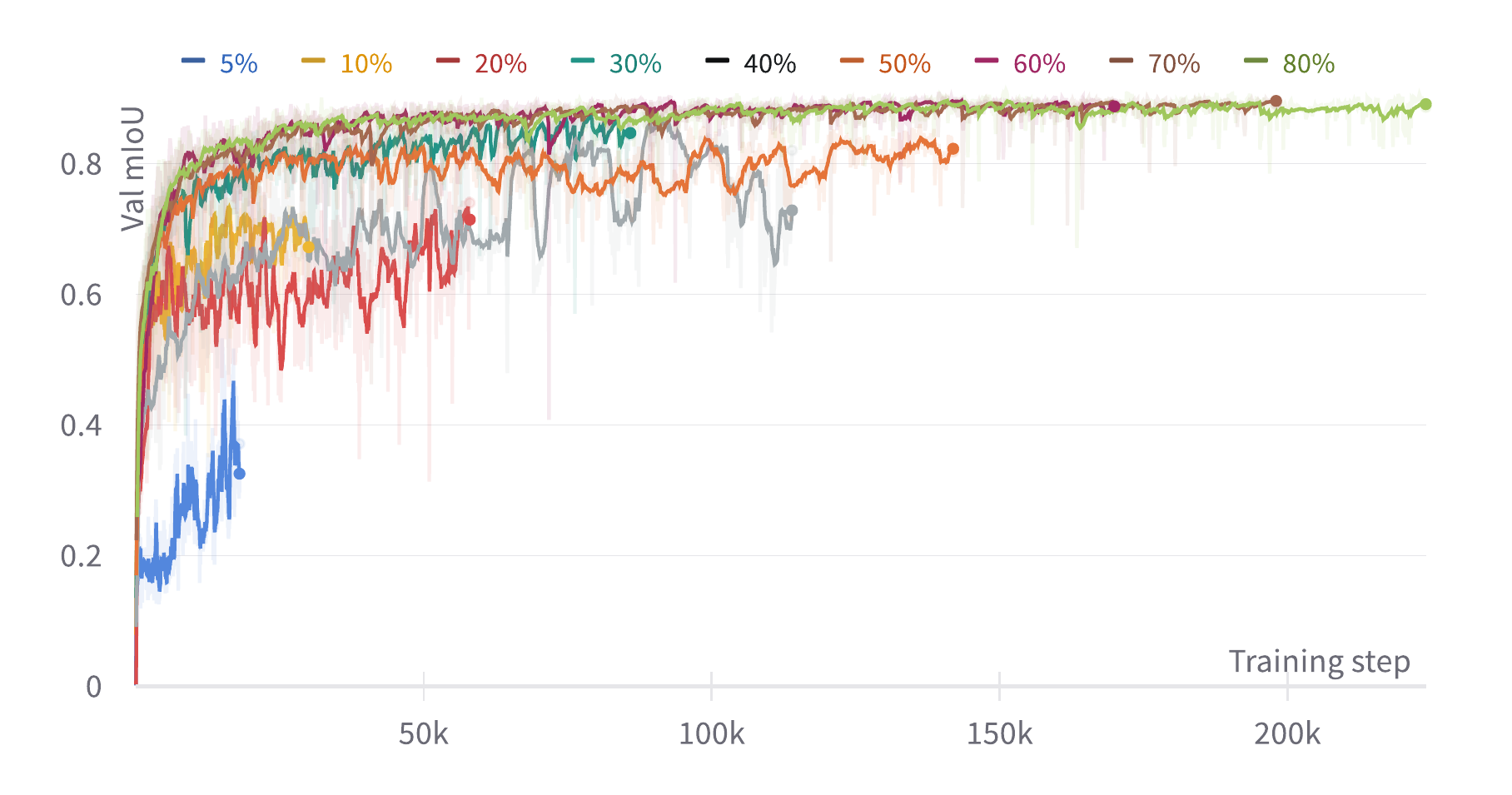}
    \caption{Validation \gls{mIoU} of semi-supervised models across the training steps with all percentages of labeled data. The y-axis represents the validation \gls{mIoU} scores, while the x-axis represents the number of training steps. The legend indicates the percentage of labeled data used for training. An exponential moving average with coefficient of 0.8 was applied to smooth the lines of the chart.}
    \label{Figure:val_mIoU_2_train_sets_ssl_models_all_percentages}
\end{figure}

The graph depicted in Figure \ref{Figure:val_mIoU_2_train_sets_ssl_models_all_percentages} illustrates the trend of validation \gls{mIoU} scores for semi-supervised models trained with the varying percentages of labeled data. As the percentage of labeled data increases, so does the \gls{mIoU}. The lowest \gls{mIoU} is observed when training with 5\% of labeled data, while the highest \gls{mIoU} scores are seen when training with 60\%, 70\%, and 80\% of labeled data.

Interestingly, there appears to be a plateau in the performance of semi-supervised models when the percentage of labeled data increases from 60\% to 80\%. This observation is also reflected in Tables \ref{table:mIoU-fully-vs-ssl-val} and \ref{table:mIoU-fully-vs-ssl}. In contrast, fully-supervised models exhibit a performance boost of 2\% and 3\% on the validation and test \gls{mIoU} scores respectively.

\subsubsection{A Connection Between Validation Loss and mIoU}
\label{subsubsec:connection-val-loss-miou}

Looking at Table \ref{table:mIoU-fully-vs-ssl-val}, it can be noticed that the fully-supervised and semi-supervised models have the same performance on the validation set when using 30\% and 70\% of labeled data. However, a divergence in performance is observed on the test set, as detailed in Table \ref{table:mIoU-fully-vs-ssl}. Specifically, the semi-supervised test \gls{mIoU} is higher when trained with 30\% of labeled data, whereas the fully-supervised model's test \gls{mIoU} is greater when trained with 70\% of labeled data. Consequently, we chose to delve deeper into these two particular scenarios.

\begin{figure}[H]
    \centering
    \begin{subfigure}[t]{\textwidth}
    \centering
        \raisebox{-\height}{\includegraphics[width=0.44\textwidth]{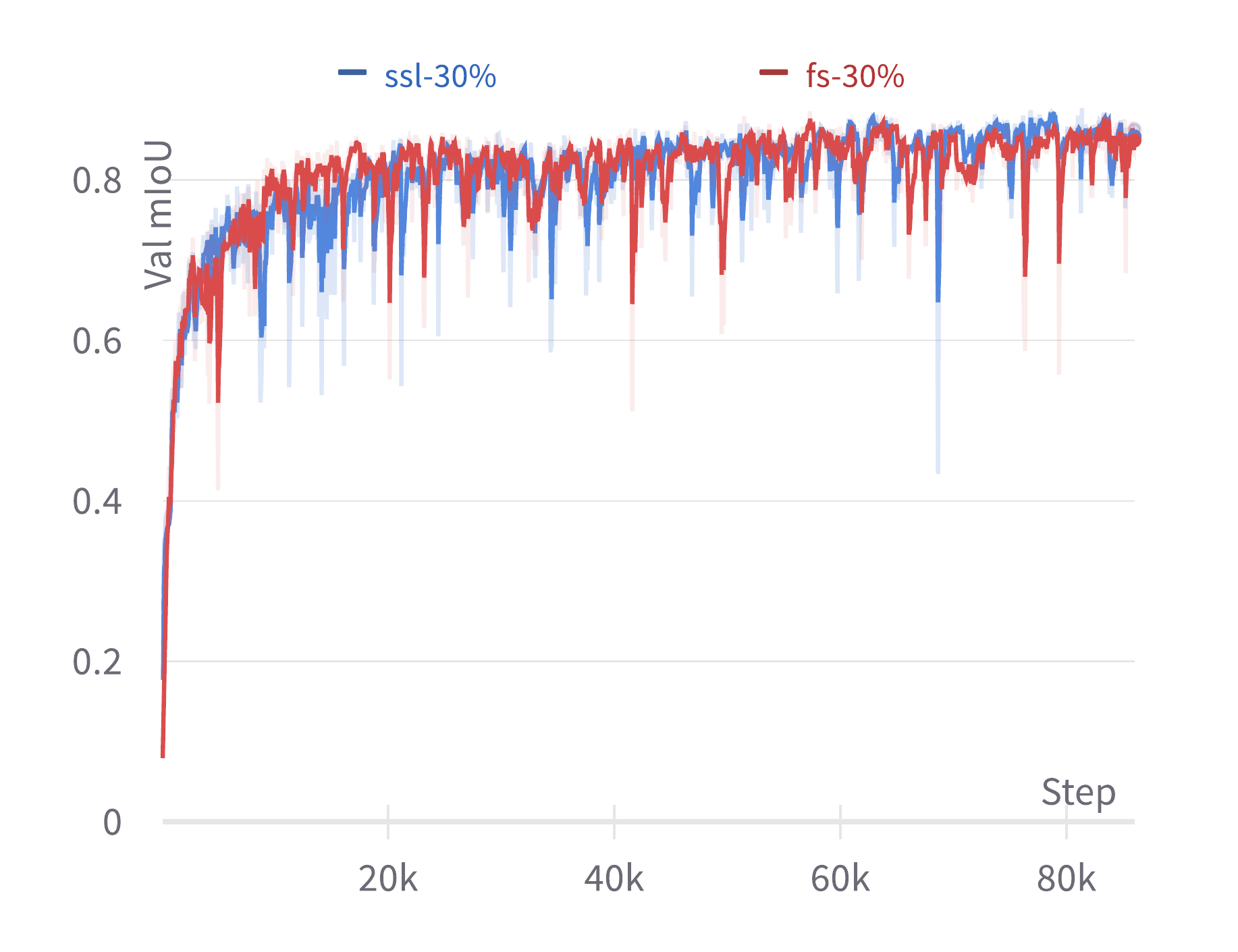}}
        \raisebox{-\height}{\includegraphics[width=0.44\textwidth]{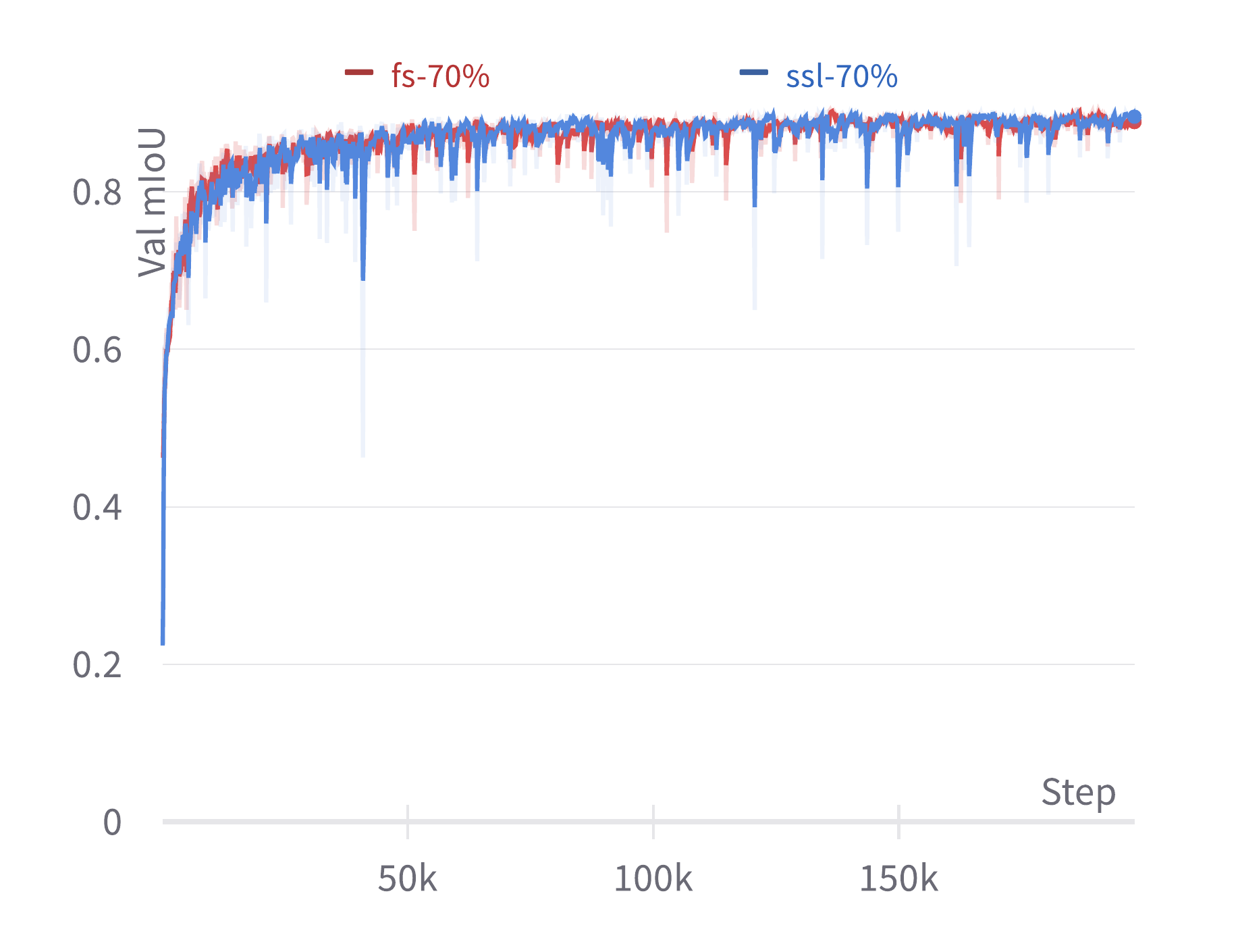}}
        \raisebox{-\height}{\includegraphics[width=0.44\textwidth]{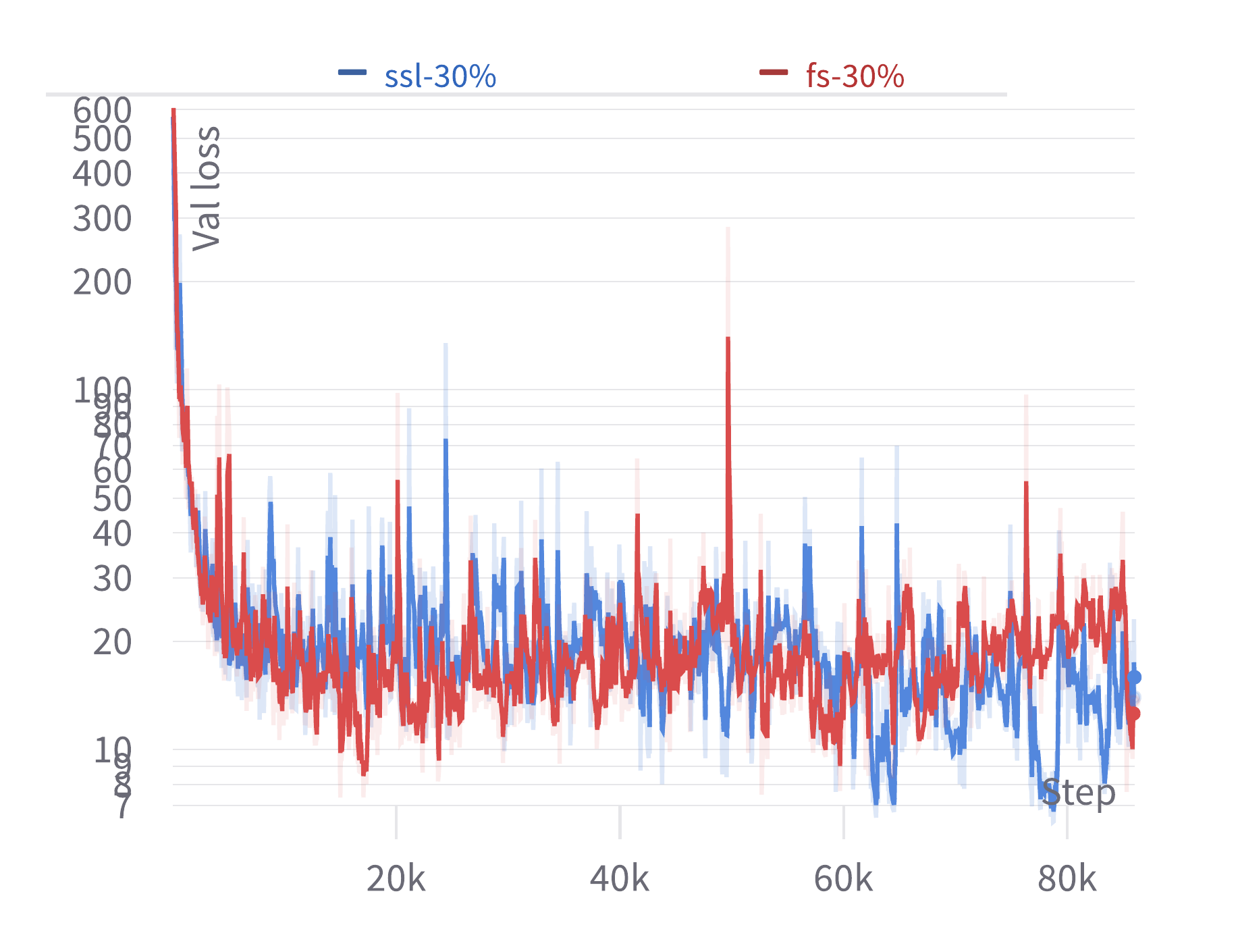}}
        \raisebox{-\height}{\includegraphics[width=0.44\textwidth]{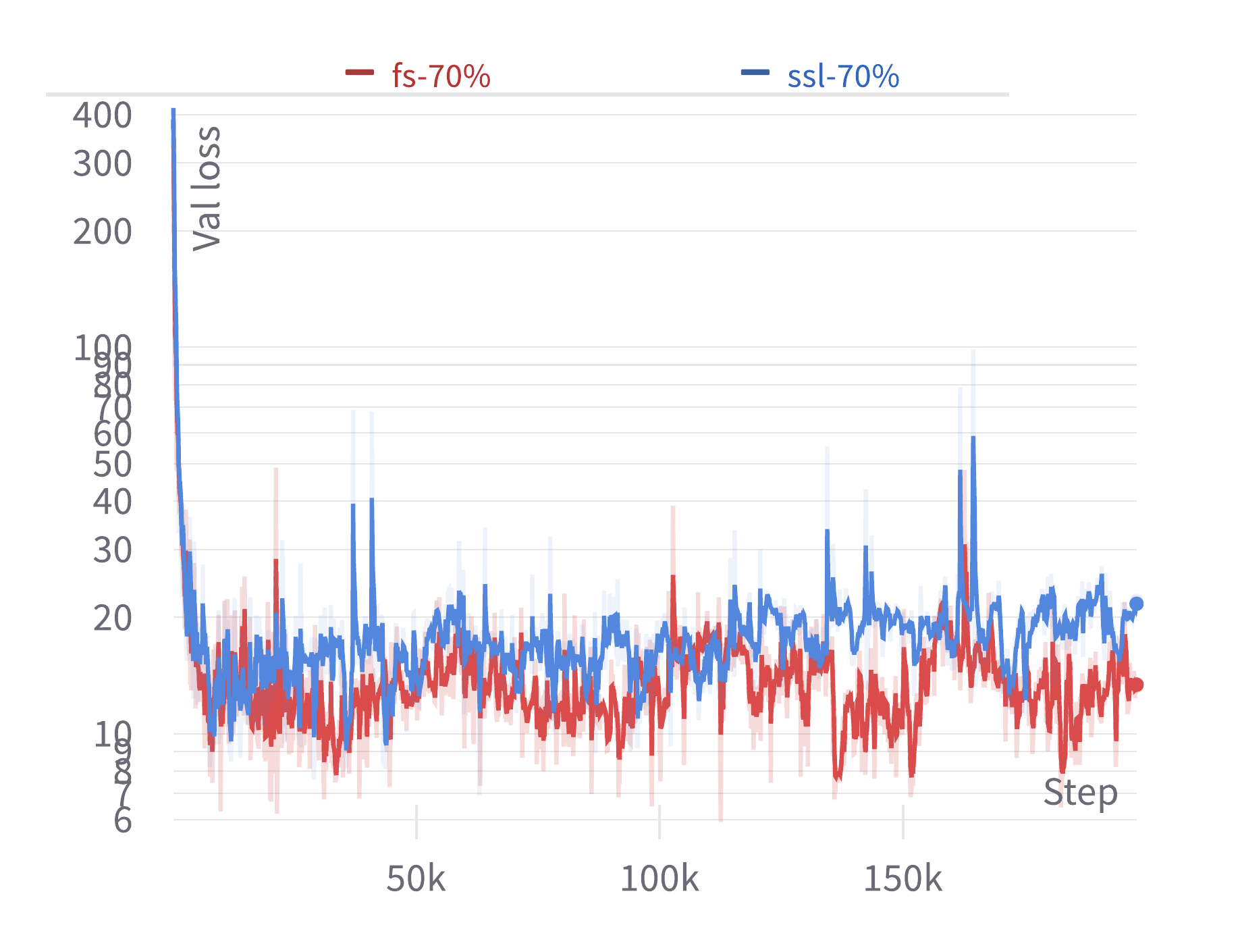}}
    \end{subfigure}
    \hfill
    \caption{Validation \gls{mIoU} and validation loss across the training steps with 30\% and 70\% of labeled data. The x-axis represents the number of training steps. The legend indicates the percentage of labeled data used for training and the type of training: fully-supervised (fs) or semi-supervised (ssl). Log scale is applied to the y-axis of the validation loss charts. An exponential moving average with coeﬀicient of 0.3 was applied to smooth the lines of all the four charts.}
    \label{Figure:val_miou_and_loss_of_30_and_70}
\end{figure}

Figure \ref{Figure:val_miou_and_loss_of_30_and_70} shows the trends of the validation \gls{mIoU} and validation loss of the fully-supervised and semi-supervised models when trained with 30\% and 70\% of labeled data. The validation \gls{mIoU} trends are very similar for the fully-supervised and semi-supervised models, in both percentages of labeled data. However, for both 30\% and 70\%, the model with the highest test \gls{mIoU} has the lower validation loss. Specifically, the semi-supervised model demonstrates a lower validation loss when trained with 30\% of labeled data, while the fully-supervised model shows a lower validation loss when trained with 70\% of labeled data.

This connection between validation loss and \gls{mIoU} is observed in these two cases where the performance on the validation set was similar, but not the one on the test set.
However, the validation loss trends of models trained with other percentages of labeled data and having different types of performance did not exhibit this same distinct pattern.

\subsection{Visually Comparing Predictions}
\label{subsec:viz-ssl-vs-fs-res}

Upon observing the trends in validation \gls{mIoU} and loss, we decided to also compare fully- and semi-supervised models by visually inspecting their predictions on the test set.

However, since there are five classes in the dataset, and given our time constraints, we opted to visually compare the predictions of fully- and semi-supervised models only for a subset of classes. We will now delineate how we chose the subset of classes whose models' predictions were visually inspected.

Firstly, we looked at the differences between the scores of fully- and semi-supervised models for each class.
Table \ref{table:mIoU-difference-classes-tw-train-sets-ssl-vs-fs} contains these differences, in terms of test \gls{IoU}, for each percentage of labeled data utilized for training.

\begin{table}[H]
\caption{Differences in \gls{IoU} scores for semi-supervised and fully-supervised models on the test set across different classes and percentages of labeled data. These differences are presented as percentages. The highest and lowest difference of each class are highlighted.}
\label{table:mIoU-difference-classes-tw-train-sets-ssl-vs-fs}
\centering
\begin{tabular}{llllllllll}
    \toprule
    \multirow{2}{*}{Semi-supervised} & \multicolumn{9}{c}{Labeled data (\%)}                                                                                                                                                                                    \\ \cmidrule{2-10} 
                                     & \multicolumn{1}{r}{5}   & \multicolumn{1}{r}{10}  & \multicolumn{1}{r}{20}  & \multicolumn{1}{r}{30} & \multicolumn{1}{r}{40}  & \multicolumn{1}{r}{50}  & \multicolumn{1}{r}{60} & \multicolumn{1}{r}{70}  & 80 \\ 
                                     \midrule
    Marine Debris                    & \multicolumn{1}{r}{\textbf{+10}} & \multicolumn{1}{r}{-3}  & \multicolumn{1}{r}{+6}  & \multicolumn{1}{r}{+7} & \multicolumn{1}{r}{\textbf{-15}} & \multicolumn{1}{r}{+4}  & \multicolumn{1}{r}{+6} & \multicolumn{1}{r}{-2}  & -3 \\ 
    Algae/Organic Material           & \multicolumn{1}{r}{\textbf{+9}}  & \multicolumn{1}{r}{+4}   & \multicolumn{1}{r}{+4}  & \multicolumn{1}{r}{-1} & \multicolumn{1}{r}{0}   & \multicolumn{1}{r}{\textbf{-2}}  & \multicolumn{1}{r}{+1} & \multicolumn{1}{r}{+1}  & -1 \\ 
    Ship                             & \multicolumn{1}{r}{+3}  & \multicolumn{1}{r}{+5} & \multicolumn{1}{r}{\textbf{+7}}  & \multicolumn{1}{r}{+3} & \multicolumn{1}{r}{+3}  & \multicolumn{1}{r}{\textbf{-6}}  & \multicolumn{1}{r}{+4} & \multicolumn{1}{r}{0}   & -2 \\ 
    Cloud                            & \multicolumn{1}{r}{\textbf{+24}} & \multicolumn{1}{r}{+14}  & \multicolumn{1}{r}{+22} & \multicolumn{1}{r}{+4} & \multicolumn{1}{r}{+1}  & \multicolumn{1}{r}{-12} & \multicolumn{1}{r}{-1} & \multicolumn{1}{r}{\textbf{-17}} & +1 \\ 
    Water                            & \multicolumn{1}{r}{\textbf{+8}}  & \multicolumn{1}{r}{+3}  & \multicolumn{1}{r}{+4}  & \multicolumn{1}{r}{+1} & \multicolumn{1}{r}{0}   & \multicolumn{1}{r}{-2}  & \multicolumn{1}{r}{0}  & \multicolumn{1}{r}{\textbf{-3}}  & \multicolumn{1}{r}{0}  \\ \bottomrule
\end{tabular}
\end{table}

After looking at Table \ref{table:mIoU-difference-classes-tw-train-sets-ssl-vs-fs}, we noticed that some classes presented higher differences in \gls{IoU} scores. Thus, we chose to visualize the models' predictions of the classes that exhibited the greatest variations in the differences of Table \ref{table:mIoU-difference-classes-tw-train-sets-ssl-vs-fs}.

To identify the classes with the most significant variations, we computed the variance of the scores' differences between the fully-supervised and semi-supervised models for each class (i.e., the variance of each row of Table \ref{table:mIoU-difference-classes-tw-train-sets-ssl-vs-fs}). To do so, we needed the mean of the differences (among the varying percentages of labeled data) of each class. However, we purposely set the mean to zero ($\bar{x} = 0$, corresponding to the scenario where the two models exhibit identical performance) so that a higher difference (positive or negative) between a fully-supervised and a semi-supervised model contributes to a higher variance.

Thus, the variance of the scores' differences between the fully-supervised and semi-supervised models for class $c$ was calculated as follows:
\begin{equation}
    Var(c) = \frac{1}{n-1} \sum_{i=1}^{n} (x_i - \bar{x})^2  
\end{equation}
where $c$ is the class we are analyzing, $n = 9$ is the number of diverse percentages of labeled data (i.e. $5, 10, 20, \dots 80$), and $x_i$ is the difference of performance in terms of \gls{IoU} for the class $c$ between the semi-supervised and the fully-supervised models trained with the percentage of labeled data corresponding to index $i$ ($i = 1$ corresponds to 5\% of labeled data, $i = 2$ to 10\%, $i = 3$ to 20\%, and so on, up to $i = 9$ corresponding to 80\%). 

The variances of the scores' differences of the classes, ranked in descending order, are as follows: "Cloud" with a variance of $213.50$, "Marine Debris" with $60.50$, "Ship" with $19.63$, "Algae/Organic Material" with $15.13$, and "Water" with $12.88$.

Thus, we decided to visually inspect the predictions of semi-supervised versus fully-supervised models of the "Cloud" and "Marine Debris" classes, as they are the ones that exhibited the highest variance.
Specifically, we compared the predictions of these classes by considering the lowest and/or highest cases of difference of performance between semi-supervised and fully-supervised models.

\subsubsection{Cloud}

Starting with the "Cloud" class, we examined the predictions made by both semi-supervised and fully-supervised models using 5\% and 70\% of labeled data. These percentages correspond to the maximum (+24\%) and minimum (-17\%) performance increase of semi-supervised models in comparison to fully-supervised models for the “Cloud” class.

\begin{figure}[H]
    \centering
    \begin{subfigure}[t]{\textwidth}
    \centering
        \raisebox{-\height}{\includegraphics[width=0.25\textwidth]{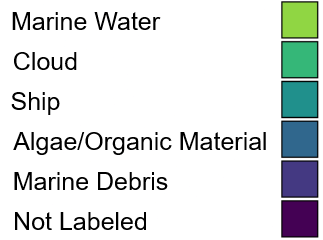}}
    
        \raisebox{-\height}{\includegraphics[width=0.24\textwidth]{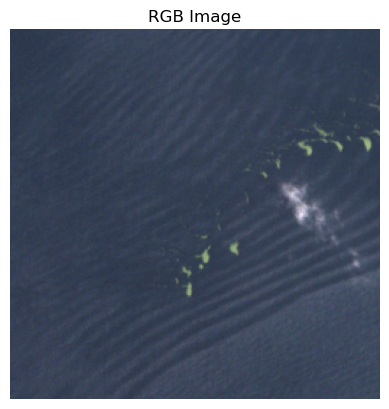}}
        \raisebox{-\height}{\includegraphics[width=0.24\textwidth]{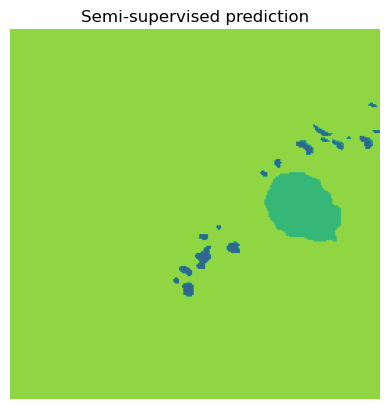}}
        \raisebox{-\height}{\includegraphics[width=0.24\textwidth]{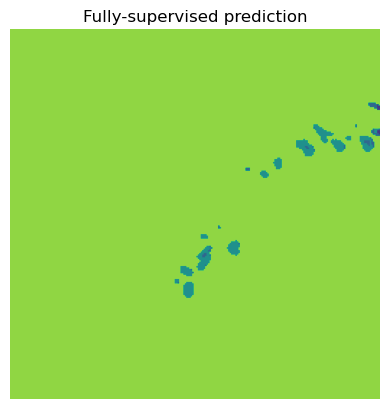}}
        \raisebox{-\height}{\includegraphics[width=0.24\textwidth]{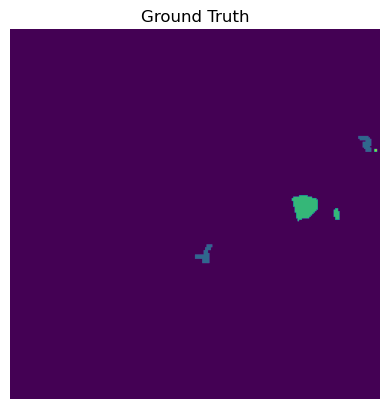}}

        \raisebox{-\height}{\includegraphics[width=0.24\textwidth]{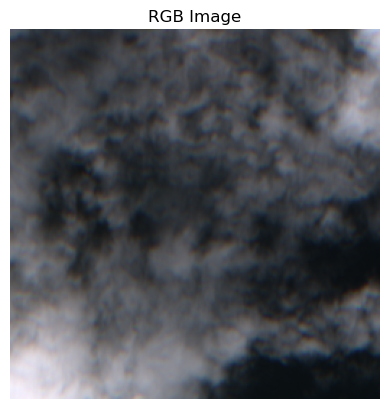}}
        \raisebox{-\height}{\includegraphics[width=0.24\textwidth]{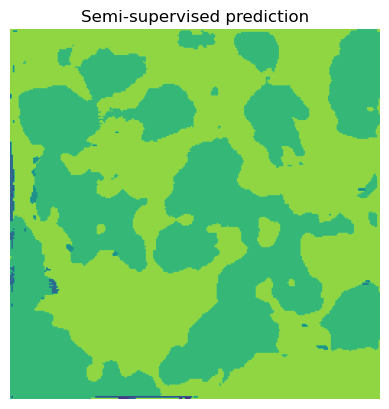}}
        \raisebox{-\height}{\includegraphics[width=0.24\textwidth]{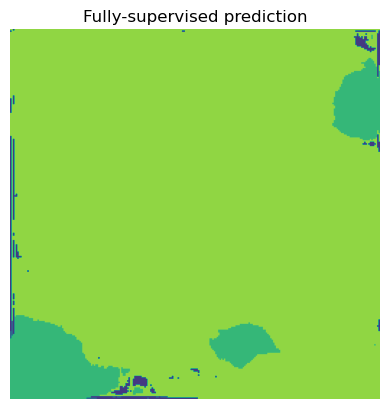}}
        \raisebox{-\height}{\includegraphics[width=0.24\textwidth]{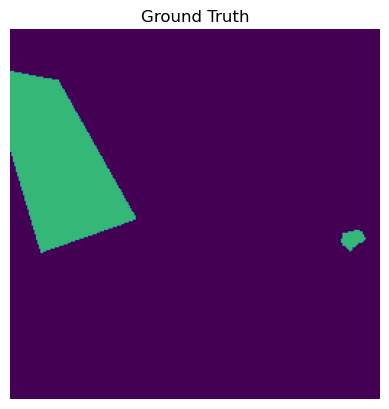}}
    
        \raisebox{-\height}{\includegraphics[width=0.24\textwidth]{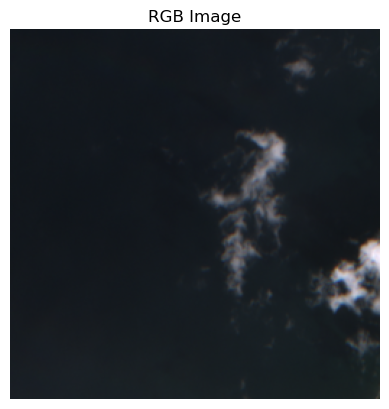}}
        \raisebox{-\height}{\includegraphics[width=0.24\textwidth]{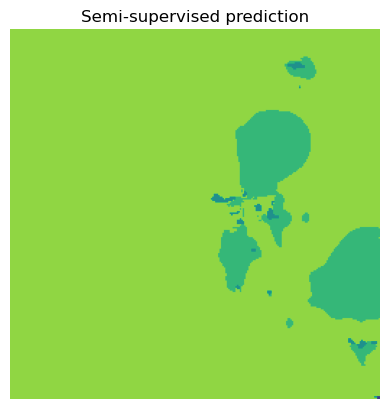}}
        \raisebox{-\height}{\includegraphics[width=0.24\textwidth]{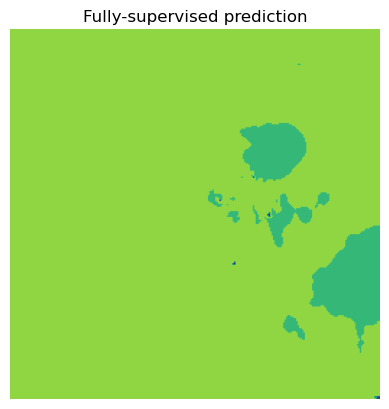}}
        \raisebox{-\height}{\includegraphics[width=0.24\textwidth]{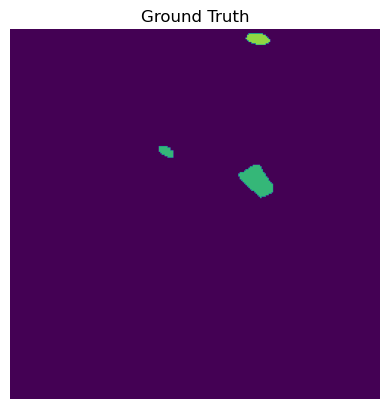}}
    \end{subfigure}
    \hfill
    \caption{Visualizing MARIDA patches, their semantic segmentation maps, and the predictions made by semi-supervised and fully-supervised models trained using 5\% of labeled data. Each row denotes a unique patch, while each column denotes the image type. The color legend is at the top of the figure.}
    \label{Figure:5_perc_labeled_data_clouds}
\end{figure}

We start by visually comparing some predictions made by the two models for the case of maximum performance in cloud detection by the semi-supervised model. Figure \ref{Figure:5_perc_labeled_data_clouds} shows three MARIDA patches containing clouds. Accompanying each patch are their corresponding semantic segmentation maps, along with the predictions of both semi-supervised and fully-supervised models trained using 5\% of labeled data.

The three patches of Figure \ref{Figure:5_perc_labeled_data_clouds} show that the semi-supervised model appears to outperform its fully-supervised counterpart in cloud detection, as it was noticed in Table \ref{table:mIoU-difference-classes-tw-train-sets-ssl-vs-fs}. This is evident in the first patch, where the semi-supervised model, albeit not flawlessly, identifies the cloud on the right side of the image, which was missed by the fully-supervised model.

\begin{figure}[H]
    \centering
    \begin{subfigure}[t]{\textwidth}
    \centering
        \raisebox{-\height}{\includegraphics[width=0.25\textwidth]{images/4_Results/2_train_sets_preds_viz/color_palette.png}}
    
        \raisebox{-\height}{\includegraphics[width=0.24\textwidth]{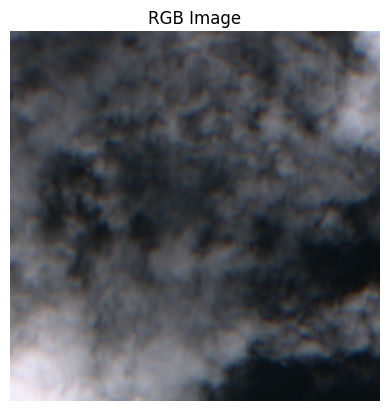}}
        \raisebox{-\height}{\includegraphics[width=0.24\textwidth]{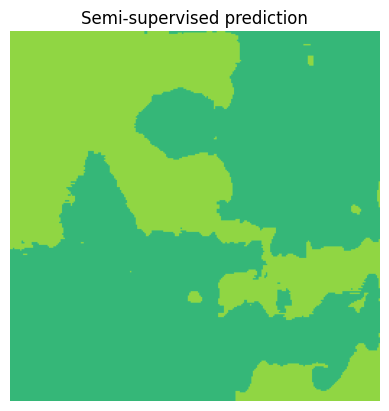}}
        \raisebox{-\height}{\includegraphics[width=0.24\textwidth]{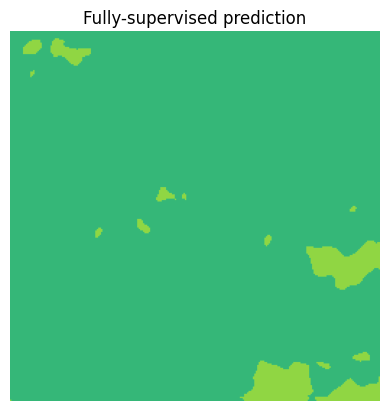}}
        \raisebox{-\height}{\includegraphics[width=0.24\textwidth]{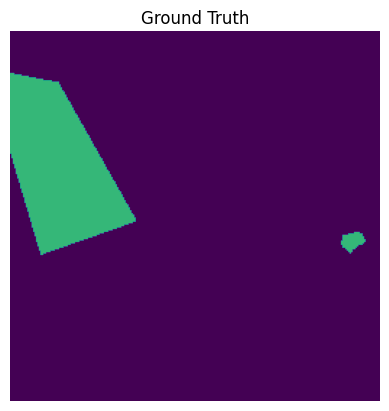}}

        \raisebox{-\height}{\includegraphics[width=0.24\textwidth]{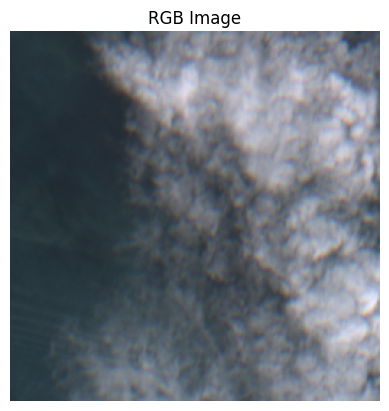}}
        \raisebox{-\height}{\includegraphics[width=0.24\textwidth]{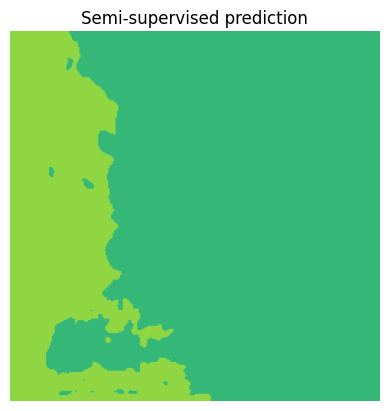}}
        \raisebox{-\height}{\includegraphics[width=0.24\textwidth]{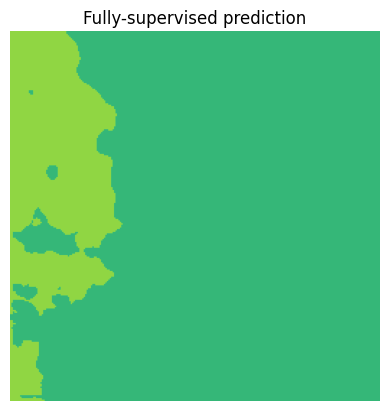}}
        \raisebox{-\height}{\includegraphics[width=0.24\textwidth]{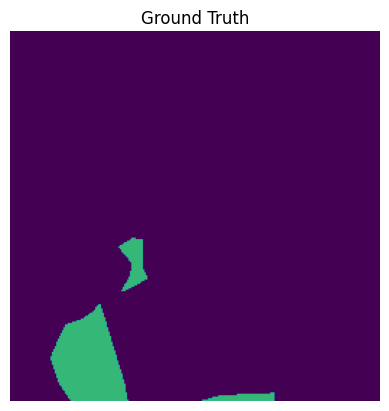}}

        \raisebox{-\height}{\includegraphics[width=0.24\textwidth]{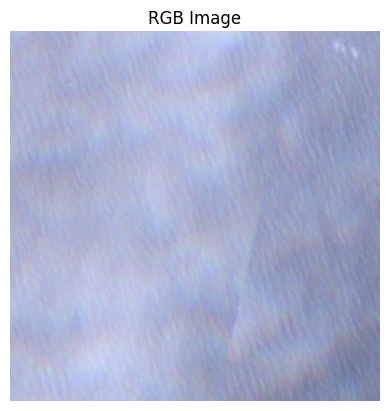}}
        \raisebox{-\height}{\includegraphics[width=0.24\textwidth]{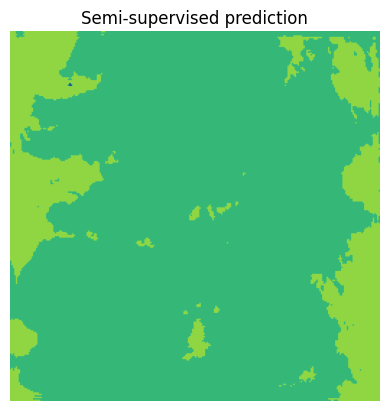}}
        \raisebox{-\height}{\includegraphics[width=0.24\textwidth]{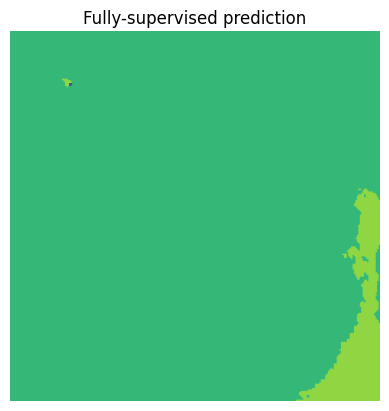}}
        \raisebox{-\height}{\includegraphics[width=0.24\textwidth]{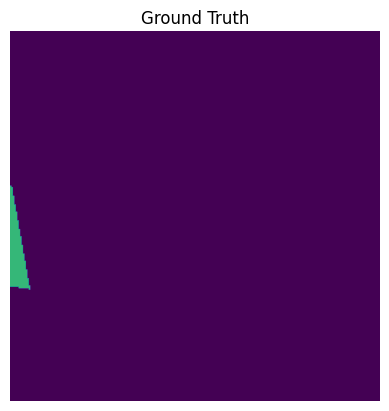}}      

    \end{subfigure}
    \hfill
    \caption{Visualizing some MARIDA patches from the test set, their semantic segmentation maps, and the predictions made by semi-supervised and fully-supervised models trained using 70\% of labeled data. Each row denotes a unique patch, while each column denotes the image type. The color legend is at the top of the figure.}
    \label{Figure:70_perc_labeled_data_clouds_majority_of_errors}
\end{figure}

In the second patch, the semi-supervised model detects a greater number of clouds than the fully-supervised model, especially in the central part of the image. We also noticed that the labeling of the second patch appears to be unusual. Specifically, a significant number of pixels that are dark in the RGB image, giving the impression of marine water, are instead labeled as "Cloud". Nonetheless, the semi-supervised model does not label several of these strangely labeled dark pixels in the central and right parts of the image as "Cloud". This patch is also depicted in Figure \ref{Figure:70_perc_labeled_data_clouds_majority_of_errors}. 

Lastly, in the third patch, the semi-supervised model detects smaller clouds situated in the center, upper-right, and lower-right sections of the image, while the fully-supervised model does not detect them.

We now consider the case where the semi-supervised model demonstrates its worst performance in cloud detection and proceed to visually compare the predictions made by both fully-supervised and semi-supervised models.

\begin{figure}[H]
    \centering
    \begin{subfigure}[t]{\textwidth}
    \centering
        \raisebox{-\height}{\includegraphics[width=0.25\textwidth]{images/4_Results/2_train_sets_preds_viz/color_palette.png}}
    
        \raisebox{-\height}{\includegraphics[width=0.24\textwidth]{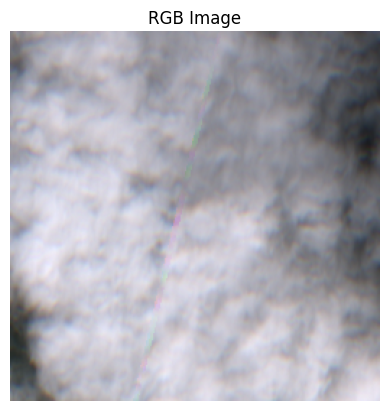}}
        \raisebox{-\height}{\includegraphics[width=0.24\textwidth]{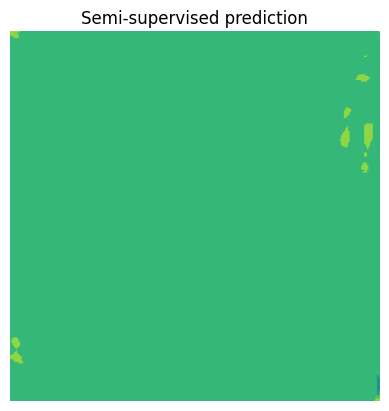}}
        \raisebox{-\height}{\includegraphics[width=0.24\textwidth]{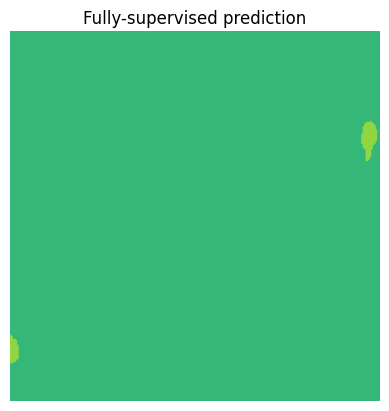}}
        \raisebox{-\height}{\includegraphics[width=0.24\textwidth]{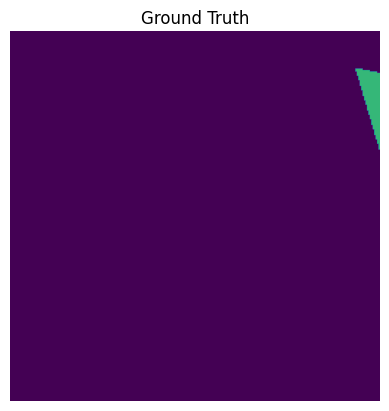}}

        \raisebox{-\height}{\includegraphics[width=0.24\textwidth]{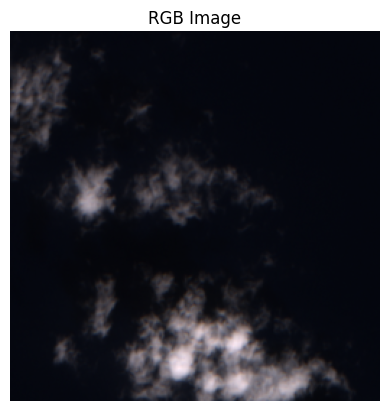}}
        \raisebox{-\height}{\includegraphics[width=0.24\textwidth]{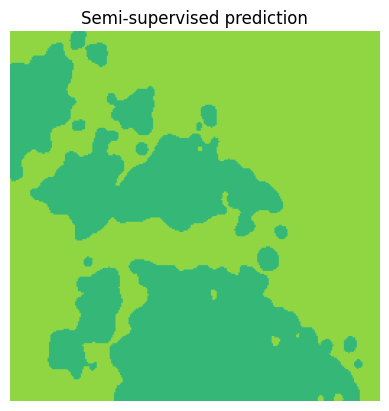}}
        \raisebox{-\height}{\includegraphics[width=0.24\textwidth]{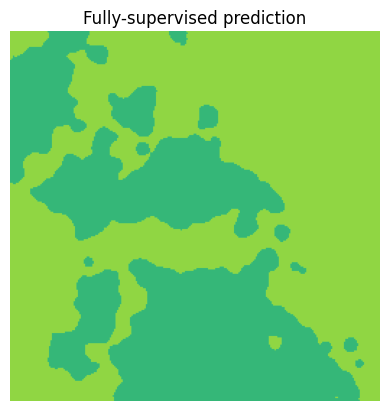}}
        \raisebox{-\height}{\includegraphics[width=0.24\textwidth]{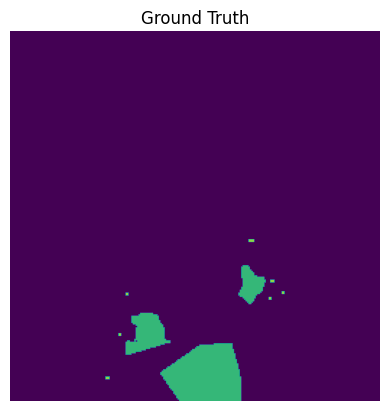}}

        \raisebox{-\height}{\includegraphics[width=0.24\textwidth]{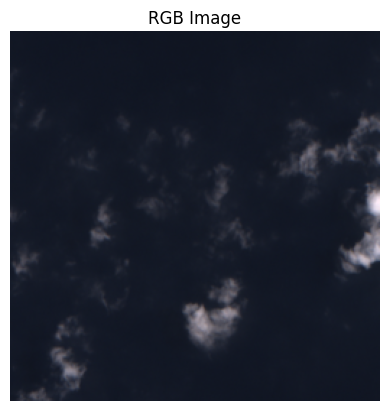}}
        \raisebox{-\height}{\includegraphics[width=0.24\textwidth]{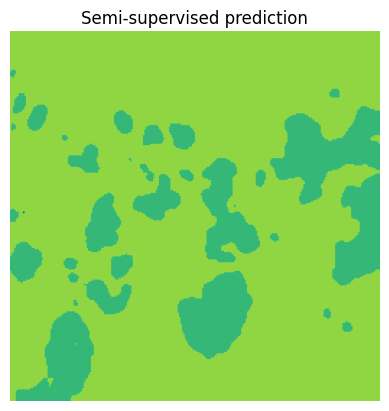}}
        \raisebox{-\height}{\includegraphics[width=0.24\textwidth]{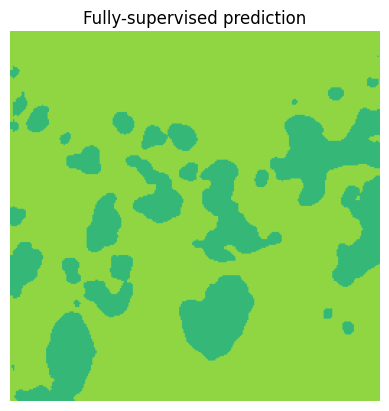}}
        \raisebox{-\height}{\includegraphics[width=0.24\textwidth]{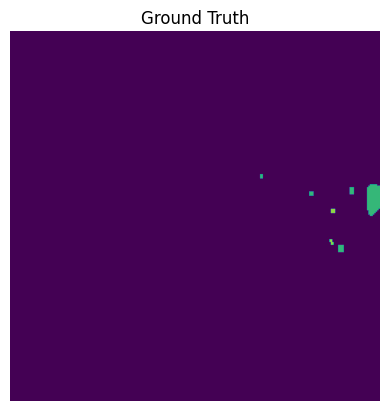}}

        \raisebox{-\height}{\includegraphics[width=0.24\textwidth]{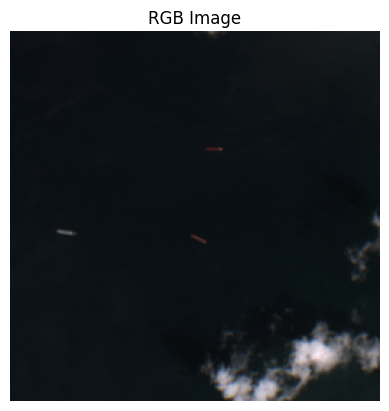}}
        \raisebox{-\height}{\includegraphics[width=0.24\textwidth]{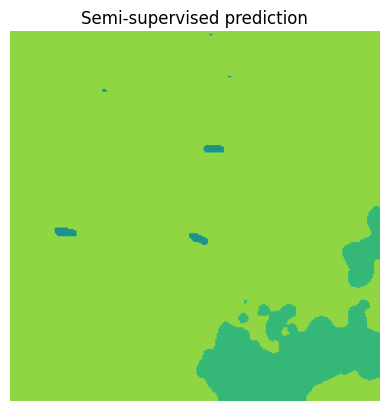}}
        \raisebox{-\height}{\includegraphics[width=0.24\textwidth]{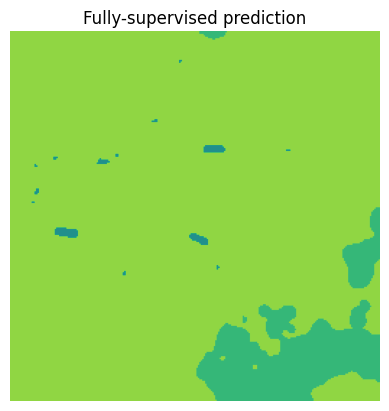}}
        \raisebox{-\height}{\includegraphics[width=0.24\textwidth]{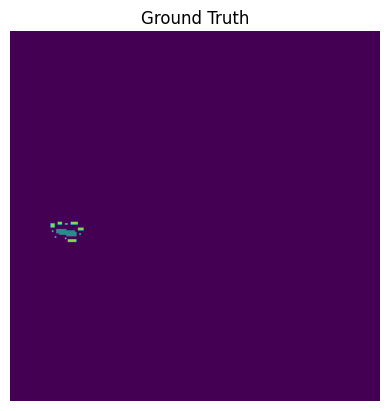}}

    \end{subfigure}
    \hfill
    \caption{Visualizing some MARIDA patches from the test set, their semantic segmentation maps, and the predictions made by semi-supervised and fully-supervised models trained using 70\% of labeled data. Each row denotes a unique patch, while each column denotes the image type. The color legend is at the top of the figure.}
    \label{Figure:70_perc_labeled_data_clouds}
\end{figure}

Figure \ref{Figure:70_perc_labeled_data_clouds} shows four MARIDA patches containing clouds. Accompanying each patch are their corresponding semantic segmentation maps, along with the predictions of both semi-supervised and fully-supervised models trained using 70\% of labeled data.

It can be noticed how both models seem to accurately detect the clouds and their predictions look quite similar. This pattern of similarity was also observed across the majority of predictions made on test set patches that contained cloud pixels.

Given these observations, we decided to examine the confusion matrix of the semi-supervised model (Table \ref{table:conf-matrix-70-perc-two-train-sets-semi-sup}) in an attempt to discern the reason behind the observed decrease in cloud detection performance by the semi-supervised model. The major issue encountered by the semi-supervised model was the misclassification of 19.64\% (6,449 out of 32,843) “Cloud” pixels, which were predicted as “Water” by the semi-supervised model.

Upon a thorough visual examination of all predictions made by the semi-supervised model on the test set, we found that 91.63\% (5,909 out of 6,449 ) of these false negatives were contained in just three test images (out of approximately 40-50 test images containing pixels labeled as "Cloud").
These three test patches are depicted in Figure \ref{Figure:70_perc_labeled_data_clouds_majority_of_errors}. Specifically, the semi-supervised model’s predictions on these three patches resulted in 4507, 1024, and 378 false negative cloud pixels, respectively, starting from the first row of Figure \ref{Figure:70_perc_labeled_data_clouds_majority_of_errors}.

This observation suggests that, had these three patches been excluded from the test set, the performance gap of -17\% in cloud detection between the semi-supervised and fully-supervised models would have been negligible because the "Cloud" \gls{IoU} of the semi-supervised model would have increased from 0.80 to 0.98.

\begin{table}[H]
\caption{Confusion matrix of the predictions on the test set of the semi-supervised model trained with 70\% of labeled data. \textit{A/OM} stands for the class "Algae/Organic Material". \textit{Sum (gt)} indicates the sum of the pixels in the ground truth segmentation maps for that class, while \textit{Sum (pred)} denotes the sum of the pixels predicted as that class. The true positive pixels of each class are reported in the main diagonal going from the top-left cell to the bottom-right of the matrix. The false positive  and false negative pixels of each class are respectively reported in each column and row of that class. The number of false negatives of the "Cloud" class that were mistakenly classified as Water are highlighted in gray.}
\label{table:conf-matrix-70-perc-two-train-sets-semi-sup}
\centering
\begin{adjustbox}{width=\columnwidth,center}
\begin{tabular}{lllllll}
    \toprule
    Class         & Marine Debris & A/OM   & Ship   & Clouds           & Water & Sum (gt)                    \\ 
    \midrule
    Marine Debris & \multicolumn{1}{r}{335}         & \multicolumn{1}{r}{16}   & \multicolumn{1}{r}{12}   & \multicolumn{1}{r}{0}              & \multicolumn{1}{r}{14}         & \multicolumn{1}{r}{377}                       \\ 
    A/OM          & \multicolumn{1}{r}{22}          & \multicolumn{1}{r}{1643} & \multicolumn{1}{r}{2}    & \multicolumn{1}{r}{0}              & \multicolumn{1}{r}{23}         & \multicolumn{1}{r}{1690} \\ 
    Ship          & \multicolumn{1}{r}{13}          & \multicolumn{1}{r}{0}    & \multicolumn{1}{r}{1077} & \multicolumn{1}{r}{0}              & \multicolumn{1}{r}{75}         & \multicolumn{1}{r}{1165}                      \\ 
    Clouds        & \multicolumn{1}{r}{0}           & \multicolumn{1}{r}{0}    & \multicolumn{1}{r}{15}   & \multicolumn{1}{r}{26379} & \multicolumn{1}{r}{\cellcolor[gray]{0.5}6449}       & \multicolumn{1}{r}{32843}            \\ 
    Water  & \multicolumn{1}{r}{88}          & \multicolumn{1}{r}{53}   & \multicolumn{1}{r}{99}   & \multicolumn{1}{r}{91}             & \multicolumn{1}{r}{158217}     & \multicolumn{1}{r}{158548}                    \\ 
    \midrule
    Sum (pred)    & \multicolumn{1}{r}{458}         & \multicolumn{1}{r}{1712} & \multicolumn{1}{r}{1205} & \multicolumn{1}{r}{26470} & \multicolumn{1}{r}{164778}     &                             \\ 
    IoU           & \multicolumn{1}{r}{0.67}          & \multicolumn{1}{r}{0.93}   & \multicolumn{1}{r}{0.83}   & \multicolumn{1}{r}{0.8}              & \multicolumn{1}{r}{0.96}         &                             \\ 
    \bottomrule
\end{tabular}
\end{adjustbox}
\end{table}

\subsubsection{Marine Debris}

Subsequently, we examined the predictions made by both semi-supervised and fully-supervised models for the “Marine Debris” class when using 40\% of labeled data. This percentage is associated to the maximum performance decrease (-15\%) of semi-supervised models in comparison to fully-supervised models when predicting the the “Marine Debris” class.

\begin{figure}[H]
    \centering
    \begin{subfigure}[t]{\textwidth}
    \centering
        \raisebox{-\height}{\includegraphics[width=0.25\textwidth]{images/4_Results/2_train_sets_preds_viz/color_palette.png}}
    
        \raisebox{-\height}{\includegraphics[width=0.24\textwidth]{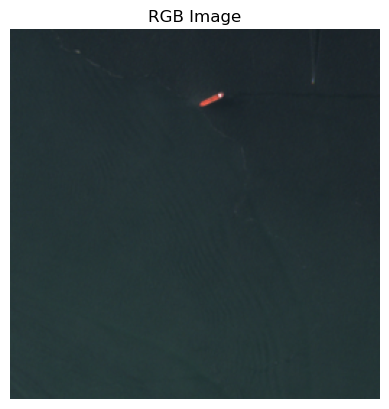}}
        \raisebox{-\height}{\includegraphics[width=0.24\textwidth]{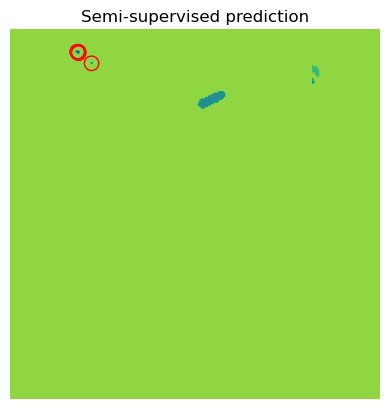}}
        \raisebox{-\height}{\includegraphics[width=0.24\textwidth]{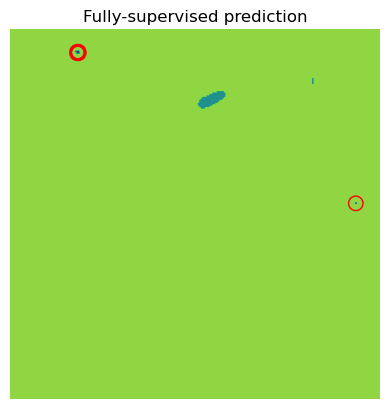}}
        \raisebox{-\height}{\includegraphics[width=0.24\textwidth]{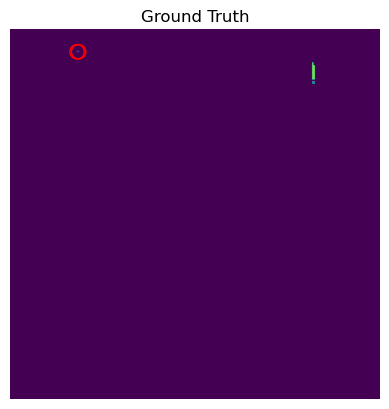}}
        
        \raisebox{-\height}{\includegraphics[width=0.24\textwidth]{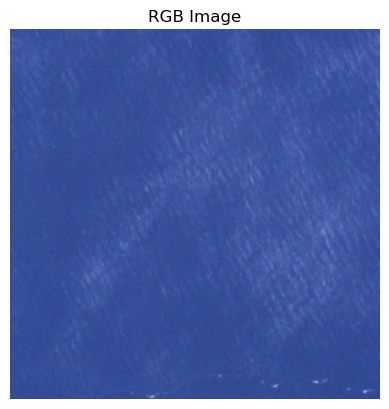}}
        \raisebox{-\height}{\includegraphics[width=0.24\textwidth]{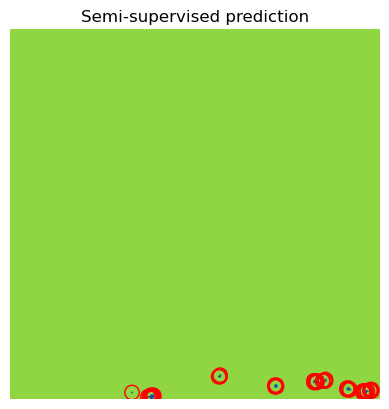}}
        \raisebox{-\height}{\includegraphics[width=0.24\textwidth]{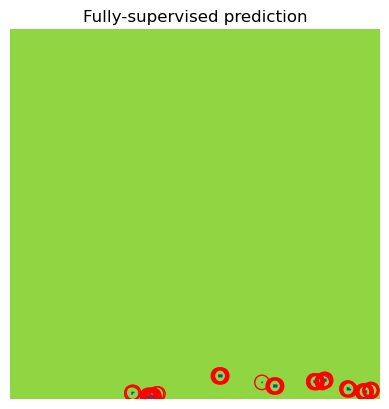}}
        \raisebox{-\height}{\includegraphics[width=0.24\textwidth]{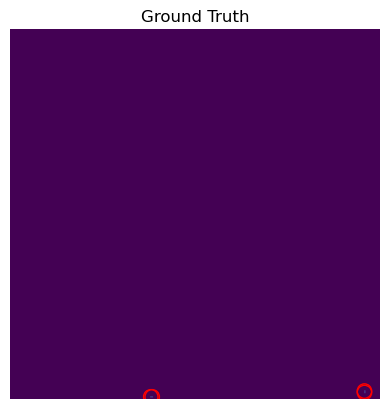}}

        \raisebox{-\height}{\includegraphics[width=0.24\textwidth]{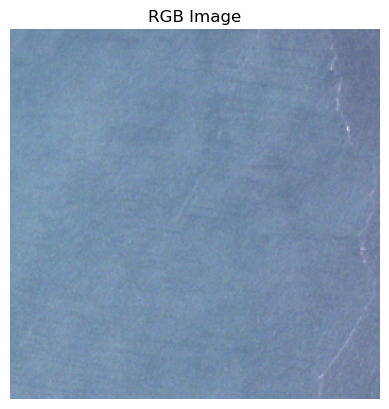}}
        \raisebox{-\height}{\includegraphics[width=0.24\textwidth]{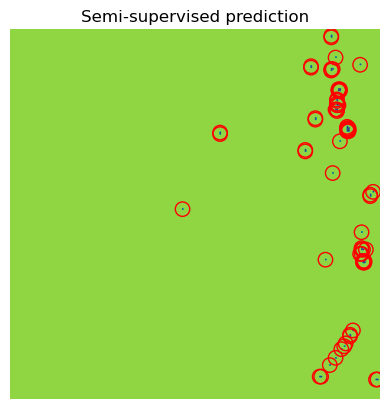}}
        \raisebox{-\height}{\includegraphics[width=0.24\textwidth]{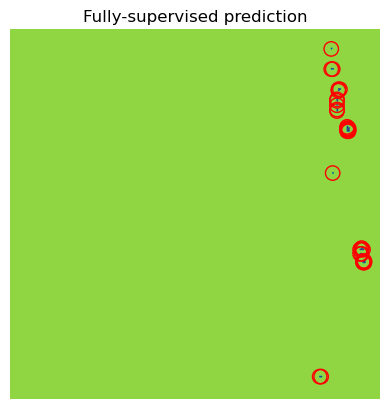}}
        \raisebox{-\height}{\includegraphics[width=0.24\textwidth]{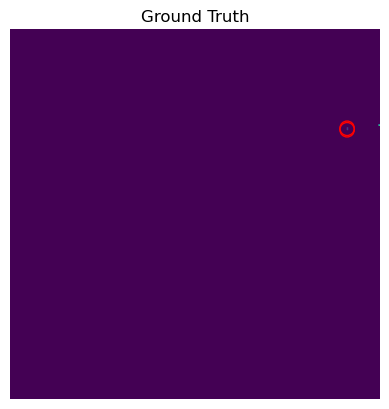}}

        \raisebox{-\height}{\includegraphics[width=0.24\textwidth]{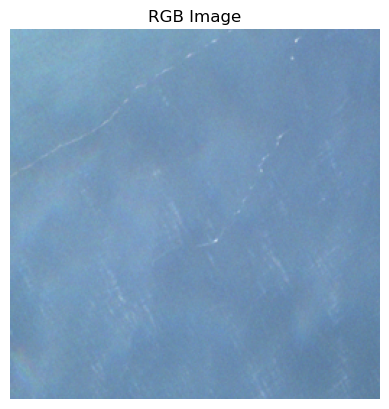}}
        \raisebox{-\height}{\includegraphics[width=0.24\textwidth]{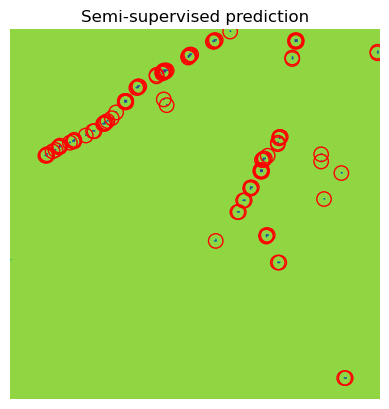}}
        \raisebox{-\height}{\includegraphics[width=0.24\textwidth]{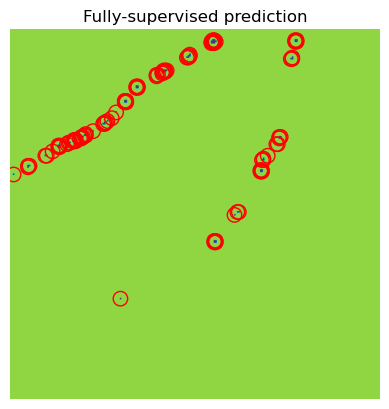}}
        \raisebox{-\height}{\includegraphics[width=0.24\textwidth]{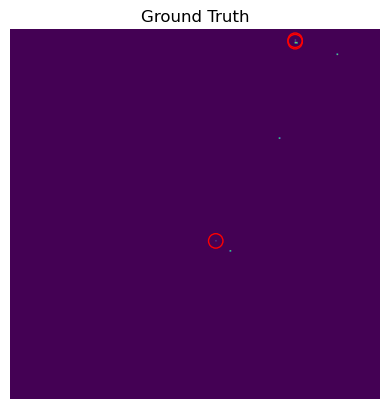}}
    \end{subfigure}
    \hfill
    \caption{Visualizing some MARIDA patches from the test set, their semantic segmentation maps, and the predictions made by semi-supervised and fully-supervised models trained using 40\% of labeled data. Each row denotes a unique patch, while each column denotes the image type. The color legend is at the top of the figure. Each pixel labeled as "Marine Debris" in the predictions and ground truth segmentation maps is highlighted with a red circle.}
    \label{Figure:md_ssl_vs_fs_40_perc_labeled_data}
\end{figure}

Figure \ref{Figure:md_ssl_vs_fs_40_perc_labeled_data} illustrates four MARIDA patches containing marine debris. Accompanying each patch are their corresponding semantic segmentation maps, along with the predictions of both semi-supervised and fully-supervised models trained using 40\% of labeled data. The marine debris pixels in both prediction and ground truth segmentation maps are highlighted with red circles to make them more visible and to avoid confusion with other labeled pixels.

In the first two patches (first two rows of Figure \ref{Figure:md_ssl_vs_fs_40_perc_labeled_data}) both the semi-supervised and fully-supervised models accurately detect the annotated marine debris pixels. They also identify additional debris on unlabeled pixels. 

In the last two patches, both models continue to correctly identify marine debris and to detect additional debris on unlabeled pixels. However, in these two cases, the semi-supervised model detects more debris than the fully-supervised model on unlabeled pixels.

Figure \ref{Figure:md-40-two-train-sets-additional-examples} illustrates additional patches containing marine debris and their corresponding semantic segmentation and predictions.

After the visual inspection of both models' predictions, we decided to examine the confusion matrices of the fully-supervised and semi-supervised models to understand the cause of the 15\% performance drop in detecting marine debris of the semi-supervised model compared to the fully-supervised one.

Tables \ref{table:conf-matrix-40-perc-two-train-sets-fully-sup} and \ref{table:conf-matrix-40-perc-two-train-sets-semi-sup} reveal that the semi-supervised model correctly detects five more marine debris pixels than the fully-supervised model. However, its score is lowered due to the detection of false positives for the “Water” class. This is because the \gls{IoU} of a class is calculated by the area of intersection of predicted and ground truth pixels of that class, divided by the area of union of predicted and ground truth pixels of that class (see Subsection \ref{subsubsection:miou}).

\begin{table}[H]
\caption{Confusion matrix of the predictions on the test set of the fully-supervised model trained with 40\% of labeled data. \textit{A/OM} stands for the class "Algae/Organic Material". \textit{Sum (gt)} indicates the sum of the pixels in the ground truth segmentation maps for that class, while \textit{Sum (pred)} denotes the sum of the pixels predicted as that class. The true positive pixels of each class are reported in the main diagonal going from the top-left cell to the bottom-right of the matrix. The false positive  and false negative pixels of each class are respectively reported in each column and row of that class. The number of true positives of the Marine "Debris class" that were correctly detected are highlighted in gray.}
\label{table:conf-matrix-40-perc-two-train-sets-fully-sup}
\centering
\begin{adjustbox}{width=\columnwidth,center}
\begin{tabular}{lllllll}
    \toprule
    Class         & Marine Debris & A/OM   & Ship   & Cloud   & Water    & Sum (gt) \\ 
    \midrule
    Marine Debris & \multicolumn{1}{r}{\cellcolor[gray]{0.5}342}         & \multicolumn{1}{r}{9}    & \multicolumn{1}{r}{8}    & \multicolumn{1}{r}{0}     & \multicolumn{1}{r}{18}     & \multicolumn{1}{r}{377}    \\ 
    A/OM          & \multicolumn{1}{r}{33}          & \multicolumn{1}{r}{1580} & \multicolumn{1}{r}{0}    & \multicolumn{1}{r}{0}     & \multicolumn{1}{r}{77}     & \multicolumn{1}{r}{1690}   \\ 
    Ship          & \multicolumn{1}{r}{15}          & \multicolumn{1}{r}{0}   & \multicolumn{1}{r}{1060} & \multicolumn{1}{r}{0}     & \multicolumn{1}{r}{90}     & \multicolumn{1}{r}{1165}   \\ 
    Clouds        & \multicolumn{1}{r}{0}           & \multicolumn{1}{r}{0}    & \multicolumn{1}{r}{0}    & \multicolumn{1}{r}{32260} & \multicolumn{1}{r}{583}    & \multicolumn{1}{r}{32843}  \\ 
    Water  & \multicolumn{1}{r}{51}          & \multicolumn{1}{r}{17}   & \multicolumn{1}{r}{111}  & \multicolumn{1}{r}{280}   & \multicolumn{1}{r}{158089} & \multicolumn{1}{r}{158548} \\ 
    \midrule
    Sum (pred)    & \multicolumn{1}{r}{441}         & \multicolumn{1}{r}{1606} & \multicolumn{1}{r}{1179} & \multicolumn{1}{r}{32540} & \multicolumn{1}{r}{158857} &          \\
    IoU           & \multicolumn{1}{r}{0.72}          & \multicolumn{1}{r}{0.92}   & \multicolumn{1}{r}{0.83}   & \multicolumn{1}{r}{0.97}    & \multicolumn{1}{r}{0.99}     &                               \\ 
    \bottomrule
\end{tabular}
\end{adjustbox}
\end{table}

\begin{table}[H]
\caption{Confusion matrix of the predictions on the test set of the semi-supervised model trained with 40\% of labeled data. \textit{A/OM} stands for the class "Algae/Organic Material". \textit{Sum (gt)} indicates the sum of the pixels in the ground truth segmentation maps for that class, while \textit{Sum (pred)} denotes the sum of the pixels predicted as that class. The true positive pixels of each class are reported in the main diagonal going from the top-left cell to the bottom-right of the matrix. The false positive  and false negative pixels of each class are respectively reported in each column and row of that class. The number of true positives of the "Marine Debris" class that were correctly detected and the "Water" pixels mistakenly classified as "Marine Debris" are highlighted in gray.}
\label{table:conf-matrix-40-perc-two-train-sets-semi-sup}
\centering
\begin{adjustbox}{width=\columnwidth,center}
\begin{tabular}{lllllll}
    \toprule
    Class         & Marine Debris & A/OM   & Ship   & Cloud   & Water    & Sum (gt) \\ 
    \midrule
    Marine Debris & \multicolumn{1}{r}{\cellcolor[gray]{0.5}347}         & \multicolumn{1}{r}{4}    & \multicolumn{1}{r}{14}   & \multicolumn{1}{r}{0}     & \multicolumn{1}{r}{12}     & \multicolumn{1}{r}{377}    \\ 
    A/OM          & \multicolumn{1}{r}{32}         & \multicolumn{1}{r}{1578} & \multicolumn{1}{r}{2}    & \multicolumn{1}{r}{0}     & \multicolumn{1}{r}{78}     & \multicolumn{1}{r}{1690}   \\ 
    Ship          & \multicolumn{1}{r}{20}          & \multicolumn{1}{r}{0}    & \multicolumn{1}{r}{1084} & \multicolumn{1}{r}{0}     & \multicolumn{1}{r}{61}     & \multicolumn{1}{r}{1165}   \\ 
    Clouds        & \multicolumn{1}{r}{0}           & \multicolumn{1}{r}{0}    & \multicolumn{1}{r}{0}    & \multicolumn{1}{r}{32622} & \multicolumn{1}{r}{221}    & \multicolumn{1}{r}{32843}  \\ 
    Water  & \multicolumn{1}{r}{\cellcolor[gray]{0.5}177}         & \multicolumn{1}{r}{25}   & \multicolumn{1}{r}{85}   & \multicolumn{1}{r}{277}   & \multicolumn{1}{r}{157984} & \multicolumn{1}{r}{158548} \\
    \midrule
    Sum (pred)    & \multicolumn{1}{r}{576}         & \multicolumn{1}{r}{1607} & \multicolumn{1}{r}{1185} & \multicolumn{1}{r}{32899} & \multicolumn{1}{r}{158356} &          \\ 
    IoU           & \multicolumn{1}{r}{0.57}          & \multicolumn{1}{r}{0.92}   & \multicolumn{1}{r}{0.86}   & \multicolumn{1}{r}{0.98}    & \multicolumn{1}{r}{0.99}     &                               \\ 
    \bottomrule
\end{tabular}
\end{adjustbox}
\end{table}

\cleardoublepage

\chapter{Discussion}
\label{ch:discussion}


The purpose of this chapter is to discuss the results of the experiments conducted in our work.

Section \ref{sec:discussion_prel_exp} discusses the results of the two preliminary hyperparameter selection experiments. Section \ref{subsec:disc-fs-vs-ssl-2-train-sets} discusses the results of the main experiment, which compares the performance of semi-supervised and fully-supervised models. 

\section{Preliminary Experiments}
\label{sec:discussion_prel_exp}

This section discusses the results of two hyperparameter selection experiments we conducted (refer to Section \ref{sec:exp-preliminary-1} for the details of the methodology of these experiments). Specifically, 
Subsection \ref{exp_disc:finding_best_threshold_max_val_miou_10} discusses the results of the experiment aimed at determining the optimal probability threshold ($\tau$ in Eq. \ref{eq:Lu_fixmatch_sem_seg}) under conditions of limited labeled training data.
Finally, Subsection \ref{subsec:disc-ce-vs-focal-loss} discusses the use of cross-entropy and focal loss in the unsupervised component of semi-supervised model loss.

\subsection{Finding the Best Probability Threshold}
\label{exp_disc:finding_best_threshold_max_val_miou_10}

The aim of this experiment is to determine the optimal probability threshold in FixMatch for Semantic Segmentation that yields the best performance.

To find the answer, we conducted training on five semi-supervised learning models. All models had identical hyperparameters, with the exception of the probability threshold, as outlined in Table \ref{table:hyperparams_best_prob_threshold}. Due to time limitations, we used only 10\% of the labeled training data for model training.

The \gls{mIoU} scores presented in Table \ref{table:finding_best_threshold_max_val_miou_10} indicate that a probability threshold of 0.9 yields the best results both on the validation and test sets. As depicted in Figure \ref{Figure:finding_best_threshold_unsup_loss_prob_thresholds}, a higher probability threshold results in a reduced unsupervised loss component, particularly during the initial stages of training.

Conversely, a lower probability threshold leads to an increased unsupervised loss component. This component significantly outweighs the supervised component throughout all of the training process. We hypothesize that this imbalance, which amplifies the impact of the unsupervised component at the expense of the supervised component during the backward pass, is particularly detrimental at the beginning of training. This is when the model has not yet fully optimized its weights based on the supervised component and is thus more susceptible to generating incorrect pseudo-labels. 

Incorrect pseudo-labels may cause the model to suffer from the confirmation bias \cite{arazo2020pseudo}, which further exacerbates the model’s learning process. We think that this is why a very high threshold of 0.9 performs best. However, it could be pointed out that a threshold of 0.6 yielded a higher score than the thresholds of 0.7 and 0.8 (refer to Table \ref{table:finding_best_threshold_max_val_miou_10}). This may be due to the fact that we trained and evaluated a single model for each threshold value, and did not average the performance of multiple trained models.

\subsection{Cross Entropy Vs Focal loss in the Unsupervised Component}
\label{subsec:disc-ce-vs-focal-loss}

This experiment compares the use of cross entropy and the focal loss in the unsupervised component of semi-supervised learning models.

When we introduced the formulation of the focal loss in Subsection \ref{subsubsec:focal-loss}, we highlighted that the cross entropy ($\gamma = 0$) function is higher than the focal loss function (see Figure \ref{Figure:focal_loss}). 

In this experiment we use a probability threshold of $0.9$ for the unsupervised component of the loss. This means that only pseudo-labels with a predicted probability of $0.9$ or higher are retained when computing the unsupervised component of the loss. For this reason, we hypothesized that using the cross entropy, which yields higher values than the focal loss (also for pseudo-labels predicted with a high probability, as the focal loss is close to zero), could potentially contribute more to the unsupervised component and yield better results.

However, we can notice from the results on the test set of Table \ref{table:mIoU_exp_ce_vs_focal} that the cross entropy and the focal loss led to \gls{mIoU} scores that differ only by $0.01$. Moreover, Figures \ref{Figure:ce_vs_focal_sup_loss} and \ref{Figure:ce_vs_focal_unsup_loss} respectively show that the trend of the supervised and the unsupervised components of the total loss are very similar.

Nonetheless, minor variations can be seen in Figure \ref{Figure:ce_vs_focal_unsup_loss}. In particular, the trend of the unsupervised component appears to be higher for the cross entropy case during the initial 200-250 epochs and the concluding 900 epochs.


We think that during the initial training phase, the model is still not yet fully optimized
, and so the unsupervised loss computed as cross entropy is higher than the one computed as focal loss. This behaviour is expected, since only samples predicted with a probability greater or equal to 0.9 are considered when computing the unsupervised component. Therefore, as stated before, the cross entropy function outputs higher values than the focal loss function. 

However, we are not sure how to justify the very similar trends of the unsupervised component with the cross entropy and the focal loss during the middle training stage. 

It would be interesting to perform the same experiment with other percentages of labeled data too to see if the performance of using the two losses as the unsupervised component is still similar or not.

\section{Main Experiment - Comparing Semi-Supervised and Fully-Supervised Learning Models}
\label{subsec:disc-fs-vs-ssl-2-train-sets}

This section discusses the results of the 
main experiment 
of our work (refer to Section \ref{subsec:exp_design_exp_2_train_sets} for the details of the methodology of this 
experiment
).
Specifically, the difference of performance between fully-supervised and semi-supervised learning models is discussed in Subsection \ref{subsec:perf-ssl-vs-fs-discussion}. Instead, Subsection \ref{subsec:viz-ssl-vs-fs-discussion} discusses the differences in the visual predictions of semi-supervised and fully-supervised models.

\subsection{Comparing the Performance of Semi-Supervised and Fully-Supervised Learning Models}
\label{subsec:perf-ssl-vs-fs-discussion}

In this Subsection, we discuss the difference of performance among semi-supervised and fully-supervised models and suggest two hypothesis that may explain the observed difference. Moreover, Subsubsection \ref{subsubsec:connection-val-loss-miou-discussion} discusses the importance of monitoring not only the evaluation metric trend, but also the loss trend. This happened to be particularly crucial when the scores of the evaluation metric of fully- and semi-supervised models on the validation set (but not on the test set) were equal.


As outlined in Tables \ref{table:mIoU-fully-vs-ssl-val} and \ref{table:mIoU-fully-vs-ssl}, semi-supervised models tend to outperform fully-supervised models when trained with a smaller fraction of labeled data. 
We claim that this happens because fully-supervised models have just very little data to learn from, while semi-supervised models have more data available. 
However, the trend reverses when a larger fraction of labeled data is used for training, with fully-supervised models showing superior performance.

Furthermore, as observed in Subsection \ref{subsec:perf-comp-ssl-vs-fs-res}, when semi-supervised models surpass the performance of fully-supervised models, the margin of superiority is typically larger.

It might be suggested that training should also have been conducted with 90\% and 95\% labeled data for fairness, especially considering that we also trained models by only using 5\% and 10\% of labeled data. However, it is worth noticing that the 20\% case already demonstrates a significant performance increase for the semi-supervised model (+8\%). This increase is notably larger than the performance increase observed for the fully-supervised model (+1\%) in the 80\% scenario. Thus, the current testing range is deemed sufficient.

We are now going to name two potential factors that could explain why semi-supervised models tend to outperform fully-supervised models when trained with low percentages of labeled data, but not when trained with high percentages of labeled data.

The first hypothesis is that, with high percentages of labeled data, the task of marine anomaly detection already achieves high \gls{mIoU} scores using just the labeled data. Consequently, it may be that only the pseudo-labels of easy samples surpass the probability threshold. These easy samples, such as pixels of dense white clouds that are unlikely to be confused with other classes, are the pixels that the semi-supervised model is already proficient at classifying.
This may be the reason of why semi-supervised models do not provide additional improvements when trained with high percentages of labeled data.

The second hypothesis is that semi-supervised models experience a performance drop due to a reduction in unlabeled data. Specifically, in the “Two Training Sets” setup (refer to Subsection \ref{subsec:two-train-set}), an increase in labeled data corresponds to a decrease in unlabeled data. A patch has both labeled and unlabeled pixels and is either part of the labeled set $\mathcal{D}^l$ (where only its labeled pixels are used during training) or part of the unlabeled set $\mathcal{D}^u$ (where all its pixels, even if labeled, are considered as unlabeled during training). In general, the majority of the pixels of a patch of the MARIDA dataset are unlabeled, whilst the minority are labeled. Thus, moving a patch from $\mathcal{D}^u$ to $\mathcal{D}^l$ means that while some pixels will now be considered as labeled, a significantly larger number of unlabeled pixels will be completely excluded from the training process.
So, the more patches are in $\mathcal{D}^l$, the fewer patches there will be in $\mathcal{D}^u$, resulting in less unlabeled data for the semi-supervised model to train on.

One or both of these explanations could potentially also account for the performance plateau observed in semi-supervised models (as seen in Figure \ref{Figure:val_mIoU_2_train_sets_ssl_models_all_percentages} and Tables \ref{table:mIoU-fully-vs-ssl-val} and \ref{table:mIoU-fully-vs-ssl}) when the percentage of labeled data increases from 60\% to 80\%, while the performance of fully-supervised models continues to improve.

\subsubsection{The Importance of Monitoring the Loss}
\label{subsubsec:connection-val-loss-miou-discussion}

By looking at Figure \ref{Figure:val_miou_and_loss_of_30_and_70} and Tables \ref{table:mIoU-fully-vs-ssl-val} and \ref{table:mIoU-fully-vs-ssl}, we observed that the validation loss is a valuable metric to consider when semi-supervised and fully-supervised models exhibit similar performance on the validation set, display comparable trends for the validation \gls{mIoU} scores, but differ in their performance on the test set. 
Specifically, we found that the model with the lowest validation loss during the final training stage also achieved the highest \gls{mIoU} on the test set, as it may be expected. This underscores the importance of monitoring the loss trend, not just the trend of the evaluation metric.

\subsection{Visually Comparing Predictions}
\label{subsec:viz-ssl-vs-fs-discussion}

Subsequently, we visually compared the prediction of the fully-supervised and semi-supervised models. However, due to time constraints, we could not inspect thoroughly the predictions for all of the five classes.

Therefore, we examined the performance fluctuations between semi-supervised and fully-supervised models across the five classes, taking into account all the tested percentages of labeled data. We observed that the classes “Clouds” and “Marine Debris” exhibited the most significant variance in performance.

\subsubsection{Cloud}

The highest performance improvement (+24\%) in cloud detection by the semi-supervised model occurred when training with the lowest percentage of labeled data (5\%). Despite the semi-supervised model not perfectly segmenting clouds, it often successfully detects some clouds that the fully-supervised model fails to identify (refer to Figure \ref{Figure:5_perc_labeled_data_clouds}).

In the scenario where the semi-supervised model exhibited the poorest performance in cloud detection compared to the fully-supervised model, which happened when training with 70\% of labeled data, we observed that the 17\% reduction on the test set was predominantly due to three patches (see Figure \ref{Figure:70_perc_labeled_data_clouds_majority_of_errors}) out of approximately 40-50 test patches containing pixels labeled as "Cloud". In fact, the predictions for the majority of the patches were quite similar and accurate for both the semi-supervised and fully-supervised models, as depicted in some examples in Figure \ref{Figure:70_perc_labeled_data_clouds}.

The semi-supervised model's predictions on the three patches that mainly contributed to the drop of performance in cloud detection are not entirely accurate, yet they are not completely wrong either. Moreover, the semi-supervised model seems to more accurately distinguish between water and cloud in the bottom-right part of the first image. It is also worth noting that the labeling of the first patch's ground truth appears somewhat unusual (refer to the first row of Figure \ref{Figure:70_perc_labeled_data_clouds_majority_of_errors}). In fact, we believe that some pixels, particularly the darker ones in the left portion of the RGB visualization of the patch, should not have been labeled as “Cloud”. 

So, we claim that, in this case, the semi-supervised model's performance is not actually worse, but appears to be an artifact of \first the bad labeling of one of these three test images, \second the low quantity of labeled pixels (low compared to the total amount of pixels contained in an image) of these three test images.

\subsubsection{Marine Debris}

Afterwards, we examined the case of most significant drop in performance of semi-supervised models (-15\%, refer to Table \ref{table:mIoU-difference-classes-tw-train-sets-ssl-vs-fs}) in detecting marine debris compared to fully-supervised models, which occurred when training with 40\% of labeled data. The semi-supervised model identifies slightly more true positive pixels than the fully-supervised model, as it can be observed in the confusion matrices shown in Tables \ref{table:conf-matrix-40-perc-two-train-sets-fully-sup} and \ref{table:conf-matrix-40-perc-two-train-sets-semi-sup}. The performance decline of the semi-supervised model is primarily due to the detection of more false positives (specifically, “Marine Debris” pixels predicted as “Water”).

Moreover, as depicted in the last two patches of Figure \ref{Figure:md_ssl_vs_fs_40_perc_labeled_data}, the semi-supervised model detects more marine debris on unlabeled pixels compared to its fully-supervised counterpart. These prediction on unlabeled data may be correct or not. Depending on the final application of the model, detecting more marine debris could be either advantageous or disadvantageous. For instance, if human verification is planned after the predictions, 
it might be more beneficial to detect more anomalies (resulting in more false positives) for subsequent verification of their correctness, rather than missing true anomalies (resulting in more false negatives).

\cleardoublepage

\chapter{Conclusions and Future work}
\label{ch:conclusionsAndFutureWork}

In this chapter, we derive conclusions from our work (Section \ref{sec:conclusions}), outline the limitations of our work (Section \ref{sec:limitations}) and future work (Section \ref{sec:futureWork}). Finally, we reflect on the ethical, societal and environmental implications of this project (Section \ref{sec:reflections}).

\section{Conclusions}
\label{sec:conclusions}

In our work, we tackled the task of semantically segmenting marine anomalies (marine debris, algae, ships, and clouds) when having limited labeled training data. We evaluated deep semi-supervised models, which utilize both labeled and unlabeled data, against fully-supervised models. Specifically, we implemented FixMatch for Semantic Segmentation as our semi-supervised approach, modifying an existing implementation of FixMatch \cite{sohn2020fixmatch} to handle semantic segmentation and multispectral images.


Firstly, we conducted two preliminary hyperparameter selection experiments, whose results were then used for our main experiment. The results showed that 
\first a high probability threshold of 0.9 was optimal for semi-supervised models with limited labeled training data, and \second the performance was similar when using cross entropy or focal loss as the unsupervised component of the loss of semi-supervised models.

In our main experiment, we compared fully-supervised and semi-supervised models trained with varying percentages of labeled data. Semi-supervised models outperformed fully-supervised models when trained with low percentages of labeled data, likely due to the additional unlabeled data. However, fully-supervised models performed better with high percentages of labeled data. We hypothesize that this may due to two reasons. Firstly, with more labeled data, the models achieve a high performance just by using labeled data. Therefore, it might be that only the pseudo-labels of easy samples exceed the probability threshold. These are the samples that the semi-supervised model is already good at classifying. Secondly, the performance of semi-supervised models may decrease due to the reduction of unlabeled data, which is given by the increase of labeled data (refer to Subsection \ref{subsec:perf-ssl-vs-fs-discussion} for a more detailed explanation).

We examined cases 
where the semi-supervised model's performance was at its lowest in comparison to the the fully-supervised model.  

In the scenario where the semi-supervised model had the poorest performance in cloud detection
, we found that the performance difference between the two models was primarily due to just three images in the test set, which had partial and/or unusual labeling. 

Instead, in the scenario where the semi-supervised model had the poorest performance in marine debris detection, we noticed that the semi-supervised model detected more true positives, despite performing worse overall. The cause of the drop in performance were "Water" pixels misdetected as “Marine Debris”. However, having more false positives could be advantageous for marine anomaly detection applications, especially if there are subsequent processing steps where humans verify the \gls{AI} model’s predictions.

In conclusion, we demonstrated the potential of semi-supervised learning for detecting marine anomalies, particularly when there is a scarcity of labeled data. This approach can expedite the deployment of anomaly detection applications by reducing the amount of data that needs to be annotated, thereby also reducing costs. To ensure that our results are relevant, we used a model architecture compatible with the computational capabilities of space hardware, indicating the possibility of onboard inference.



\section{Limitations}
\label{sec:limitations}

Our results are subject to certain limitations. Firstly, we made assumptions regarding the feasibility of deploying the \gls{AI} model onboard an artificial satellite. In our study, we utilized atmospherically corrected and perfectly co-registered and georeferenced images, which is not typically the case onboard artificial satellites. As such, the potential impact of atmospheric effects and co-registration errors was not considered in our analysis. Additionally, if the data is not georeferenced, it becomes impossible to transmit the georeferenced coordinates of any anomalies to ground stations.

Secondly, the number of labeled pixels of most of the classes we considered was quite limited. This scarcity of data makes our evaluation dependent on these few pixels, potentially limiting the generalization of our findings to other anomalies located in different regions of the world.

Thirdly, the MARIDA dataset lacks any pixels labeled as land. As a result, the model might identify numerous anomalies in images that are not entirely captured over a water body.

\section{Future work}
\label{sec:futureWork}

We will now focus on the future work that could be done to expand our project. Specifically, Subsection \ref{sec:impro_mode_perf} outlines ideas for further enhancing the performance of the deep learning models. Subsection \ref{sec:impro_mode_charact} suggests work to better characterize the model. Finally, Subsection \ref{sec:moving_model_onboard} outlines potential steps to make the model's onboard usage more realistic.  

\subsection{Improving Model Performance}
\label{sec:impro_mode_perf}

Additional experiments could be conducted to improve model performance. For instance, a larger hyperparameter space could be explored for each utilized hyperparameter. Due to time constraints and the computationally demanding nature of hyperparameter exploration, we only tested one or a few values for each hyperparameter.

Techniques for decaying the learning rate could also be employed, such as decaying it when a metric has ceased to improve.

Moreover, we aim to conduct the experiment titled “Finding the Best Probability Threshold” (Subsection \ref{exp_finding_best_thresh_results}) with larger percentages of labeled data. This is because our previous attempt was limited to a scenario with only 10\% of the data being labeled.

Lastly, it could be interesting to train the deep learning models for additional epochs and with more parameters to see if new optimal minima could be reached by leveraging the deep double descent phenomenon \cite{nakkiran2021deep}.

\subsection{Improving Model Characterization}
\label{sec:impro_mode_charact}

In the future, we aim to test the performance of semi-supervised models by training them using both labeled and unlabeled pixels of each image. This approach deviates from our previous method, where only the labeled or the unlabeled pixels of each image were used during training, as each image belonged either to the labeled or unlabeled training set.
Beyond the need for performance evaluation, this novel methodology offers the potential advantage of speeding up the training process. This efficiency stems from the requirement of only a single dataloader, thereby utilizing just one batch during each training step. 
The operations are applied directly to this batch, rather than sequentially to the labeled and then the unlabeled batch. While parallel computation could speed up operations when having two batches, hardware resources are often limited in space missions. Therefore, this new approach could be crucial in such scenarios.

Additionally, we plan to incorporate a broader range of evaluation metrics, to better assess the impacts of false positives and false negatives. There have been instances where semi-supervised models, despite detecting more true positives of marine anomalies, performed worse than the fully-supervised models. This decline in performance was primarily attributed to the semi-supervised models’ misdetection of marine anomalies as water. However, the semi-supervised models also detected more marine anomalies on unlabeled pixels, which we do not know if are correct or not. Having a higher number of anomaly detections could be beneficial for marine anomaly detection applications, particularly if there are subsequent stages where human operators verify the \gls{AI} model’s predictions.

Lastly, we intend to evaluate the deep learning models on labeled data of new regions to see it they generalize well. If they do not generalize well, it would be beneficial to add labeled data from other regions worldwide.

\subsection{Moving the Model Onboard}
\label{sec:moving_model_onboard}

In the future, we plan to consider aspects of having the model onboard that we did not take into account. Specifically, we intend to emulate the lack of co-registration among the multispectral bands and the lack of georefenced pixels by utilizing data products that are below Level-1C. 

Furthermore, we aim to test the model on space-rated hardware to evaluate its latency and power consumption. Other factors to consider are the limited bandwidth and communication windows between the satellite and the ground stations, which serve for transmitting predictions to the ground stations or updating the onboard model. Finally, we will consider potential challenges such as the impact of electromagnetic radiation on the hardware's functionality and constraints related to battery capacity.

\section{Ethical, Societal and Environmental Reflections}
\label{sec:reflections}

We believe that this project does not pose substantial ethical issues as it utilizes open-source satellite data, whose highest resolution is of 10 meters per pixel. This resolution is insufficient to identify individuals, thereby mitigating concerns related to privacy invasion.

In terms of societal implications, further development of this work could lead to the provision of notifications and locations of marine anomalies, benefiting various stakeholders. For example, non-governmental organizations and startups could coordinate the removal of detected marine debris. Maritime surveillance entities could monitor the traffic of ships, and space agencies and remote sensing researchers could benefit from cloud detection, enabling the retrieval of images with minimal cloud coverage.

The thesis has the potential to make a significant contribution to the \gls{UN}\enspace\gls{SDG} \cite{Agenda2023} number 
14 "Life Below Water", which 
advocates for the conservation and sustainable use of our oceans, seas, and marine resources to promote sustainable development.

Specifically, our project may contribute to the \gls{SDG} 14 by potentially detecting harmful algae blooms, which pose a serious threat to aquatic ecosystems health \cite{hallegraeff2003harmful, tsikoti2021review}.

In addition, our project supports \gls{SDG} 14 through the automatic detection of marine debris. This allows for timely notification to relevant authorities for its removal, leading to cleaner marine habitats. It also creates safer conditions for marine life that could otherwise be entangled in plastic, and aids in preventing the spread of alien species via marine litter \cite{lincoln2022marine}.

The automatic detection of ships enhances our potential contribution to \gls{SDG} 14. If further investigated by human operators, the detection of ships that are not expected to be in a certain area could aid in the identification of illegal fishing and overfishing activities.

Finally, utilizing \gls{AI} on board satellites could significantly expedite the detection and notification process for marine anomalies, thereby potentially reducing the latency in the detection, which is, generally, another factor of primary importance for monitoring applications.



\cleardoublepage
\renewcommand{\bibname}{References}
\addcontentsline{toc}{chapter}{References}

\ifbiblatex
    \printbibliography[heading=bibintoc]
\else
    \bibliography{00_main}
\fi


\appendix
\renewcommand{\chaptermark}[1]{\markboth{Appendix \thechapter\relax:\thinspace\relax#1}{}}
\chapter{Additional Examples of Marine Debris predictions}

\begin{figure}[H]
    \centering
    \begin{subfigure}[t]{\textwidth}
    \centering
        \raisebox{-\height}{\includegraphics[width=0.25\textwidth]{images/4_Results/2_train_sets_preds_viz/color_palette.png}}
    
        \raisebox{-\height}{\includegraphics[width=0.24\textwidth]{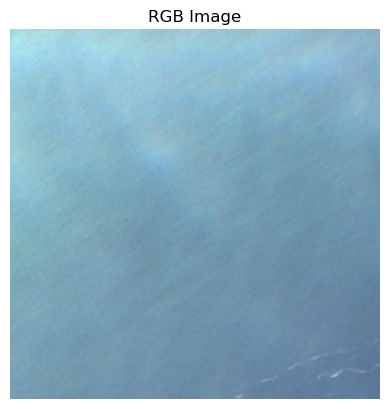}}
        \raisebox{-\height}{\includegraphics[width=0.24\textwidth]{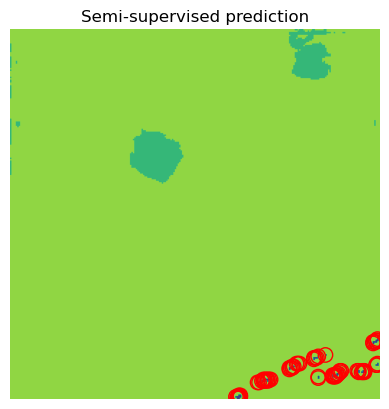}}
        \raisebox{-\height}{\includegraphics[width=0.24\textwidth]{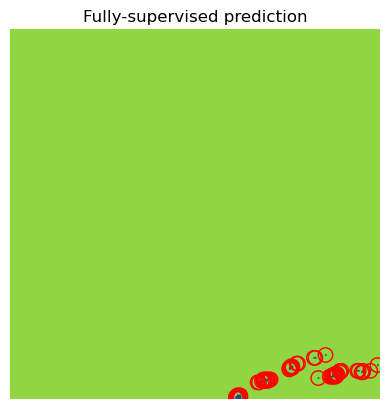}}
        \raisebox{-\height}{\includegraphics[width=0.24\textwidth]{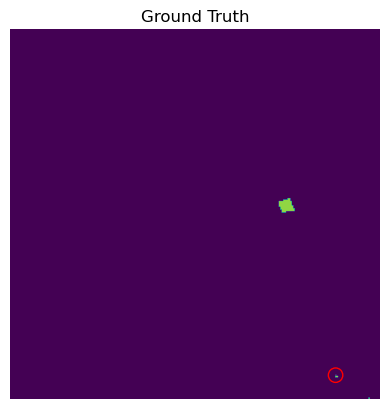}}

        \raisebox{-\height}{\includegraphics[width=0.24\textwidth]{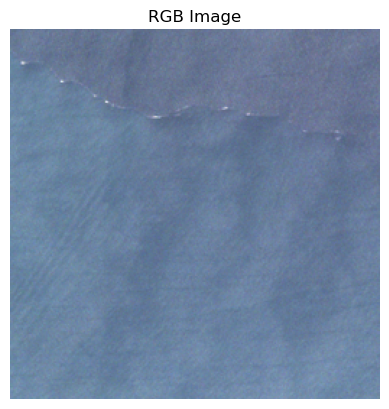}}
        \raisebox{-\height}{\includegraphics[width=0.24\textwidth]{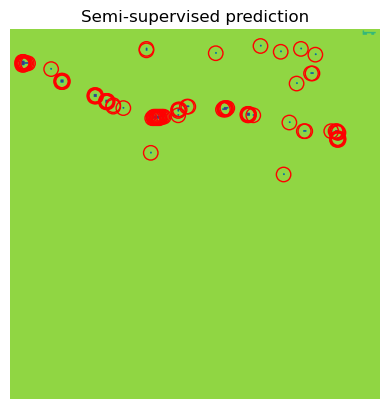}}
        \raisebox{-\height}{\includegraphics[width=0.24\textwidth]{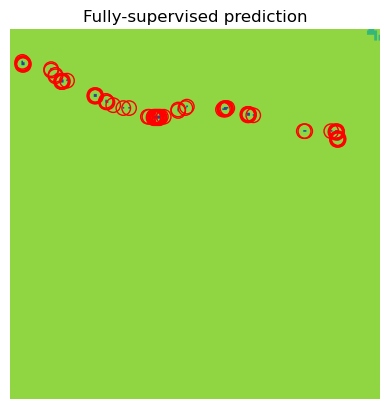}}
        \raisebox{-\height}{\includegraphics[width=0.24\textwidth]{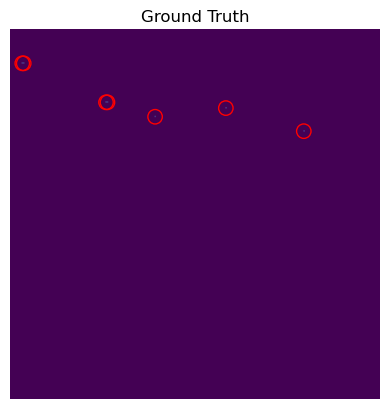}}

        \raisebox{-\height}{\includegraphics[width=0.24\textwidth]{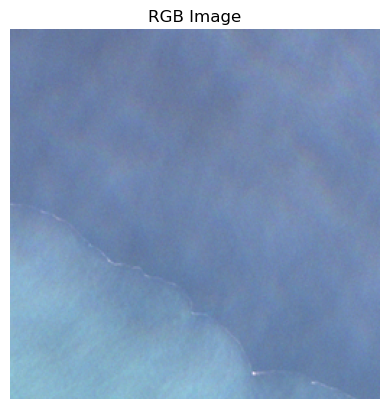}}
        \raisebox{-\height}{\includegraphics[width=0.24\textwidth]{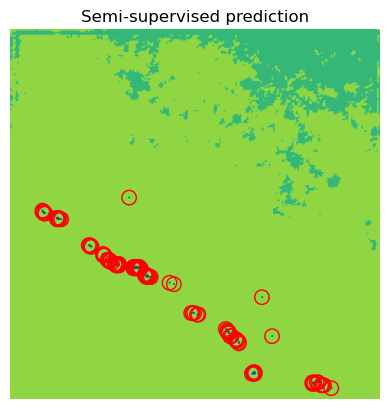}}
        \raisebox{-\height}{\includegraphics[width=0.24\textwidth]{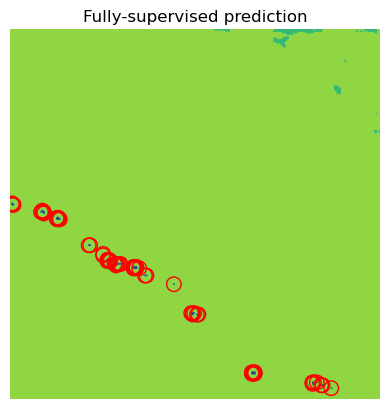}}
        \raisebox{-\height}{\includegraphics[width=0.24\textwidth]{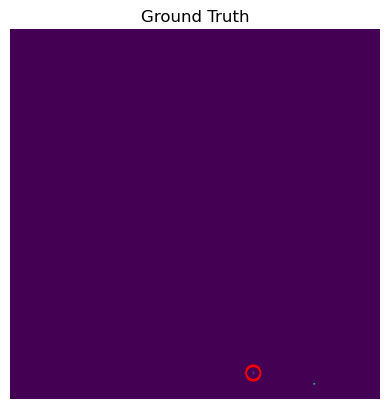}}
    \end{subfigure}
    \hfill
    \caption{Visualizing some MARIDA patches from the test set, their semantic segmentation maps, and the predictions made by semi-supervised and fully-supervised models trained using 40\% of labeled data. Each row denotes a unique patch, while each column denotes the image type. The color legend is at the top of the figure. Each pixel labeled as "Marine Debris" in the predictions and ground truth segmentation maps are highlighted with red circles.}
    \label{Figure:md-40-two-train-sets-additional-examples}
\end{figure}

Figure \ref{Figure:md-40-two-train-sets-additional-examples} shows additional patches containing marine debris and their corresponding semantic segmentation and predictions. The predictions are made by fully-supervised and semi-supervised models trained with 40\% of labeled data.

%
%
%
%
%
%
%


\label{pg:lastPageofMainmatter}

\kthbackcover
\fancyhead{}  



\end{document}